\documentclass[11pt,a4paper]{article}
\usepackage[hyperref]{eacl2021}
\usepackage[T5]{fontenc}

\usepackage[utf8]{inputenc}
\usepackage{times}
\usepackage{latexsym}

\usepackage{amssymb}
\usepackage{amsmath}
\usepackage{xspace} 
\usepackage{tablefootnote}
\usepackage{booktabs}
\usepackage{lingmacros}
\usepackage{colortbl}
\usepackage{graphicx} 
\usepackage{subfig} 

\usepackage{enumitem}

\setlist{nolistsep}
\usepackage{microtype}

\usepackage{soulutf8}

\usepackage{titlesec}

\iffalse
    \usepackage[final]{proofing}
  \renewcommand\hl[1]{{#1}}  
   {\draftnote{\red{#2}}}
   \newcommand\redHL[1]{}
  \newcommand\todo[1]{}

  \newcommand{\Djame}[1]{}
\newcommand{\bm}[1]{}
\newcommand{\bs}[1]{}
\newcommand{\ds}[1]{}
\newcommand{\ye}[1]{}

\else  

\usepackage[final]{proofing}
\aclfinalcopy  
\newcommand{\Djame}[1]{
{\textcolor{red}{\hl{Djame: #1}}}
}

\newcommand\red[1]{{\textbf{\textcolor{red}{#1}}}}

\let\oldred\red
\renewcommand\red[1]{{\bf \oldred{{#1}}}}

 \newcommand\redHL[1]{\red{\hl{#1}}}
\let\olddraftnote\draftnote
\renewcommand\draftnote[1]{\olddraftnote{\red{#1}}}

\newcommand{\bm}[1]{{\color{red}\textbf{BM}: #1}}
\newcommand{\bs}[1]{{\color{blue}\textbf{BS}: #1}}
\newcommand{\ds}[1]{{\color{orange}\textbf{DS}: #1}}
\newcommand{\ye}[1]{{\color{violet}\textbf{YE}: #1}}
\newcommand{\hg}[1]{{\color{cyan}\textbf{HG}: #1}}

\fi

\graphicspath{{../}{./}}

\usepackage{url}

\usepackage{graphicx}
\usepackage{tabu}
\usepackage{multirow}
\usepackage{microtype}
\usepackage{arydshln}

\definecolor{lightgreen}{rgb}{0.67, 0.94, 0.82}
\definecolor{darkred}{rgb}{0.6, 0.0, 0.0}
\definecolor{brightlavender}{rgb}{0.75, 0.58, 0.89}

\definecolor{mediumspringgreen}{rgb}{0.0, 0.98, 0.6}
\definecolor{inchworm}{rgb}{0.0, 0.98, 0.6}

\newcommand{\mbert}{mBERT\xspace}

\newcommand{\inlang}{same-language\xspace}
\newcommand{\crosslang}{cross-language\xspace}
\newcommand{\outdomain}{out-of-domain\xspace}
\newcommand{\outdist}{out-of-distribution\xspace}

\newcommand{\reinit}{\textsc{Random-init}\xspace}

\newcommand{\gap}{\textit{cross-lang gap}\xspace}

\newcommand{\similarity}{cross-lingual similarity\xspace}

\graphicspath{{../}{./}}

\newcommand{\standardFineTune}{\textsc{Ref}\xspace}

\title{First Align, then Predict: \\Understanding the Cross-Lingual Ability of Multilingual BERT 
}
\author{Benjamin Muller$^{ 1,2}$\quad Yanai Elazar$^{3,4}$\quad\large\textbf{ Beno\^it Sagot$^1$\quad Djam\'e Seddah$^1$}\\
   $^{1}$Inria, Paris, France \quad
  $^{2}$Sorbonne Universit\'e, Paris, France\\ $^{3}$Computer Science Department, Bar Ilan University\\
  $^{4}$Allen Institute for Artificial Intelligence\\
  \texttt{\{benjamin.muller, benoit.sagot,djame.seddah\}@inria.fr}\\ 
  \texttt{yanaiela@gmail.com}}

\date{}

\begin{document}
\maketitle
\begin{abstract}

Multilingual pretrained language models have demonstrated remarkable zero-shot cross-lingual transfer capabilities. Such transfer emerges by fine-tuning on a task of interest in one language and evaluating on a distinct language, not seen during the fine-tuning. Despite promising results, we still lack a proper understanding of the source of this transfer. Using a novel layer ablation technique and analyses of the model's internal representations, we show that multilingual BERT, a popular multilingual language model, can be viewed as the stacking of two sub-networks: a multilingual encoder followed by a task-specific language-agnostic predictor.
While the encoder is crucial for cross-lingual transfer and remains mostly unchanged during fine-tuning, the task predictor has little importance on the transfer and can be reinitialized during fine-tuning. We present extensive experiments with three distinct tasks, seventeen typologically diverse languages and multiple domains to support our hypothesis. 


\end{abstract}

\section{Introduction}
Zero-shot Cross-Lingual transfer 
aims at building models for a \textit{target} language by reusing knowledge acquired from a \textit{source} language. 
Historically, it has been tackled with a two-step \textit{standard cross-lingual pipeline}\draftnote{We should add the Tenney et al. 2019 paper, see Rev1 comment -ds \bm{Tenney et. al study only the mononlingual case}} \citep{ruder2019survey}: (1)~Building a shared multilingual representation of text, typically by aligning textual representations across languages. This step can be done using feature extraction \citep{aone1993language,schultz2001language} as with the delexicalized approach \cite{zeman2008cross,sogaard2011data} or using word embedding techniques \citep{mikolov2013exploiting,smith2017offline} by projecting monolingual embeddings onto a shared multilingual embedding space, this step requiring \draftreplace{annotated data}{explicit supervision signal} in the target language in the form of features or parallel data. (2)~Training a task-specific model using supervision on a source language on top of the shared representation. 

Recently, the rise of multilingual language models entailed a paradigm shift in this field. Multilingual pretrained language models \cite{devlin-etal-2019-bert,conneau2019cross} have been shown to perform efficient zero-shot cross-lingual transfer for many tasks and languages \cite{pires-etal-2019-multilingual,wu2019beto}. 
Such transfer relies on three-steps: (i)~pretraining a mask-language model (e.g.~\citet{devlin-etal-2019-bert}) 
on the concatenation of monolingual corpora across multiple languages, 
(ii)~fine-tuning the model on a specific task in the source language, and (iii)~using the fine-tuned model on  a target language.
The success of this approach is remarkable, and in contrast to the standard cross-lingual pipeline, the model sees neither aligned data nor task-specific annotated data in the target language at any training stage.

The source of such a successful transfer is still largely unexplained. \citet{pires-etal-2019-multilingual} hypothesize that these models learn shared multilingual representations during pretraining.
Focusing on syntax,  \citet{chi-etal-2020-finding} recently showed that the multilingual version of BERT (\mbert) \citep{devlin-etal-2019-bert}, encodes linguistic properties in shared multilingual sub-spaces.
Recently, \citet{gonen2020s} \draftreplace{suggests}{suggest} that \mbert learns a language encoding component and an abstract cross-lingual component.
In this work, we are interested in understanding the mechanism that leads \mbert to perform zero-shot cross-lingual transfer. More specifically, we ask \textbf{what parts of the model and what mechanisms support cross-lingual transfer?}

By combining behavioral and structural analyses \citep{belinkov2020interpretability}, we show that mBERT operates as the stacking of two modules: (1)~A multilingual encoder, located in the lower part of the model, critical for cross-lingual transfer, is in charge of aligning multilingual representations; and (2)~a task-specific, language-agnostic predictor which has little importance for cross-lingual transfer and is dedicated to performing the downstream task. This mechanism that emerges out-of-the-box, without any explicit supervision, suggests that mBERT behaves like the standard cross-lingual pipeline\draftnote{We had those references before \cite{aone1993language,schultz2001language,zeman2008cross,sogaard2011data,mikolov2013exploiting,smith2017offline,cao2020multilingual} should we put them back? at least one or two -ds \bm{we've cited all of then at the begining, wouldn't it be too much to cite them again here?}}. 
Our contributions advance the understanding of multilingual language models and as such have the potential to support the development of better pretraining processes. 

\section{Analysis Techniques}

We study mBERT with a novel behavioral test that disentangles the task fine-tuning influence from the pretraining step (\S \ref{sec:controlled_exp}), and a structural analysis on the intermediate representations (\S \ref{sec:alignement_description}). Combining the results from these analyses allows us to locate the cross-lingual transfer \draftreplace{}{and gain insights into the mechanisms that enable it}.

\subsection{Locating Transfer with \reinit}
\label{sec:controlled_exp}

In order to disentangle the impact of the pretraining step from the fine-tuning, we propose a new behavioral 
technique: \reinit. 
First, we randomly initialize a set of parameters (e.g. all the parameters of a given layer) instead of using the parameters learned during the pretraining step. Then, we fine-tune the modified pretrained model and measure the downstream performance.\footnote{Note that we perform the same optimization procedure for the model with and w/o \reinit (optimal learning rate and batch size are chosen with grid-search)
.}

By replacing a given set of pretrained parameters and fine-tuning the model, \textit{all other factors being equal}, \reinit enables us to quantify the contribution of a given set of pretrained parameters on downstream performance and therefore to locate which pretrained parameters contribute to the cross-lingual transfer. 

If the \crosslang performance is significantly lower than \inlang performance, we conclude that these layers are more important to \crosslang performance than they are for \inlang performance. If the \crosslang score does not change, it indicates that \crosslang transfer does not rely on these layers.

This technique is reminiscent of the recent \textit{Amnesic Probing} method \cite{amnesic_probing}, that removes from the representation a specific feature, e.g. Part-of-Speech, and then measures the outcome on the downstream task. In contrast, \reinit allows to study a specific architecture component, instead of specific features.

\subsection{Hidden State Similarities across Languages}
\label{sec:alignement_description}

To strengthen the behavioral evidence \draftadd{brought by \reinit}, and provide finer analyses that \draftreplace{focuses -sing}{focus} on individual layers, we \draftremove{propose to }study how the textual representations differ between parallel sentences in different languages. We hypothesize that \draftreplace{a good}{an efficient} fine-tuned model should be able to represent similar sentences in the source and target languages similarly, even-though it was fine-tuned only on the source language.

To measure the similarities of the representation across languages, we use the Central Kernel Alignment metric (CKA), introduced by \citet{kornblith2019similarity}. 
We follow \citet{wu2019emerging} who use the CKA as a similarity metric   \draftreplace{}{to compare the representations of monolingual and bilingual pretrained models across languages. In our work, we use the CKA to study the representation difference between source and target languages in pretrained and fine-tuned multilingual models}. For every layer, we average  all contextualized tokens in a sentence to get a single vector.\footnote{After removing [CLS] and [SEP] special tokens.} 
Then we compute the similarity between target and source representations and compare it across layers in the pretrained and fine-tuned models.
\draftreplace{We refer to this metric as}{We call this metric} the \textit{\similarity}.

\section{Experimental Setup}
\label{sec:exp}
\paragraph{Tasks, Datasets and Evaluation}
We consider three tasks covering both syntactic and semantic aspects of language: Part-Of-Speech Tagging (POS), dependency parsing, and Named-Entity Recognition (NER). For POS tagging and parsing we use the Universal Dependency \citep{ud22} treebanks, and
for NER, we use the WikiANN dataset \citep{pan-etal-2017-cross}. 
We evaluate our systems with the standard metrics per task; 
word-level accuracy for POS tagging, F1 for NER and labeled attachment score (LAS) for parsing. All the reported scores are computed on the test \draftreplace{split}{set} of each dataset. 

We experiment with English, Russian and Arabic as source languages, and fourteen typologically diverse target languages, including Chinese, Czech, German and Hindi. The complete list can be found in the Appendix~\ref{sec:data_appendix}.

The results of a model that is fine-tuned and evaluated on the \textit{same} language are referred to as \textit{\inlang} and those evaluated on \textit{distinct} languages are referred to as \textit{\crosslang}.

\paragraph{Multilingual Model}
We focus on mBERT \citep{devlin-etal-2019-bert}, a 12-layer model trained 
on the concatenation of 104 monolingual Wikipedia corpora, including our languages of study.
\paragraph{Fine-Tuning}
We fine-tune the model for each task following the standard methodology of \citet{devlin-etal-2019-bert}.
The exact details for reproducing our results can be found in the Appendix. 
All reported scores are averaged on 5 runs with different random seeds.

\section{Results}

\subsection{Disentangling the Pretraining Effect}
\label{sec:bottom_remove_description}

\begin{table}[t!]
\footnotesize\centering
\scalebox{0.7}{ 
\begin{tabular}{lrrrrrrr}
\toprule
& \multicolumn{7}{c}{\underline{\textit{\reinit \textit{of layers}}}} \\
\textsc{Src-Trg}&\textsc{\standardFineTune} & $\Delta$ \textsc{1-2} & $\Delta$ \textsc{3-4} & $\Delta$ \textsc{5-6} & $\Delta$ \textsc{7-8} & $\Delta$ \textsc{9-10} &  $\Delta$ \textsc{11-12} \\
&\multicolumn{7}{c}{\textit{Parsing}} \\

\textsc{En - En} & 88.98 & 
\cellcolor{lightgreen}-0.96 & \cellcolor{lightgreen}-0.66 & \cellcolor{lightgreen}-0.93 & \cellcolor{lightgreen}-0.55 & \cellcolor{mediumspringgreen}0.04 & \cellcolor{lightgreen}-0.09 \\

\textsc{Ru - Ru} & 85.15 & \cellcolor{lightgreen}-0.82 & \cellcolor{lightgreen}-1.38 & \cellcolor{lightgreen}-1.51 & \cellcolor{lightgreen}-0.86 & \cellcolor{lightgreen}-0.29 & \cellcolor{mediumspringgreen}0.18 \\

\textsc{Ar - Ar} & 59.54 & \cellcolor{lightgreen}-0.78 & \cellcolor{pink}-2.14 & \cellcolor{lightgreen}-1.20 & \cellcolor{lightgreen}-0.67 & \cellcolor{lightgreen}-0.27 & \cellcolor{mediumspringgreen}0.08 \\

\hdashline
\textsc{En -  X} & 53.23 & \cellcolor{brightlavender} -15.77 & \cellcolor{brightlavender} -6.51 & \cellcolor{pink} -3.39 & \cellcolor{lightgreen} -1.47 & \cellcolor{mediumspringgreen} 0.29 & \cellcolor{mediumspringgreen} 1.00 \\

\textsc{Ru -  X} &  55.41 & \cellcolor{brightlavender} -7.69 & \cellcolor{pink} -3.71 & \cellcolor{pink} -3.13 & \cellcolor{lightgreen} -1.70 & \cellcolor{mediumspringgreen} 0.92 & \cellcolor{mediumspringgreen} 0.94 \\

\textsc{Ar -  X} &  27.97 & \cellcolor{pink} -4.91 & \cellcolor{pink} -3.17 & \cellcolor{lightgreen} -1.48 & \cellcolor{lightgreen} -1.68 & \cellcolor{lightgreen} -0.36 & \cellcolor{lightgreen} -0.14 \\
\midrule

&\multicolumn{7}{c}{\textit{POS}} \\
\textsc{En - En} & 96.51 & \cellcolor{lightgreen}-0.30 & \cellcolor{lightgreen}-0.25 & \cellcolor{lightgreen}-0.40 & \cellcolor{lightgreen}-0.00 & \cellcolor{mediumspringgreen}0.05 & \cellcolor{mediumspringgreen}0.02 \\

\textsc{Ru - Ru} & 96.90 & \cellcolor{lightgreen}-0.52 & \cellcolor{lightgreen}-0.55 & \cellcolor{lightgreen}-0.40 & \cellcolor{lightgreen}-0.07 & \cellcolor{mediumspringgreen}0.02 & \cellcolor{lightgreen}-0.03 \\
\textsc{Ar - Ar} & 79.28 & \cellcolor{lightgreen}-0.35 & \cellcolor{lightgreen}-0.49 & \cellcolor{lightgreen}-0.36 & \cellcolor{lightgreen}-0.19 & \cellcolor{lightgreen}-0.05 & \cellcolor{lightgreen}-0.00 \\

\hdashline
\textsc{En -  X} & 79.37 & \cellcolor{brightlavender} -8.94 & \cellcolor{pink} -2.49 & \cellcolor{lightgreen} -1.66 & \cellcolor{lightgreen} -0.88 & \cellcolor{mediumspringgreen} 0.20 & \cellcolor{lightgreen} -0.14 \\

\textsc{Ru -  X} & 79.25 & \cellcolor{brightlavender} -10.08 & \cellcolor{pink} -2.83 & \cellcolor{lightgreen} -1.65 & \cellcolor{pink} -2.74 & \cellcolor{mediumspringgreen} 0.01 & \cellcolor{lightgreen} -0.45 \\

\textsc{Ar -  X} &  64.81 & \cellcolor{brightlavender} -6.73 & \cellcolor{pink} -3.50 & \cellcolor{lightgreen} -1.63 & \cellcolor{lightgreen} -1.56 & \cellcolor{lightgreen} -0.73 & \cellcolor{lightgreen} -1.29 \\

\midrule
&\multicolumn{7}{c}{\textit{NER}} \\
\textsc{En - En} & 83.30 & \cellcolor{pink}-2.66 & \cellcolor{pink}-2.14 & \cellcolor{lightgreen}-1.43 & \cellcolor{lightgreen}-0.63 & \cellcolor{lightgreen}-0.23 & \cellcolor{lightgreen}-0.12 \\

\textsc{Ru - Ru} & 88.20 & \cellcolor{pink}-2.08 & \cellcolor{pink}-2.13 & \cellcolor{lightgreen}-1.52 & \cellcolor{lightgreen}-0.64 & \cellcolor{lightgreen}-0.33 & \cellcolor{lightgreen}-0.13 \\

\textsc{Ar - Ar} & 87.97 & \cellcolor{pink}-2.37 & \cellcolor{pink}-2.11 & \cellcolor{lightgreen}-0.96 & \cellcolor{lightgreen}-0.39 & \cellcolor{lightgreen}-0.15 & \cellcolor{mediumspringgreen}0.21 \\

\hdashline
\textsc{En - X} &  64.17 & \cellcolor{brightlavender} -8.28 & \cellcolor{brightlavender} -5.09 & \cellcolor{pink} -3.07 & \cellcolor{lightgreen} -0.79 & \cellcolor{lightgreen} -0.47 & \cellcolor{lightgreen} -0.13 \\

\textsc{Ru - X} &  62.13 & \cellcolor{brightlavender} -15.85 & \cellcolor{brightlavender} -9.36 & \cellcolor{brightlavender} -5.50 & \cellcolor{pink} -2.44 & \cellcolor{lightgreen} -1.16 & \cellcolor{lightgreen} -0.06 \\

\textsc{Ar - X} & 65.59 & \cellcolor{brightlavender} -16.10 & \cellcolor{brightlavender} -8.42 & \cellcolor{pink} -3.73 & \cellcolor{lightgreen} -1.40 & \cellcolor{lightgreen} -0.25 & \cellcolor{mediumspringgreen} 0.67 \\
\bottomrule
\end{tabular}
}
\caption{Relative Zero shot Cross-Lingual performance of \mbert with \reinit (\S\ref{sec:controlled_exp}) on pairs of consecutive layers compared to \mbert without any random-initialization (\standardFineTune). In \textsc{Src-Trg}, \textsc{Src} indicates the source language on which we fine-tune \mbert, and \textsc{Trg} the target language on which we evaluate it. 
\textsc{Src-X} is the average across all 17 target language with X $\neq$ \textsc{Src}. Detailed results per target language are reported in tables \ref{tab:delta_pud_reinit_parsing}, \ref{tab:delta_pud_reinit_pos} and \ref{tab:delta_pud_reinit_ner} \draftadd{in the Appendix}.
Coloring is computed based on how \mbert with \reinit performs compared to the \standardFineTune model. 
\colorbox{mediumspringgreen}{$\geq$ \textsc{\standardFineTune}}
\colorbox{lightgreen}{$<$  \textsc{\standardFineTune}}
\colorbox{pink}{$\leq$ -2 points}
\colorbox{brightlavender}{$\leq$ -5 points }
}
\label{tab:delta_standard_re_init_x2}
\end{table}

For each experiment, we measure the impact of randomly-initializing specific layers as the difference between the model performance without any random-initialization (\standardFineTune) and with random-initialization (\reinit). Results for two consecutive layers are shown in Table~\ref{tab:delta_standard_re_init_x2}. The rest of the results, which exhibit similar trends, can be found in the Appendix (Table \ref{tab:re_init_full_ablation}).

For all tasks, we observe sharp drops in the \crosslang performance at the lower layers of the model but only moderate drops in the \inlang performance. For instance, the parsing experiment with English as the source language, results in a performance drop on English of only 0.96 points (\textsc{En-En}), when randomly-initializing layers 1 and 2. However, it leads to an average drop of 15.77 points on other languages (\textsc{En-X}).

Furthermore, we show that applying \reinit to the upper layers does not harm \inlang and \crosslang performances (e.g. when training on parsing for English, the performance slightly decreases by 0.09 points in the \inlang while it increases by 1.00 in the \crosslang case). This suggests that the upper layers are \textit{task-specific} and \textit{language-agnostic}, since re-initializing them have minimal change on performance. We conclude that mBERT's upper layers do not contribute to \crosslang transfer.


\paragraph{Does the Target Domain Matter?}
\label{sec:domain_analysis}

In order to test whether this behavior is specific to the \crosslang setting and is not general to \outdist (OOD) transfer, we repeat the same \reinit experiment by evaluating on \inlang setting while varying the evaluated domain.\footnote{Although other factors might play a part in out-of-distribution, we suspect that domains plays a crucial part in transfer. Moreover, it was shown that BERT encodes out-of-the-box domain information \cite{aharoni-goldberg-2020-unsupervised}} If the drop is similar to \crosslang performance, it means that lower layers are important for \outdist transfer in general. Otherwise, it would confirm that these layers play a specific role for \crosslang transfer.

\begin{table}[ht!]
\footnotesize\centering
\scalebox{0.69}{ 
\begin{tabular}{lrrrrrrrr}
    \toprule
       &&  \multicolumn{7}{c}{\underline{\textit{\reinit \textit{of layers}}}} \\
     \textsc{Src - Trg}&\textsc{\standardFineTune}  & $\Delta$\textsc{0-1}&$\Delta$\textsc{2-3} & $\Delta$\textsc{4-5} & $\Delta$\textsc{6-7} & $\Delta$\textsc{8-9} &  $\Delta$\textsc{10-11} \\
    \multicolumn{1}{c}{\textit{Domain Analyses}} &&\multicolumn{7}{c}{\textit{Parsing}}\\
     
 
 \textsc{En - En } & 90.40 & \cellcolor{lightgreen}-1.41 & \cellcolor{pink}-2.33 & \cellcolor{lightgreen}-1.57 & \cellcolor{lightgreen}-1.43 & \cellcolor{lightgreen}-0.60 & \cellcolor{lightgreen}-0.46 \\ 
\textsc{En - En Lit.} & 77.91 & \cellcolor{lightgreen}-0.91 & \cellcolor{lightgreen}-1.38 & \cellcolor{lightgreen}-1.85 & \cellcolor{lightgreen}-0.83 & \cellcolor{lightgreen}-0.23 & \cellcolor{lightgreen}-0.17 \\ 
\textsc{En - En Web} & 75.77 & \cellcolor{pink}-2.14 & \cellcolor{pink}-2.42 & \cellcolor{pink}-2.54 & \cellcolor{lightgreen}-1.42 & \cellcolor{lightgreen}-0.71 & \cellcolor{lightgreen}-0.69 \\ 
\textsc{En - En UGC} & 45.90 & \cellcolor{lightgreen}-1.97 & \cellcolor{pink}-2.75 & \cellcolor{pink}-2.10 & \cellcolor{lightgreen}-1.04 & \cellcolor{lightgreen}-0.39 & \cellcolor{lightgreen}-0.25 \\ 
\hdashline
\multicolumn{1}{c}{\textit{Cross-Language}}&\\
\textsc{En - Fr tran.} & 83.25 & \cellcolor{brightlavender}-5.82 & \cellcolor{pink}-2.69 & \cellcolor{pink}-2.42 & \cellcolor{lightgreen}-0.44 & \cellcolor{mediumspringgreen}0.25 & \cellcolor{mediumspringgreen}0.94 \\ 
\textsc{En - Fr Wiki} & 71.29 & \cellcolor{brightlavender}-7.86 & \cellcolor{pink}-4.33 & \cellcolor{pink}-4.64 & \cellcolor{lightgreen}-0.92 & \cellcolor{lightgreen}-0.11 & \cellcolor{mediumspringgreen}0.33 \\ 
\midrule
     \multicolumn{1}{c}{\textit{Domain Analyses}}&&\multicolumn{7}{c}{\textit{POS}}\\
\textsc{En - En} & 96.83 & \cellcolor{lightgreen}-1.35 & \cellcolor{lightgreen}-0.98 & \cellcolor{lightgreen}-0.70 & \cellcolor{lightgreen}-0.40 & \cellcolor{lightgreen}-0.28 & \cellcolor{lightgreen}-0.24 \\ 
\textsc{En - En Lit.} & 93.09 & \cellcolor{lightgreen}-0.58 & \cellcolor{lightgreen}-0.65 & \cellcolor{lightgreen}-0.28 & \cellcolor{lightgreen}-0.04 & \cellcolor{lightgreen}-0.06 & \cellcolor{mediumspringgreen}0.12 \\ 
\textsc{En - En Web} & 89.67 & \cellcolor{lightgreen}-1.07 & \cellcolor{lightgreen}-1.21 & \cellcolor{lightgreen}-0.41 & \cellcolor{lightgreen}-0.10 & \cellcolor{mediumspringgreen}0.03 & \cellcolor{mediumspringgreen}0.21 \\ 
\textsc{En - En UGC} & 68.93 & \cellcolor{pink}-2.38 & \cellcolor{lightgreen}-1.07 & \cellcolor{lightgreen}-0.14 & \cellcolor{mediumspringgreen}0.54 & \cellcolor{lightgreen}-0.04 & \cellcolor{mediumspringgreen}0.63 \\ 
\hdashline
\multicolumn{1}{c}{\textit{Cross-Language}}&\\
\textsc{En - Fr Tran.} & 93.43 & \cellcolor{pink}-3.59 & \cellcolor{lightgreen}-0.88 & \cellcolor{lightgreen}-1.31 & \cellcolor{lightgreen}-0.56 & \cellcolor{mediumspringgreen}0.46 & \cellcolor{mediumspringgreen}0.25 \\ 
\textsc{En - Fr.} & 91.13 & \cellcolor{brightlavender}-5.10 & \cellcolor{lightgreen}-0.93 & \cellcolor{lightgreen}-1.16 & \cellcolor{lightgreen}-0.74 & \cellcolor{mediumspringgreen}0.15 & \cellcolor{lightgreen}-0.07 \\ 
\midrule
\multicolumn{1}{c}{\textit{Domain Analyses}}&&\multicolumn{7}{c}{\textit{NER}}\\
\textsc{En - En} & 83.22 & \cellcolor{pink}-2.45 & \cellcolor{pink}-2.15 & \cellcolor{lightgreen}-1.28 & \cellcolor{lightgreen}-0.49 & \cellcolor{lightgreen}-0.15 & \cellcolor{lightgreen}-0.06 \\ 
\textsc{En - News} & 51.72 & \cellcolor{lightgreen}-1.32 & \cellcolor{lightgreen}-1.05 & \cellcolor{lightgreen}-0.80 & \cellcolor{lightgreen}-0.14 & \cellcolor{lightgreen}-0.31 & \cellcolor{lightgreen}-0.33 \\ 
 \hdashline
 \multicolumn{1}{c}{\textit{Cross-Language}}&\\
 \textsc{En - Fr} &76.16 & \cellcolor{brightlavender}-5.14 & \cellcolor{pink}-2.82 & \cellcolor{lightgreen}-1.97 & \cellcolor{lightgreen}-0.33 & \cellcolor{mediumspringgreen}0.52 & \cellcolor{mediumspringgreen}0.34 \\


\bottomrule

\end{tabular}
}
\caption{
Relative Zero shot Cross-Lingual performance of \mbert with \reinit (\S \ref{sec:controlled_exp}) on pairs of consecutive layers compared to \mbert without any random-initialization (\standardFineTune). We present experiments with English as the source language and evaluate across various target domains in English in comparison with the cross-lingual setting when we evaluate on French. \\
\textsc{EN-Lit.} refers to the Literature Domain. \textsc{UGC} refers to User-Generated Content. \textsc{FR-Tran.} refers to sentences translated from the English \textit{In-Domain} test set, hence reducing the domain-gap to its minimum. \\
\colorbox{mediumspringgreen}{$\geq$ \textsc{\standardFineTune}}
\colorbox{lightgreen}{$<$  \textsc{\standardFineTune}}
\colorbox{pink}{$\leq$ -2 points}
\colorbox{brightlavender}{$\leq$ -5 points }
}
\label{tab:re_init_x2_domain_delta_small}
\end{table}

We report the results in Table \ref{tab:re_init_x2_domain_delta_small}.
\draftreplace{In the \crosslang setting, applying \reinit to the two first layers of the model leads to a large drop in performance compared to the standard fine-tuned model (\standardFineTune):. This contrasts with the moderate drop when we evaluate on several domains in English.}{For all analyzed domains (Web, News, Literature, etc.) applying \reinit to the two first layers of the models leads to very moderate drops (e.g. -0.91 when the target domain is English Literature for parsing), while it leads to large drops when the evaluation is done on a distinct language (e.g. -5.82 when evaluated on French).} \draftnote{\hg{start with explicitly saying what you get for domains, then contrast with what we have seen for cross lingual}}
The trends are similar for all the domains and tasks we tested on. We conclude that the pretrained parameters at the lower layers are consistently more critical for \crosslang transfer than for \inlang transfer, and cannot be explained by the possibly different domain of the evaluated datasets.\draftnote{\hg{cool, important experiment!}}


\subsection{Cross-Lingual Similarity in mBERT}
\label{sec:alignement}

\begin{figure}[t!]
\centering
\includegraphics[width=.8\columnwidth]{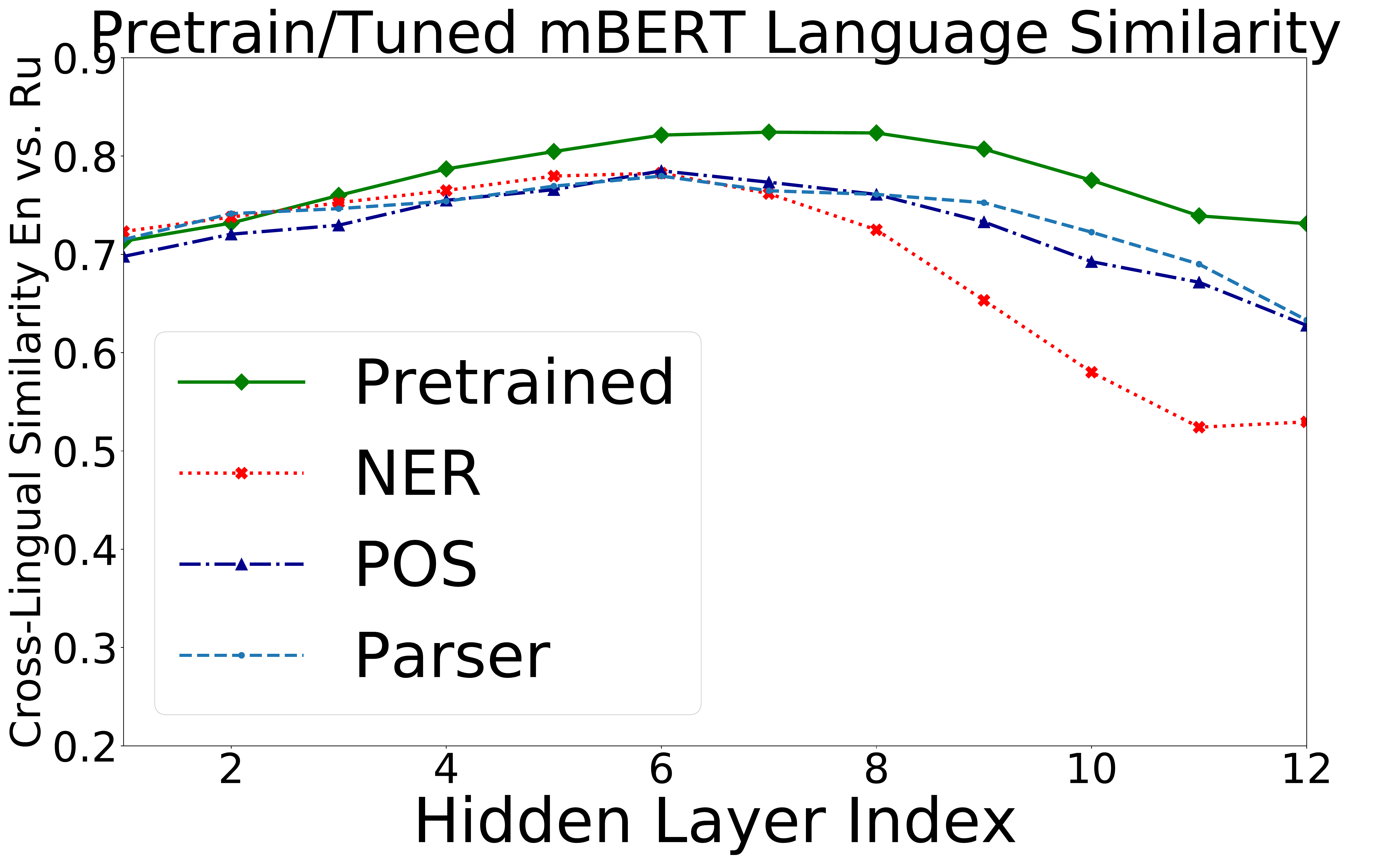}
\caption{Cross-Lingual similarity (CKA) between representations of pretrained and fine-tuned models on POS, NER and Parsing between English and Russian.}
\label{fig:layer_influence_unmasked}
\end{figure}

The results from the previous sections suggest that the lower layers of the model are responsible for the cross lingual transfer, whereas the upper layers are language-agnostic. In this section, we assess the transfer 
by directly analyzing the intermediate representations and measuring the similarities of the hidden state representations between source and target languages. 
We compute the CKA metric (cf. \S\ref{sec:alignement_description}) between the source and the target representations for pretrained and fine-tuned models using parallel sentences from the PUD dataset \citep{zeman-etal-2017-conll}. 
In Figure~\ref{fig:layer_influence_unmasked}, we present the similarities between Russian and English with mBERT pretrained and fine-tuned on the three tasks.\footnote{We report the comparisons for 5 other languages in Figure~\ref{fig:pretrained_tuned_models} in the Appendix.} 

The \similarity between the representations constantly increases up to layer 5 for all the three tasks (reaching 78.1\%, 78.1\% and 78.2\%  for parsing, POS tagging and NER respectively).
From these layers forward, the similarity decreases.
We observe the same trends across all languages (cf. Figure \ref{fig:pretrained_tuned_models}). This demonstrates that the fine-tuned model creates similar representations regardless of the language and task, and hints on an alignment that occurs in the lower part of the model. Interestingly, the same trend is also observed in the pretrained model, suggesting that the fine-tuning step preserves the multilingual alignment.

\begin{figure}[t!]
\centering
\includegraphics[width=.8\columnwidth]{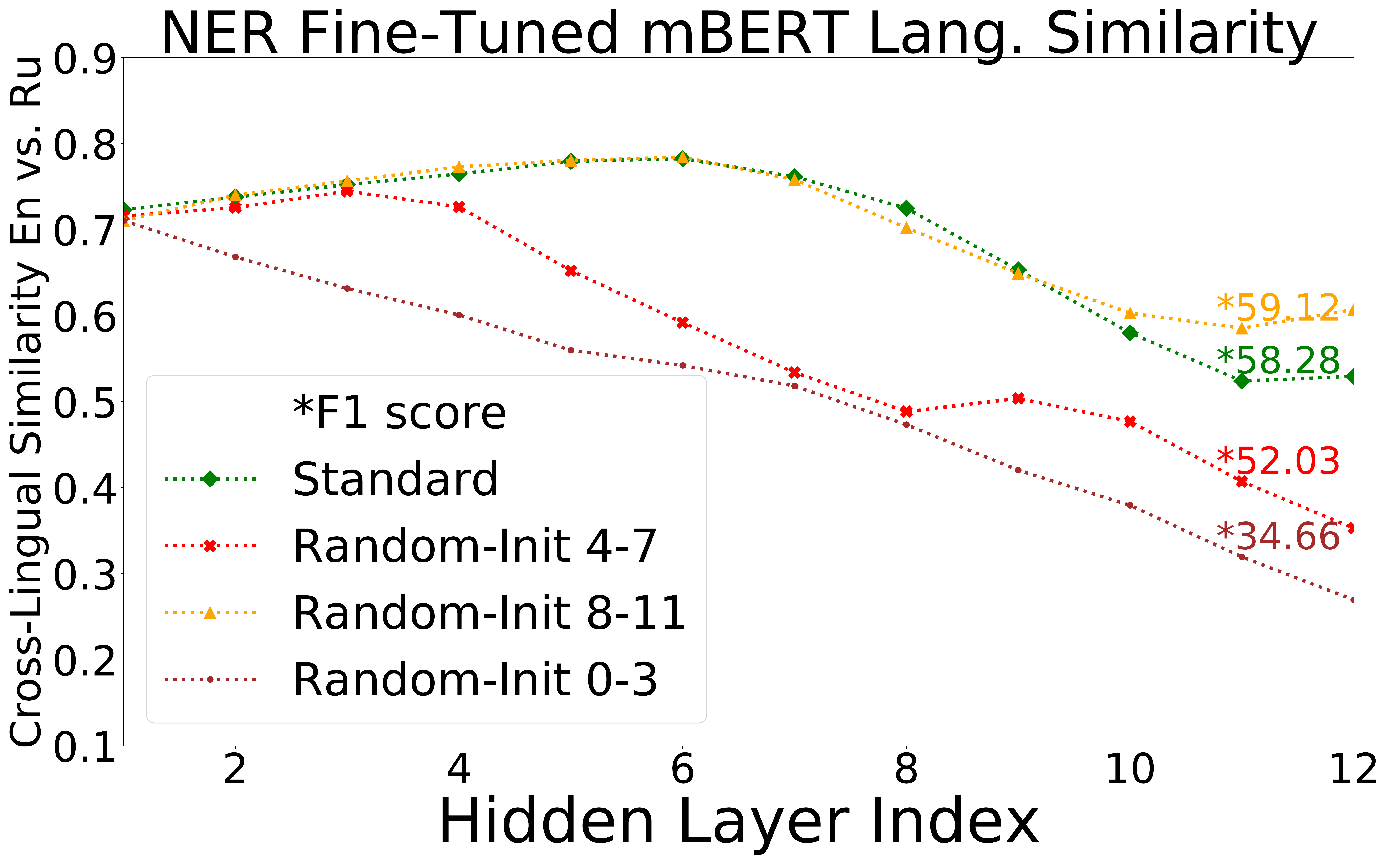}
\caption{Cross-Lingual similarity (CKA) of the representations of  a fine-tuned model on NER with and w/o \reinit between English (source) and Russian (target). The higher the score the greater the similarity.}
\label{fig:layer_re_init_NER}
\end{figure}


These results do not match the findings of \citet{singh2019bert}, who found no language alignment across layers, although they inspected Natural Language Inference, a more ``high-level task'' \cite{dagan2005pascal,bowman-etal-2015-large}.
We leave the inspection of this mismatch to future work.


\subsection{Better Alignment Leads to Better Cross-Lingual Transfer
}
\label{sec:downstream_perf_and_inner_sim}

In the previous section we showed that fine-tuned models align the representations between parallel sentences, across languages. Moreover, we demonstrated that the lower part of the model is critical for \crosslang transfer but hardly impacts the \inlang performance. In this section, we show that the alignment measured plays a critical role in cross-lingual transfer. 

As seen in Figure~\ref{fig:layer_re_init_NER} in the case of English to Russian (and in \draftnote{works in mode 'final' only} Figures~\ref{fig:plot_parsing_random_init}-\ref{fig:plot_ner_random_init} in the Appendix for other languages), when we randomly-initialize the lower part of the model, \draftremove{the alignment is low}{there is no alignment}: the similarity between the source and target languages decreases. We observe the same trend for all other languages and tasks and report it in the Appendix in Figures~\ref{fig:plot_parsing_random_init}-\ref{fig:plot_ner_random_init}. This result matches the drop in cross-lingual performance that occurs when we apply \reinit to the lower part of the model while impacting moderately \inlang performance. 

For a more systematic view of the link between the cross-lingual similarities and the \crosslang transfer, we measure the Spearman correlation between the \gap (i.e the difference between the \inlang perfromance and the \crosslang performance) \citep{hu2020xtreme} and the \similarity averaged over all the layers. 
We measure it with the \similarity computed on the pretrained and fine-tuned models (without random-initialization) on all the languages. 
We find that the \similarity correlates significantly with the \gap for all three tasks, both on the fine-tuned and pretrained models.\draftnote{Shouldn't we include that table from the appendix here ? \bm{we have no space, I think it is fine this way}} The spearman correlation  for the fine-tuned models are 0.76, 0.75 and 0.47 for parsing, POS and NER, respectively.\footnote{Correlations for both the pretrained and the fine-tuned models are reported in the Appendix Table~\ref{tab:correlation_table}.} In summary, our results show that the cross-lingual alignment is highly correlated with the cross-lingual transfer.

\section{Discussion}

Understanding the behavior of pretrained language models is currently a fundamental challenge in NLP. A popular approach consists of probing the intermediate representations with external classifiers \citep{AlainICLR,adiProbing,conneau-etal-2018-cram} to measure if a specific layer captures a given property. Using this technique, \citet{tenney-etal-2019-bert} showed that BERT encodes linguistic properties in the same order as the ``classical NLP pipeline''.  
However, probing techniques only indirectly explain the behavior of a model and do not explain the relationship between the information captured in the representations and its effect on the task \citep{amnesic_probing}. 
Moreover, recent works have questioned the usage of probing as an interpretation tool \cite{hewitt2019designing,ravichander2020probing}.
This motivates our approach to combine a structural analysis based on representation similarity with behavioral analysis. In this regard, our findings extend 
recent work from \citet{merchant-etal-2020-happens} in the multilingual setting, who show that fine-tuning impacts mainly the upper layers of the model and preserves the linguistic features learned during pretraining. In our case, we show that the lower layers are in charge of aligning representations across languages and that this cross-lingual alignment learned during pretraining is preserved after fine-tuning.

\section{Conclusion}

The remarkable performance of multilingual languages models in zero-shot cross-lingual transfer is still not well understood. 
In this work, we combine a \textit{structural} analysis of the similarities between hidden representation across languages with a novel \textit{behavioral} analysis that randomly-initialize the models' parameters to understand it.
\draftreplace{By combining the extensive experiments with these two techniques \hg{I would rephrase the prefix}}{By combining those experiments} on 17 languages and 3 tasks, we show that mBERT is constructed from: (1)~a multilingual encoder in the lower layers,  which aligns hidden representations across languages and is critical for \crosslang transfer, and (2)~a task-specific, language-agnostic predictor that has little effect to \crosslang transfer, in the upper layers.
Additionally, we demonstrate that hidden cross-lingual similarity strongly correlates with downstream cross-lingual performance suggesting that this alignment is at the root of these cross-lingual transfer abilities. This shows that mBERT reproduces the standard cross-lingual pipeline described by \citet{ruder2019survey} without any explicit supervision signal for it. Practically speaking, our findings provide a concrete tool to measure cross-lingual representation similarity that could be used to design better multilingual pretraining processes.

\section*{Acknowledgments}
We thank Hila Gonen, Shauli Ravfogel and Ganesh Jawahar for their careful review and insightful comments.
We also thank the anonymous reviewers for their valuable suggestions.
This work was partly funded by two French National funded projects granted to Inria and other partners by the Agence Nationale de la Recherche, namely projects PARSITI (ANR-16-CE33-0021) and SoSweet (ANR-15-CE38-0011), as well as by the third author's chair in the PRAIRIE institute funded by the French national agency ANR as part of the ``Investissements d’avenir'' programme under the reference \mbox{ANR-19-P3IA-0001}.
Yanai Elazar is grateful to be partially supported by the PBC fellowship for outstanding Phd candidates in Data Science.

\newpage

\bibliography{emnlp2020}

\begin{thebibliography}{37}
\expandafter\ifx\csname natexlab\endcsname\relax\def\natexlab#1{#1}\fi

\bibitem[{Adi et~al.(2017)Adi, Kermany, Belinkov, Lavi, and
  Goldberg}]{adiProbing}
Yossi Adi, Einat Kermany, Yonatan Belinkov, Ofer Lavi, and Yoav Goldberg. 2017.
\newblock \href {https://openreview.net/forum?id=BJh6Ztuxl} {Fine-grained
  analysis of sentence embeddings using auxiliary prediction tasks}.
\newblock In \emph{5th International Conference on Learning Representations,
  {ICLR} 2017, Toulon, France, April 24-26, 2017, Conference Track
  Proceedings}. OpenReview.net.

\bibitem[{Aharoni and Goldberg(2020)}]{aharoni-goldberg-2020-unsupervised}
Roee Aharoni and Yoav Goldberg. 2020.
\newblock \href {https://doi.org/10.18653/v1/2020.acl-main.692} {Unsupervised
  domain clusters in pretrained language models}.
\newblock In \emph{Proceedings of the 58th Annual Meeting of the Association
  for Computational Linguistics}, pages 7747--7763, Online. Association for
  Computational Linguistics.

\bibitem[{Alain and Bengio(2017)}]{AlainICLR}
Guillaume Alain and Yoshua Bengio. 2017.
\newblock \href {https://openreview.net/forum?id=HJ4-rAVtl} {Understanding
  intermediate layers using linear classifier probes}.
\newblock In \emph{5th International Conference on Learning Representations,
  {ICLR} 2017, Toulon, France, April 24-26, 2017, Workshop Track Proceedings}.
  OpenReview.net.

\bibitem[{Aone and McKee(1993)}]{aone1993language}
Chinatsu Aone and Douglas McKee. 1993.
\newblock A language-independent anaphora resolution system for understanding
  multilingual texts.
\newblock In \emph{31st Annual Meeting of the Association for Computational
  Linguistics}, pages 156--163.

\bibitem[{Belinkov et~al.(2020)Belinkov, Gehrmann, and
  Pavlick}]{belinkov2020interpretability}
Yonatan Belinkov, Sebastian Gehrmann, and Ellie Pavlick. 2020.
\newblock Interpretability and analysis in neural nlp.
\newblock In \emph{Proceedings of the 58th Annual Meeting of the Association
  for Computational Linguistics: Tutorial Abstracts}, pages 1--5.

\bibitem[{Bowman et~al.(2015)Bowman, Angeli, Potts, and
  Manning}]{bowman-etal-2015-large}
Samuel~R. Bowman, Gabor Angeli, Christopher Potts, and Christopher~D. Manning.
  2015.
\newblock \href {https://doi.org/10.18653/v1/D15-1075} {A large annotated
  corpus for learning natural language inference}.
\newblock In \emph{Proceedings of the 2015 Conference on Empirical Methods in
  Natural Language Processing}, pages 632--642, Lisbon, Portugal. Association
  for Computational Linguistics.

\bibitem[{Chi et~al.(2020)Chi, Hewitt, and Manning}]{chi-etal-2020-finding}
Ethan~A. Chi, John Hewitt, and Christopher~D. Manning. 2020.
\newblock \href {https://doi.org/10.18653/v1/2020.acl-main.493} {Finding
  universal grammatical relations in multilingual {BERT}}.
\newblock In \emph{Proceedings of the 58th Annual Meeting of the Association
  for Computational Linguistics}, pages 5564--5577, Online. Association for
  Computational Linguistics.

\bibitem[{Conneau et~al.(2018)Conneau, Kruszewski, Lample, Barrault, and
  Baroni}]{conneau-etal-2018-cram}
Alexis Conneau, German Kruszewski, Guillaume Lample, Lo{\"\i}c Barrault, and
  Marco Baroni. 2018.
\newblock \href {https://doi.org/10.18653/v1/P18-1198} {What you can cram into
  a single {\$}{\&}!{\#}* vector: Probing sentence embeddings for linguistic
  properties}.
\newblock In \emph{Proceedings of the 56th Annual Meeting of the Association
  for Computational Linguistics (Volume 1: Long Papers)}, pages 2126--2136,
  Melbourne, Australia. Association for Computational Linguistics.

\bibitem[{Conneau and Lample(2019)}]{conneau2019cross}
Alexis Conneau and Guillaume Lample. 2019.
\newblock Cross-lingual language model pretraining.
\newblock In \emph{Advances in Neural Information Processing Systems}, pages
  7057--7067.

\bibitem[{Conneau et~al.(2020)Conneau, Wu, Li, Zettlemoyer, and
  Stoyanov}]{wu2019emerging}
Alexis Conneau, Shijie Wu, Haoran Li, Luke Zettlemoyer, and Veselin Stoyanov.
  2020.
\newblock \href {https://doi.org/10.18653/v1/2020.acl-main.536} {Emerging
  cross-lingual structure in pretrained language models}.
\newblock In \emph{Proceedings of the 58th Annual Meeting of the Association
  for Computational Linguistics}, pages 6022--6034, Online. Association for
  Computational Linguistics.

\bibitem[{Dagan et~al.(2005)Dagan, Glickman, and Magnini}]{dagan2005pascal}
Ido Dagan, Oren Glickman, and Bernardo Magnini. 2005.
\newblock The pascal recognising textual entailment challenge.
\newblock In \emph{Machine Learning Challenges Workshop}, pages 177--190.
  Springer.

\bibitem[{Devlin et~al.(2019)Devlin, Chang, Lee, and
  Toutanova}]{devlin-etal-2019-bert}
Jacob Devlin, Ming-Wei Chang, Kenton Lee, and Kristina Toutanova. 2019.
\newblock \href {https://doi.org/10.18653/v1/N19-1423} {{BERT}: Pre-training of
  deep bidirectional transformers for language understanding}.
\newblock In \emph{Proceedings of the 2019 Conference of the North {A}merican
  Chapter of the Association for Computational Linguistics: Human Language
  Technologies, Volume 1 (Long and Short Papers)}, pages 4171--4186,
  Minneapolis, Minnesota. Association for Computational Linguistics.

\bibitem[{Elazar et~al.(2020)Elazar, Ravfogel, Jacovi, and
  Goldberg}]{amnesic_probing}
Yanai Elazar, Shauli Ravfogel, Alon Jacovi, and Y.~Goldberg. 2020.
\newblock Amnesic probing: Behavioral explanation with amnesic counterfactuals.
\newblock \emph{arXiv: Computation and Language}.

\bibitem[{Gonen et~al.(2020)Gonen, Ravfogel, Elazar, and Goldberg}]{gonen2020s}
Hila Gonen, Shauli Ravfogel, Yanai Elazar, and Yoav Goldberg. 2020.
\newblock It’s not greek to mbert: Inducing word-level translations from
  multilingual bert.
\newblock In \emph{Proceedings of the Third BlackboxNLP Workshop on Analyzing
  and Interpreting Neural Networks for NLP}, pages 45--56.

\bibitem[{van~der Goot and van Noord(2018)}]{van2018modeling}
Rob van~der Goot and Gertjan van Noord. 2018.
\newblock Modeling input uncertainty in neural network dependency parsing.
\newblock In \emph{Proceedings of the 2018 Conference on Empirical Methods in
  Natural Language Processing}, pages 4984--4991.

\bibitem[{Hewitt and Liang(2019)}]{hewitt2019designing}
John Hewitt and Percy Liang. 2019.
\newblock Designing and interpreting probes with control tasks.
\newblock In \emph{Proceedings of the 2019 Conference on Empirical Methods in
  Natural Language Processing and the 9th International Joint Conference on
  Natural Language Processing (EMNLP-IJCNLP)}, pages 2733--2743.

\bibitem[{Hu et~al.(2020)Hu, Ruder, Siddhant, Neubig, Firat, and
  Johnson}]{hu2020xtreme}
Junjie Hu, Sebastian Ruder, Aditya Siddhant, Graham Neubig, Orhan Firat, and
  Melvin Johnson. 2020.
\newblock \href {http://proceedings.mlr.press/v119/hu20b.html} {{XTREME}: A
  massively multilingual multi-task benchmark for evaluating cross-lingual
  generalisation}.
\newblock In \emph{Proceedings of the 37th International Conference on Machine
  Learning}, volume 119 of \emph{Proceedings of Machine Learning Research},
  pages 4411--4421. PMLR.

\bibitem[{Kingma and Ba(2015)}]{Kingma2015AdamAM}
Diederik~P. Kingma and Jimmy Ba. 2015.
\newblock Adam: A method for stochastic optimization.
\newblock \emph{CoRR}, abs/1412.6980.

\bibitem[{Kornblith et~al.(2019)Kornblith, Norouzi, Lee, and
  Hinton}]{kornblith2019similarity}
Simon Kornblith, Mohammad Norouzi, Honglak Lee, and Geoffrey Hinton. 2019.
\newblock Similarity of neural network representations revisited.
\newblock In \emph{International Conference on Machine Learning}, pages
  3519--3529.

\bibitem[{McDonald et~al.(2013)McDonald, Nivre, Quirmbach-Brundage, Goldberg,
  Das, Ganchev, Hall, Petrov, Zhang, T{\"a}ckstr{\"o}m, Bedini,
  Bertomeu~Castell{\'o}, and Lee}]{mcdonald-etal-2013-universal}
Ryan McDonald, Joakim Nivre, Yvonne Quirmbach-Brundage, Yoav Goldberg, Dipanjan
  Das, Kuzman Ganchev, Keith Hall, Slav Petrov, Hao Zhang, Oscar
  T{\"a}ckstr{\"o}m, Claudia Bedini, N{\'u}ria Bertomeu~Castell{\'o}, and
  Jungmee Lee. 2013.
\newblock \href {https://www.aclweb.org/anthology/P13-2017} {Universal
  dependency annotation for multilingual parsing}.
\newblock In \emph{Proceedings of the 51st Annual Meeting of the Association
  for Computational Linguistics (Volume 2: Short Papers)}, pages 92--97, Sofia,
  Bulgaria. Association for Computational Linguistics.

\bibitem[{Merchant et~al.(2020)Merchant, Rahimtoroghi, Pavlick, and
  Tenney}]{merchant-etal-2020-happens}
Amil Merchant, Elahe Rahimtoroghi, Ellie Pavlick, and Ian Tenney. 2020.
\newblock \href {https://doi.org/10.18653/v1/2020.blackboxnlp-1.4} {What
  happens to {BERT} embeddings during fine-tuning?}
\newblock In \emph{Proceedings of the Third BlackboxNLP Workshop on Analyzing
  and Interpreting Neural Networks for NLP}, pages 33--44, Online. Association
  for Computational Linguistics.

\bibitem[{Mikolov et~al.(2013)Mikolov, Le, and
  Sutskever}]{mikolov2013exploiting}
Tomas Mikolov, Quoc~V Le, and Ilya Sutskever. 2013.
\newblock Exploiting similarities among languages for machine translation.
\newblock \emph{arXiv preprint arXiv:1309.4168}.

\bibitem[{Nivre et~al.(2018)Nivre, Abrams, Agi{\'c}, Ahrenberg, Antonsen,
  Aranzabe, Arutie, Asahara, Ateyah, Attia, Atutxa, Augustinus, Badmaeva,
  Ballesteros, Banerjee, Bank, Barbu~Mititelu, Bauer, Bellato, Bengoetxea,
  Bhat, Biagetti, Bick, Blokland, Bobicev, B{\"o}rstell, Bosco, Bouma, Bowman,
  Boyd, Burchardt, Candito, Caron, Caron, Cebiro{\u g}lu~Eryi{\u g}it, Celano,
  Cetin, Chalub, Choi, Cho, Chun, Cinkov{\'a}, Collomb, {\c C}{\"o}ltekin,
  Connor, Courtin, Davidson, de~Marneffe, de~Paiva, Diaz~de Ilarraza,
  Dickerson, Dirix, Dobrovoljc, Dozat, Droganova, Dwivedi, Eli, Elkahky,
  Ephrem, Erjavec, Etienne, Farkas, Fernandez~Alcalde, Foster, Freitas,
  Gajdo{\v s}ov{\'a}, Galbraith, Garcia, G{\"a}rdenfors, Gerdes, Ginter,
  Goenaga, Gojenola, G{\"o}k{\i}rmak, Goldberg, G{\'o}mez~Guinovart,
  Gonz{\'a}les~Saavedra, Grioni, Gr{\=u}z{\={\i}}tis, Guillaume,
  Guillot-Barbance, Habash, Haji{\v c}, Haji{\v c}~jr., H{\`a}~M{\~y}, Han,
  Harris, Haug, Hladk{\'a}, Hlav{\'a}{\v c}ov{\'a}, Hociung, Hohle, Hwang, Ion,
  Irimia, Jel{\'{\i}}nek, Johannsen, J{\o}rgensen, Ka{\c s}{\i}kara, Kahane,
  Kanayama, Kanerva, Kayadelen, Kettnerov{\'a}, Kirchner, Kotsyba, Krek, Kwak,
  Laippala, Lambertino, Lando, Larasati, Lavrentiev, Lee, L{\^e}~H{\`{\^o}}ng,
  Lenci, Lertpradit, Leung, Li, Li, Li, Lim, Ljube{\v s}i{\'c}, Loginova,
  Lyashevskaya, Lynn, Macketanz, Makazhanov, Mandl, Manning, Manurung, M{\u
  a}r{\u a}nduc, Mare{\v c}ek, Marheinecke, Mart{\'{\i}}nez~Alonso, Martins,
  Ma{\v s}ek, Matsumoto, {McDonald}, Mendon{\c c}a, Miekka, Missil{\"a},
  Mititelu, Miyao, Montemagni, More, Moreno~Romero, Mori, Mortensen,
  Moskalevskyi, Muischnek, Murawaki, M{\"u}{\"u}risep, Nainwani,
  Navarro~Hor{\~n}iacek, Nedoluzhko, Ne{\v s}pore-B{\=e}rzkalne,
  Nguy{\~{\^e}}n~Th{\d i}, Nguy{\~{\^e}}n Th{\d i}~Minh, Nikolaev, Nitisaroj,
  Nurmi, Ojala, Ol{\'u}{\`o}kun, Omura, Osenova, {\"O}stling, {\O}vrelid,
  Partanen, Pascual, Passarotti, Patejuk, Peng, Perez, Perrier, Petrov,
  Piitulainen, Pitler, Plank, Poibeau, Popel, Pretkalni{\c n}a, Pr{\'e}vost,
  Prokopidis, Przepi{\'o}rkowski, Puolakainen, Pyysalo, R{\"a}{\"a}bis,
  Rademaker, Ramasamy, Rama, Ramisch, Ravishankar, Real, Reddy, Rehm,
  Rie{\ss}ler, Rinaldi, Rituma, Rocha, Romanenko, Rosa, Rovati, Roșca, Rudina,
  Sadde, Saleh, Samard{\v z}i{\'c}, Samson, Sanguinetti, Saul{\={\i}}te,
  Sawanakunanon, Schneider, Schuster, Seddah, Seeker, Seraji, Shen, Shimada,
  Shohibussirri, Sichinava, Silveira, Simi, Simionescu, Simk{\'o}, {\v
  S}imkov{\'a}, Simov, Smith, Soares-Bastos, Stella, Straka, Strnadov{\'a},
  Suhr, Sulubacak, Sz{\'a}nt{\'o}, Taji, Takahashi, Tanaka, Tellier, Trosterud,
  Trukhina, Tsarfaty, Tyers, Uematsu, Ure{\v s}ov{\'a}, Uria, Uszkoreit,
  Vajjala, van Niekerk, van Noord, Varga, Vincze, Wallin, Washington, Williams,
  Wir{\'e}n, Woldemariam, Wong, Yan, Yavrumyan, Yu, {\v Z}abokrtsk{\'y},
  Zeldes, Zeman, Zhang, and Zhu}]{ud22}
Joakim Nivre, Mitchell Abrams, {\v Z}eljko Agi{\'c}, Lars Ahrenberg, Lene
  Antonsen, Maria~Jesus Aranzabe, Gashaw Arutie, Masayuki Asahara, Luma Ateyah,
  Mohammed Attia, Aitziber Atutxa, Liesbeth Augustinus, Elena Badmaeva, Miguel
  Ballesteros, Esha Banerjee, Sebastian Bank, Verginica Barbu~Mititelu, John
  Bauer, Sandra Bellato, Kepa Bengoetxea, Riyaz~Ahmad Bhat, Erica Biagetti,
  Eckhard Bick, Rogier Blokland, Victoria Bobicev, Carl B{\"o}rstell, Cristina
  Bosco, Gosse Bouma, Sam Bowman, Adriane Boyd, Aljoscha Burchardt, Marie
  Candito, Bernard Caron, Gauthier Caron, G{\"u}l{\c s}en Cebiro{\u
  g}lu~Eryi{\u g}it, Giuseppe G.~A. Celano, Savas Cetin, Fabricio Chalub, Jinho
  Choi, Yongseok Cho, Jayeol Chun, Silvie Cinkov{\'a}, Aur{\'e}lie Collomb, {\c
  C}a{\u g}r{\i} {\c C}{\"o}ltekin, Miriam Connor, Marine Courtin, Elizabeth
  Davidson, Marie-Catherine de~Marneffe, Valeria de~Paiva, Arantza Diaz~de
  Ilarraza, Carly Dickerson, Peter Dirix, Kaja Dobrovoljc, Timothy Dozat, Kira
  Droganova, Puneet Dwivedi, Marhaba Eli, Ali Elkahky, Binyam Ephrem, Toma{\v
  z} Erjavec, Aline Etienne, Rich{\'a}rd Farkas, Hector Fernandez~Alcalde,
  Jennifer Foster, Cl{\'a}udia Freitas, Katar{\'{\i}}na Gajdo{\v s}ov{\'a},
  Daniel Galbraith, Marcos Garcia, Moa G{\"a}rdenfors, Kim Gerdes, Filip
  Ginter, Iakes Goenaga, Koldo Gojenola, Memduh G{\"o}k{\i}rmak, Yoav Goldberg,
  Xavier G{\'o}mez~Guinovart, Berta Gonz{\'a}les~Saavedra, Matias Grioni,
  Normunds Gr{\=u}z{\={\i}}tis, Bruno Guillaume, C{\'e}line Guillot-Barbance,
  Nizar Habash, Jan Haji{\v c}, Jan Haji{\v c}~jr., Linh H{\`a}~M{\~y}, Na-Rae
  Han, Kim Harris, Dag Haug, Barbora Hladk{\'a}, Jaroslava Hlav{\'a}{\v
  c}ov{\'a}, Florinel Hociung, Petter Hohle, Jena Hwang, Radu Ion, Elena
  Irimia, Tom{\'a}{\v s} Jel{\'{\i}}nek, Anders Johannsen, Fredrik
  J{\o}rgensen, H{\"u}ner Ka{\c s}{\i}kara, Sylvain Kahane, Hiroshi Kanayama,
  Jenna Kanerva, Tolga Kayadelen, V{\'a}clava Kettnerov{\'a}, Jesse Kirchner,
  Natalia Kotsyba, Simon Krek, Sookyoung Kwak, Veronika Laippala, Lorenzo
  Lambertino, Tatiana Lando, Septina~Dian Larasati, Alexei Lavrentiev, John
  Lee, Phương L{\^e}~H{\`{\^o}}ng, Alessandro Lenci, Saran Lertpradit, Herman
  Leung, Cheuk~Ying Li, Josie Li, Keying Li, {KyungTae} Lim, Nikola Ljube{\v
  s}i{\'c}, Olga Loginova, Olga Lyashevskaya, Teresa Lynn, Vivien Macketanz,
  Aibek Makazhanov, Michael Mandl, Christopher Manning, Ruli Manurung, C{\u
  a}t{\u a}lina M{\u a}r{\u a}nduc, David Mare{\v c}ek, Katrin Marheinecke,
  H{\'e}ctor Mart{\'{\i}}nez~Alonso, Andr{\'e} Martins, Jan Ma{\v s}ek, Yuji
  Matsumoto, Ryan {McDonald}, Gustavo Mendon{\c c}a, Niko Miekka, Anna
  Missil{\"a}, C{\u a}t{\u a}lin Mititelu, Yusuke Miyao, Simonetta Montemagni,
  Amir More, Laura Moreno~Romero, Shinsuke Mori, Bjartur Mortensen, Bohdan
  Moskalevskyi, Kadri Muischnek, Yugo Murawaki, Kaili M{\"u}{\"u}risep, Pinkey
  Nainwani, Juan~Ignacio Navarro~Hor{\~n}iacek, Anna Nedoluzhko, Gunta Ne{\v
  s}pore-B{\=e}rzkalne, Lương Nguy{\~{\^e}}n~Th{\d i}, Huy{\`{\^e}}n
  Nguy{\~{\^e}}n Th{\d i}~Minh, Vitaly Nikolaev, Rattima Nitisaroj, Hanna
  Nurmi, Stina Ojala, Ad{\'e}day{\d{\`o}} Ol{\'u}{\`o}kun, Mai Omura, Petya
  Osenova, Robert {\"O}stling, Lilja {\O}vrelid, Niko Partanen, Elena Pascual,
  Marco Passarotti, Agnieszka Patejuk, Siyao Peng, Cenel-Augusto Perez, Guy
  Perrier, Slav Petrov, Jussi Piitulainen, Emily Pitler, Barbara Plank, Thierry
  Poibeau, Martin Popel, Lauma Pretkalni{\c n}a, Sophie Pr{\'e}vost, Prokopis
  Prokopidis, Adam Przepi{\'o}rkowski, Tiina Puolakainen, Sampo Pyysalo,
  Andriela R{\"a}{\"a}bis, Alexandre Rademaker, Loganathan Ramasamy, Taraka
  Rama, Carlos Ramisch, Vinit Ravishankar, Livy Real, Siva Reddy, Georg Rehm,
  Michael Rie{\ss}ler, Larissa Rinaldi, Laura Rituma, Luisa Rocha, Mykhailo
  Romanenko, Rudolf Rosa, Davide Rovati, Valentin Roșca, Olga Rudina, Shoval
  Sadde, Shadi Saleh, Tanja Samard{\v z}i{\'c}, Stephanie Samson, Manuela
  Sanguinetti, Baiba Saul{\={\i}}te, Yanin Sawanakunanon, Nathan Schneider,
  Sebastian Schuster, Djam{\'e} Seddah, Wolfgang Seeker, Mojgan Seraji,
  Mo~Shen, Atsuko Shimada, Muh Shohibussirri, Dmitry Sichinava, Natalia
  Silveira, Maria Simi, Radu Simionescu, Katalin Simk{\'o}, M{\'a}ria {\v
  S}imkov{\'a}, Kiril Simov, Aaron Smith, Isabela Soares-Bastos, Antonio
  Stella, Milan Straka, Jana Strnadov{\'a}, Alane Suhr, Umut Sulubacak, Zsolt
  Sz{\'a}nt{\'o}, Dima Taji, Yuta Takahashi, Takaaki Tanaka, Isabelle Tellier,
  Trond Trosterud, Anna Trukhina, Reut Tsarfaty, Francis Tyers, Sumire Uematsu,
  Zde{\v n}ka Ure{\v s}ov{\'a}, Larraitz Uria, Hans Uszkoreit, Sowmya Vajjala,
  Daniel van Niekerk, Gertjan van Noord, Viktor Varga, Veronika Vincze, Lars
  Wallin, Jonathan~North Washington, Seyi Williams, Mats Wir{\'e}n, Tsegay
  Woldemariam, Tak-sum Wong, Chunxiao Yan, Marat~M. Yavrumyan, Zhuoran Yu,
  Zden{\v e}k {\v Z}abokrtsk{\'y}, Amir Zeldes, Daniel Zeman, Manying Zhang,
  and Hanzhi Zhu. 2018.
\newblock \href {http://hdl.handle.net/11234/1-2837} {Universal dependencies
  2.2}.
\newblock {LINDAT}/{CLARIN} digital library at the Institute of Formal and
  Applied Linguistics ({{\'U}FAL}), Faculty of Mathematics and Physics, Charles
  University.

\bibitem[{Pan et~al.(2017)Pan, Zhang, May, Nothman, Knight, and
  Ji}]{pan-etal-2017-cross}
Xiaoman Pan, Boliang Zhang, Jonathan May, Joel Nothman, Kevin Knight, and Heng
  Ji. 2017.
\newblock \href {https://doi.org/10.18653/v1/P17-1178} {Cross-lingual name
  tagging and linking for 282 languages}.
\newblock In \emph{Proceedings of the 55th Annual Meeting of the Association
  for Computational Linguistics (Volume 1: Long Papers)}, pages 1946--1958,
  Vancouver, Canada. Association for Computational Linguistics.

\bibitem[{Pires et~al.(2019)Pires, Schlinger, and
  Garrette}]{pires-etal-2019-multilingual}
Telmo Pires, Eva Schlinger, and Dan Garrette. 2019.
\newblock \href {https://doi.org/10.18653/v1/P19-1493} {How multilingual is
  multilingual {BERT}?}
\newblock In \emph{Proceedings of the 57th Annual Meeting of the Association
  for Computational Linguistics}, pages 4996--5001, Florence, Italy.
  Association for Computational Linguistics.

\bibitem[{Ravichander et~al.(2020)Ravichander, Belinkov, and
  Hovy}]{ravichander2020probing}
Abhilasha Ravichander, Yonatan Belinkov, and Eduard Hovy. 2020.
\newblock Probing the probing paradigm: Does probing accuracy entail task
  relevance?
\newblock \emph{arXiv preprint arXiv:2005.00719}.

\bibitem[{Ruder et~al.(2019)Ruder, Vuli{\'c}, and S{\o}gaard}]{ruder2019survey}
Sebastian Ruder, Ivan Vuli{\'c}, and Anders S{\o}gaard. 2019.
\newblock A survey of cross-lingual word embedding models.
\newblock \emph{Journal of Artificial Intelligence Research}, 65:569--631.

\bibitem[{Schultz and Waibel(2001)}]{schultz2001language}
Tanja Schultz and Alex Waibel. 2001.
\newblock Language-independent and language-adaptive acoustic modeling for
  speech recognition.
\newblock \emph{Speech Communication}, 35(1-2):31--51.

\bibitem[{Singh et~al.(2019)Singh, McCann, Socher, and Xiong}]{singh2019bert}
Jasdeep Singh, Bryan McCann, Richard Socher, and Caiming Xiong. 2019.
\newblock Bert is not an interlingua and the bias of tokenization.
\newblock In \emph{Proceedings of the 2nd Workshop on Deep Learning Approaches
  for Low-Resource NLP (DeepLo 2019)}, pages 47--55.

\bibitem[{Smith et~al.(2017)Smith, Turban, Hamblin, and
  Hammerla}]{smith2017offline}
Samuel~L. Smith, David H.~P. Turban, Steven Hamblin, and Nils~Y. Hammerla.
  2017.
\newblock \href {https://openreview.net/forum?id=r1Aab85gg} {Offline bilingual
  word vectors, orthogonal transformations and the inverted softmax}.
\newblock In \emph{5th International Conference on Learning Representations,
  {ICLR} 2017, Toulon, France, April 24-26, 2017, Conference Track
  Proceedings}. OpenReview.net.

\bibitem[{S{\o}gaard(2011)}]{sogaard2011data}
Anders S{\o}gaard. 2011.
\newblock Data point selection for cross-language adaptation of dependency
  parsers.
\newblock In \emph{Proceedings of the 49th Annual Meeting of the Association
  for Computational Linguistics: Human Language Technologies: short
  papers-Volume 2}, pages 682--686. Association for Computational Linguistics.

\bibitem[{Svizzera(2014)}]{Svizzera2014ConvertingTP}
Corso Svizzera. 2014.
\newblock Converting the parallel treebank partut in universal stanford
  dependencies.

\bibitem[{Tenney et~al.(2019)Tenney, Das, and Pavlick}]{tenney-etal-2019-bert}
Ian Tenney, Dipanjan Das, and Ellie Pavlick. 2019.
\newblock \href {https://doi.org/10.18653/v1/P19-1452} {{BERT} rediscovers the
  classical {NLP} pipeline}.
\newblock In \emph{Proceedings of the 57th Annual Meeting of the Association
  for Computational Linguistics}, pages 4593--4601, Florence, Italy.
  Association for Computational Linguistics.

\bibitem[{Wolf et~al.(2020)Wolf, Debut, Sanh, Chaumond, Delangue, Moi, Cistac,
  Rault, Louf, Funtowicz, Davison, Shleifer, von Platen, Ma, Jernite, Plu, Xu,
  Le~Scao, Gugger, Drame, Lhoest, and Rush}]{wolf-etal-2020-transformers}
Thomas Wolf, Lysandre Debut, Victor Sanh, Julien Chaumond, Clement Delangue,
  Anthony Moi, Pierric Cistac, Tim Rault, Remi Louf, Morgan Funtowicz, Joe
  Davison, Sam Shleifer, Patrick von Platen, Clara Ma, Yacine Jernite, Julien
  Plu, Canwen Xu, Teven Le~Scao, Sylvain Gugger, Mariama Drame, Quentin Lhoest,
  and Alexander Rush. 2020.
\newblock \href {https://doi.org/10.18653/v1/2020.emnlp-demos.6} {Transformers:
  State-of-the-art natural language processing}.
\newblock In \emph{Proceedings of the 2020 Conference on Empirical Methods in
  Natural Language Processing: System Demonstrations}, pages 38--45, Online.
  Association for Computational Linguistics.

\bibitem[{Wu and Dredze(2019)}]{wu2019beto}
Shijie Wu and Mark Dredze. 2019.
\newblock Beto, bentz, becas: The surprising cross-lingual effectiveness of
  bert.
\newblock In \emph{Proceedings of the 2019 Conference on Empirical Methods in
  Natural Language Processing and the 9th International Joint Conference on
  Natural Language Processing (EMNLP-IJCNLP)}, pages 833--844.

\bibitem[{Zeman et~al.(2017)Zeman, Popel, Straka, Haji{\v{c}}, Nivre, Ginter,
  Luotolahti, Pyysalo, Petrov, Potthast, Tyers, Badmaeva, Gokirmak, Nedoluzhko,
  Cinkov{\'a}, Haji{\v{c}}~jr., Hlav{\'a}{\v{c}}ov{\'a}, Kettnerov{\'a},
  Ure{\v{s}}ov{\'a}, Kanerva, Ojala, Missil{\"a}, Manning, Schuster, Reddy,
  Taji, Habash, Leung, de~Marneffe, Sanguinetti, Simi, Kanayama, de~Paiva,
  Droganova, Mart{\'\i}nez~Alonso, {\c{C}}{\"o}ltekin, Sulubacak, Uszkoreit,
  Macketanz, Burchardt, Harris, Marheinecke, Rehm, Kayadelen, Attia, Elkahky,
  Yu, Pitler, Lertpradit, Mandl, Kirchner, Alcalde, Strnadov{\'a}, Banerjee,
  Manurung, Stella, Shimada, Kwak, Mendon{\c{c}}a, Lando, Nitisaroj, and
  Li}]{zeman-etal-2017-conll}
Daniel Zeman, Martin Popel, Milan Straka, Jan Haji{\v{c}}, Joakim Nivre, Filip
  Ginter, Juhani Luotolahti, Sampo Pyysalo, Slav Petrov, Martin Potthast,
  Francis Tyers, Elena Badmaeva, Memduh Gokirmak, Anna Nedoluzhko, Silvie
  Cinkov{\'a}, Jan Haji{\v{c}}~jr., Jaroslava Hlav{\'a}{\v{c}}ov{\'a},
  V{\'a}clava Kettnerov{\'a}, Zde{\v{n}}ka Ure{\v{s}}ov{\'a}, Jenna Kanerva,
  Stina Ojala, Anna Missil{\"a}, Christopher~D. Manning, Sebastian Schuster,
  Siva Reddy, Dima Taji, Nizar Habash, Herman Leung, Marie-Catherine
  de~Marneffe, Manuela Sanguinetti, Maria Simi, Hiroshi Kanayama, Valeria
  de~Paiva, Kira Droganova, H{\'e}ctor Mart{\'\i}nez~Alonso,
  {\c{C}}a{\u{g}}r{\i} {\c{C}}{\"o}ltekin, Umut Sulubacak, Hans Uszkoreit,
  Vivien Macketanz, Aljoscha Burchardt, Kim Harris, Katrin Marheinecke, Georg
  Rehm, Tolga Kayadelen, Mohammed Attia, Ali Elkahky, Zhuoran Yu, Emily Pitler,
  Saran Lertpradit, Michael Mandl, Jesse Kirchner, Hector~Fernandez Alcalde,
  Jana Strnadov{\'a}, Esha Banerjee, Ruli Manurung, Antonio Stella, Atsuko
  Shimada, Sookyoung Kwak, Gustavo Mendon{\c{c}}a, Tatiana Lando, Rattima
  Nitisaroj, and Josie Li. 2017.
\newblock \href {https://doi.org/10.18653/v1/K17-3001} {{C}o{NLL} 2017 shared
  task: Multilingual parsing from raw text to universal dependencies}.
\newblock In \emph{Proceedings of the {C}o{NLL} 2017 Shared Task: Multilingual
  Parsing from Raw Text to Universal Dependencies}, pages 1--19, Vancouver,
  Canada. Association for Computational Linguistics.

\bibitem[{Zeman and Resnik(2008)}]{zeman2008cross}
Daniel Zeman and Philip Resnik. 2008.
\newblock Cross-language parser adaptation between related languages.
\newblock In \emph{Proceedings of the IJCNLP-08 Workshop on NLP for Less
  Privileged Languages}.

\end{thebibliography}
\bibliographystyle{acl_natbib}

\newpage
\appendix
\section{Appendices}
\label{sec:appendix}

\subsection{Reproducibility}
\label{sec:reproducibility}

\subsubsection{Optimization}
\label{sec:optim}

We fine-tune our models using the standard Adam optimizer \citep{Kingma2015AdamAM}. We warmup the learning rate on the first 10\%  steps and use linear decay in the rest of the training. 
Using the validation set of the source language, we find the best combination of hyper-parameters with a grid search on batch size among \{16, 32\} and learning rate initialization among \{1e-5, 2.5e-5, 5e-5\}
We select the model with the highest validation performance out of 15 epochs for parsing and out of 6 epochs for POS tagging and NER.

\paragraph{Hyperparameters}


In Table~\ref{tab:hyperparameters}, we report the best hyper-parameters set for each task, the bound of each hyperparameter, the estimated number of grid search trial for each task as well as the estimated run time. 

\subsubsection{Data}
\label{sec:data_appendix}

\paragraph{Data Sources}

We base our experiments on data originated from two sources. The Universal Dependency project \cite{mcdonald-etal-2013-universal} downloadable here \url{https://lindat.mff.cuni.cz/repository/xmlui/handle/11234/1-2988} and the WikiNER dataset \cite{pan-etal-2017-cross}. We also make use of the CoNLL-2003 shared task NER English dataset \url{https://www.clips.uantwerpen.be/conll2003/}
 
%

\paragraph{Languages}
For all our experiments, we use English, Russian and Arabic as source languages in addition to Chinese, Czech, Finish, French, Indonesian, Italian, Japanese, German, Hindi, Polish, Portuguese, Slovenian, Spanish, and Turkish as target languages.

\paragraph{Fine-tuning Data}

For all the cross-lingual experiments, we use English, Russian and Arabic as source languages on which we fine-tune \mbert. For English, we take the English-EWT treebank for fine-tuning, for Russian the Russian-GSD treebank and for Arabic the Arabic-PADT treebank. 

\paragraph{Evaluation Data}
For all our experiments, we perform the evaluation on all the 17 languages. For Parsing and POS tagging we use the test set from the PUD treebanks released for the CoNLL Shared Task 2017 \citep{zeman-etal-2017-conll}. For NER, we use the corresponding annotated datasets in the wikiner dataset. 


\paragraph{Domain Analysis Datasets}
\label{sec:domain_dataset_details}

We list here the datasets for completing our domain analysis experiment in Section~\ref{sec:domain_analysis} reported in Table \ref{tab:re_init_x2_domain_delta_small}.  To have a full control on the source domains, we use for fine-tuning the English Partut treebank for POS tagging and parsing \citep{Svizzera2014ConvertingTP}. It is a mix of legal, news and wikipedia text. For NER, we keep the WikiANN dataset \citep{pan-etal-2017-cross}.
For the \inlang and \outdomain experiments, we use the English-EWT, English-Lines and English Lexnorm \citep{van2018modeling} treebanks  for Web Media data, Literature data and Noisy tweets respectively. For the \crosslang French evaluation, we use the translation of the English test set,\footnote{We do so by taking the French-ParTUT test set that overlaps with the English-ParTUT, which includes 110 sentences.} as well as the French-GSD treebank. For NER, we take the CoNLL-2003 shared task English data as our \outdomain evaluation extracted from the \textit{News} domain. We note that the absolute performance on this dataset is not directly comparable to the one on the source wikiner. Indeed, the CoNLL-2003 dataset uses an extra \textsc{MISC} class. In our work, we only interpret the relative performance of different models on this test set. 


\paragraph{Cross-Lingual Similarity Analysis}
\label{sec:cross_sim_langs}

 For a given source language $l$ and a target language $l'$, we collect a 1000 pairs of aligned sentences from the UD-PUD treebanks \citep{zeman-etal-2017-conll}. For a given model and for each layer, we get a single sentence embedding by averaging token-level embeddings (after excluding special tokens). We then concatenate the 1000 sentence embedding vectors and get the matrices  $X_l$ and $X_l'$. Based on these two matrices, the CKA between the language $l$ and the language $l'$ is defined as:
\begin{equation*}
    CKA(X_{l}, X_{l'}) = \frac{||X_l^TX_{l'}||_F^2}{||X_l^TX_l||_F||X_{l'}^TX_{l'}||_F}
\end{equation*}

with $||.||_F$ defining the Frobenius norm. 

We do so for each source-target language pairs using the representation of the pretrained \mbert model as well as for \mbert fine-tuned on each downstream task. 

\begin{figure}[t!]
\centering
\includegraphics[width=1.0\columnwidth]{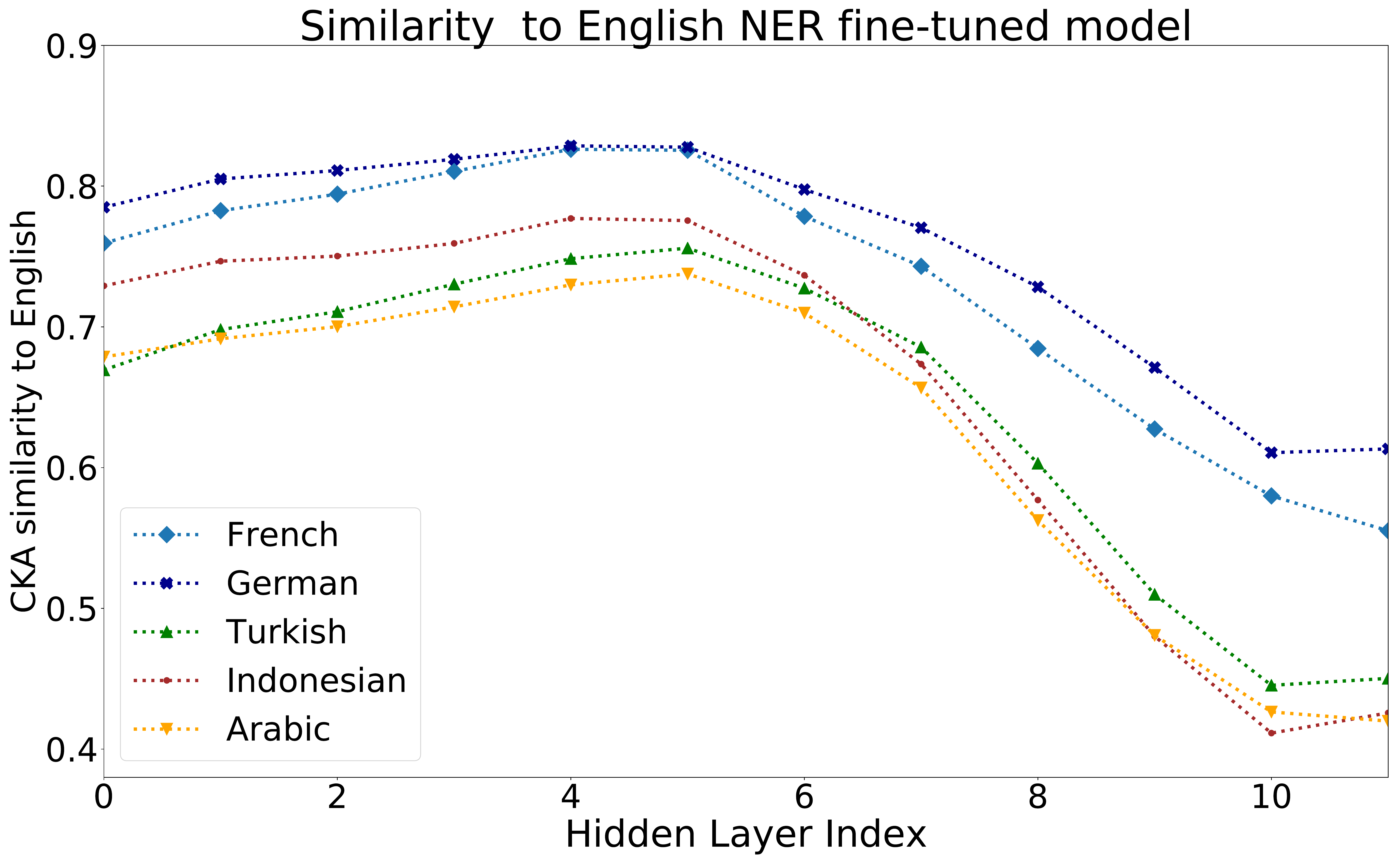}
\caption{Cross-Lingual Similarity (CKA)  (\S\ref{sec:alignement}) of hidden representations of a source language (English) sentences with target languages of \mbert fine-tuned for NER. The higher the CKA value the greater the similarity.}
\label{fig:ner_lang_comparison}
\end{figure}

In addition to the results presented in \S\ref{sec:alignement}, we report in Figure~\ref{fig:ner_lang_comparison}, a comparison of the cross-lingual similarity per hidden layer of \mbert fine-tuned on NER, across target languages. The trend is the same for all languages. 


\subsubsection{Computation}

\paragraph{Infrastructure}

Our experiments were ran on a shared cluster on the equivalent of 15 Nvidia Tesla T4 GPUs.\footnote{https://www.nvidia.com/en-sg/data-center/tesla-t4/}

\paragraph{Codebase}

All of our experiments are built using the Transformers library described in \citep{wolf-etal-2020-transformers}.  We also provide code to reproduce our experiments at \url{https://github.com/benjamin-mlr/first-align-then-predict.git}. 

\subsubsection{Preprocessing}

Our experiments are ran with word-level tokenization as provided in the datasets. We then tokenize each sequence of words at the sub-word level using the \textit{Wordpiece} algorithm of BERT and provided by the Transformers library.

\begin{table}[h!]
\footnotesize 
\centering
\begin{tabular}{lrrrr}
\toprule
  \textit{Params.}& Parsing  &  NER & POS & Bounds\\
  \hline
  batch size & 32 & 16 & 16 & [1,256] \\
  learning rate &  5e-5 & 3.5e-5 & 5e-5 & [1e-6,1e-3] \\
  epochs (best) &  15 & 6 & 6 & [1,50] \\
  \#grid & 60 & 60 & 180 & - \\
  Run-time (min) & 32 & 24 & 75 & -\\
\bottomrule

\end{tabular}

\caption{Fine-tuning best hyper-parameters for each task as selected on the validation set of the source language with bounds. \#grid: number of grid search trial. Run-time is reported in average for training and evaluation. 
}
\label{tab:hyperparameters}
\end{table}

\subsection{Cross-lingual transfer analyses}

\subsubsection{Correlation}

We report here in Figure~\ref{tab:correlation_table} the correlation between the hidden representation of each layer and the \gap between the source and the target averaged across all target languages and all layers. The correlation is strong and significant for all the tasks and for both the fine-tuned and the pretrained models. This shows that multilingual alignment that occurs within the models, learnt during pretraining is strongly related with cross-lingual transfer. 

\begin{figure}[t!]
\centering
\includegraphics[width=1.01\columnwidth]{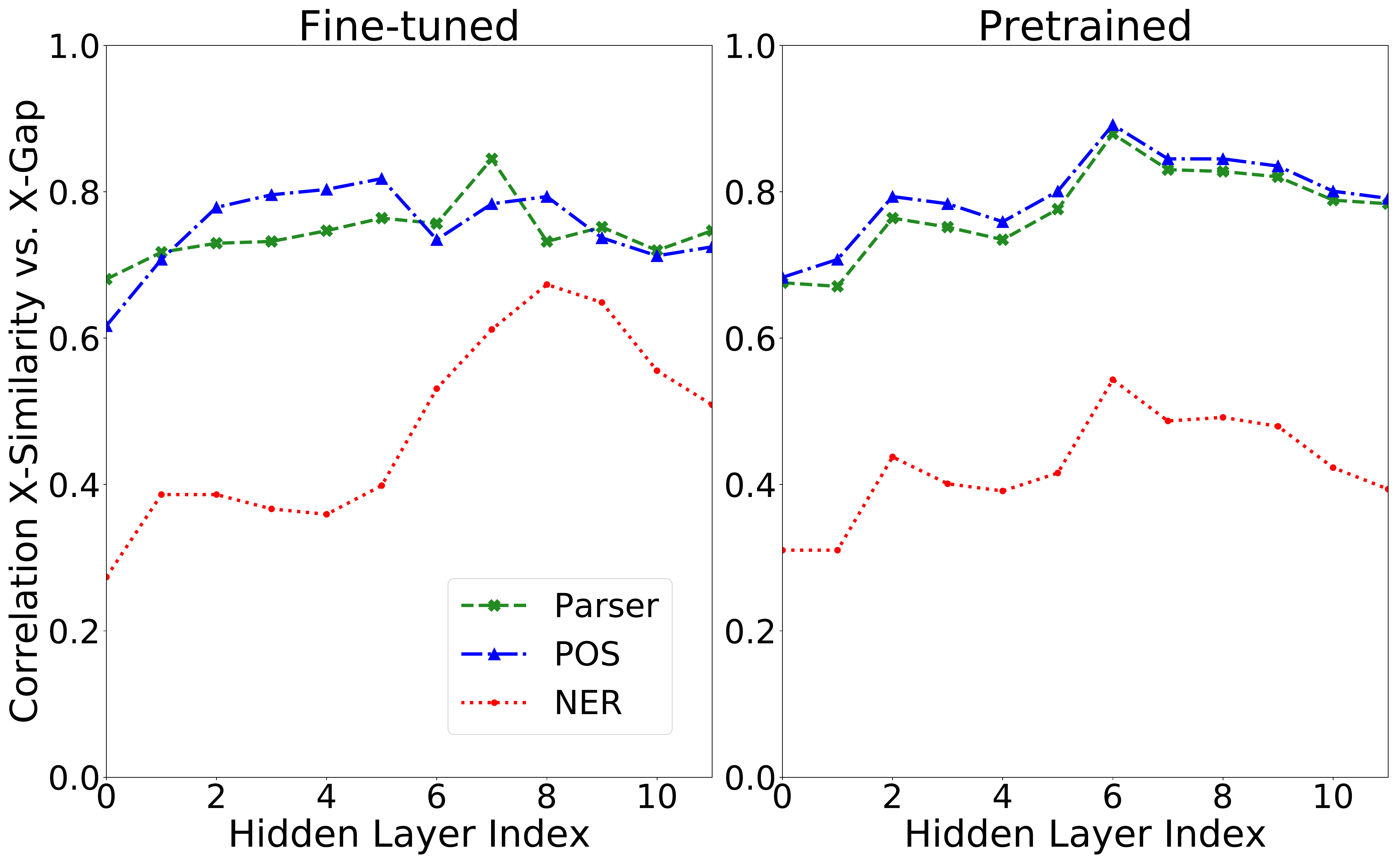}
\caption{Spearman Correlation between Cross-Lingual Similarity (CKA  between English and the target representations) and \gap averaged over all 17 target languages for each layer}
\label{fig:correlation_plot_layer}
\end{figure}

We report in Figure~\ref{fig:correlation_plot_layer}, the detail of this correlation per layer. For the pretrained model, we observe the same distribution for each task with layer 6 being the most correlated to cross-lingual transfer. We observe large variations in the fine-tuned cases, the most notable being NER. This illustrates the task-specific aspect of the relation between \similarity and cross-lingual transfer. More precisely, in the case of NER, the sharp increase and decrease in the upper part of the model provides new evidence that for this task, fine-tuning highly impacts  \similarity in the upper part of the model which correlates with \crosslang transfer.

\begin{table}[h!]
\footnotesize\centering
\scalebox{0.9}{ 
\begin{tabular}{lrr}
    \toprule
     & \multicolumn{2}{c}{Correlation} \\
    \textit{Task} &
    \multicolumn{2}{c}{X-Gap vs. X-Similarity} \\
    &\underline{\textsc{Fine-Tuned}} & \underline{\textsc{Pretrained}}\\
    Parsing & 0.76 & 0.79 \\
    POS &  0.74 & 0.82 \\ 
    NER* &  0.47 & 0.43\\
\bottomrule

\end{tabular}
}
\caption{
\small{Spearman-Rank Correlation between the \textit{Cross-lingual Gap} (X-Lang Gap) and the \textit{Cross-lingual Similarity} between the source and the target languages of the fine-tuned models and the pretrained model averaged over all the hidden layers and all the 17 target languages (sample size per task: 17). 
For NER, \gap measured on wikiner data and not on the parrallel data itself in constrast with Parsing and POS tagging. 
Complete list of languages can be found in Appendix~\ref{sec:cross_sim_langs}
}}
\label{tab:correlation_table}
\end{table}

\begin{table*}[h!]
\footnotesize\centering
\scalebox{0.81}{ 
\begin{tabular}{lrrrrrrrrrrrrr}
    \toprule
     &  \multicolumn{13}{c}{\underline{\textit{\reinit \textit{of layers}}}} \\
     Eval&\textsc{\standardFineTune} & \textsc{All} & \textsc{1} & \textsc{2} & \textsc{1-2} &
     \textsc{3-4} &\textsc{1-3} &\textsc{4-6} &\textsc{7-9} &\textsc{10-12} & \textsc{1-4} & \textsc{5-8} & \textsc{9-12} \\
    \midrule

& \multicolumn{13}{c}{\textit{Parsing}} \\

 \textsc{English Dev} & 88.52 & \cellcolor{brightlavender}74.66 & \cellcolor{lightgreen}87.77 & \cellcolor{lightgreen}88.03 & \cellcolor{lightgreen}87.28 & \cellcolor{lightgreen}86.81 & \cellcolor{lightgreen}83.77 & \cellcolor{lightgreen}85.86 & \cellcolor{lightgreen}87.53 & \cellcolor{mediumspringgreen}88.78 & \cellcolor{lightgreen}84.30 & \cellcolor{lightgreen}85.41 & \cellcolor{lightgreen}88.35 \\ 
   \textsc{English Test} & 88.59 & \cellcolor{brightlavender}74.58 & \cellcolor{lightgreen}87.77 & \cellcolor{lightgreen}88.09 & \cellcolor{lightgreen}87.25 & \cellcolor{lightgreen}86.79 & \cellcolor{pink}83.37 & \cellcolor{lightgreen}85.54 & \cellcolor{lightgreen}87.36 & \cellcolor{mediumspringgreen}88.62 & \cellcolor{pink}83.10 & \cellcolor{lightgreen}85.37 & \cellcolor{mediumspringgreen}88.69 \\ 
   
  \hdashline
 \textsc{French} & 68.94 & \cellcolor{brightlavender}3.70 & \cellcolor{lightgreen}65.73 & \cellcolor{lightgreen}65.21 & \cellcolor{brightlavender}55.31 & \cellcolor{pink}61.31 & \cellcolor{brightlavender}43.81 & \cellcolor{pink}61.77 & \cellcolor{lightgreen}67.03 & \cellcolor{mediumspringgreen}69.36 & \cellcolor{brightlavender}37.29 & \cellcolor{pink}61.82 & \cellcolor{mediumspringgreen}69.26 \\ 
  \textsc{German} & 67.43 & \cellcolor{brightlavender}4.73 & \cellcolor{lightgreen}64.97 & \cellcolor{lightgreen}65.20 & \cellcolor{brightlavender}57.08 & \cellcolor{pink}60.62 & \cellcolor{brightlavender}47.85 & \cellcolor{pink}58.93 & \cellcolor{lightgreen}64.12 & \cellcolor{lightgreen}66.67 & \cellcolor{brightlavender}36.05 & \cellcolor{pink}59.37 & \cellcolor{lightgreen}67.21 \\ 
  \textsc{Turkish} & 28.40 & \cellcolor{brightlavender}2.76 & \cellcolor{pink}21.65 & \cellcolor{lightgreen}23.77 & \cellcolor{brightlavender}16.78 & \cellcolor{pink}21.21 & \cellcolor{brightlavender}10.69 & \cellcolor{pink}20.23 & \cellcolor{lightgreen}25.39 & \cellcolor{mediumspringgreen}30.43 & \cellcolor{brightlavender}9.70 & \cellcolor{pink}20.94 & \cellcolor{mediumspringgreen}29.33 \\ 
 \textsc{Indonesian} & 45.13 & \cellcolor{brightlavender}4.99 & \cellcolor{lightgreen}43.33 & \cellcolor{lightgreen}43.48 & \cellcolor{pink}39.83 & \cellcolor{pink}39.09 & \cellcolor{brightlavender}33.06 & \cellcolor{lightgreen}40.65 & \cellcolor{lightgreen}44.42 & \cellcolor{mediumspringgreen}46.96 & \cellcolor{brightlavender}30.35 & \cellcolor{lightgreen}40.85 & \cellcolor{mediumspringgreen}47.53 \\ 
  \textsc{Russian} & 59.70 & \cellcolor{brightlavender}2.95 & \cellcolor{lightgreen}57.81 & \cellcolor{lightgreen}57.53 & \cellcolor{pink}54.10 & \cellcolor{pink}53.51 & \cellcolor{brightlavender}47.01 & \cellcolor{pink}52.37 & \cellcolor{lightgreen}56.45 & \cellcolor{mediumspringgreen}61.41 & \cellcolor{brightlavender}38.58 & \cellcolor{pink}52.41 & \cellcolor{mediumspringgreen}60.72 \\ 
  \textsc{Arabic} & 23.37 & \cellcolor{brightlavender}3.19 & \cellcolor{mediumspringgreen}23.66 & \cellcolor{mediumspringgreen}23.49 & \cellcolor{lightgreen}21.01 & \cellcolor{lightgreen}19.55 & \cellcolor{pink}16.17 & \cellcolor{lightgreen}18.84 & \cellcolor{lightgreen}20.70 & \cellcolor{mediumspringgreen}24.54 & \cellcolor{brightlavender}13.26 & \cellcolor{pink}18.27 & \cellcolor{mediumspringgreen}23.93 \\ 
 
\midrule
& \multicolumn{13}{c}{\textit{POS}} \\

  \textsc{English Dev} & 96.45 & \cellcolor{pink}87.47 & \cellcolor{lightgreen}96.04 & \cellcolor{lightgreen}96.06 & \cellcolor{lightgreen}95.92 & \cellcolor{lightgreen}95.81 & \cellcolor{lightgreen}95.38 & \cellcolor{lightgreen}95.43 & \cellcolor{lightgreen}96.25 & \cellcolor{mediumspringgreen}96.58 & \cellcolor{lightgreen}94.01 & \cellcolor{lightgreen}95.35 & \cellcolor{lightgreen}96.39 \\

 \textsc{English Test} & 96.53 & \cellcolor{pink}87.71 & \cellcolor{lightgreen}96.08 & \cellcolor{lightgreen}96.24 & \cellcolor{lightgreen}95.94 & \cellcolor{lightgreen}95.72 & \cellcolor{lightgreen}95.40 & \cellcolor{lightgreen}95.59 & \cellcolor{lightgreen}96.34 & \cellcolor{mediumspringgreen}96.74 & \cellcolor{lightgreen}94.05 & \cellcolor{lightgreen}95.45 & \cellcolor{lightgreen}96.51 \\

 \hdashline
  \textsc{French} & 88.25 & \cellcolor{brightlavender}28.96 & \cellcolor{lightgreen}86.70 & \cellcolor{lightgreen}87.66 & \cellcolor{pink}79.84 & \cellcolor{lightgreen}87.14 & \cellcolor{brightlavender}69.43 & \cellcolor{lightgreen}86.42 & \cellcolor{lightgreen}86.94 & \cellcolor{mediumspringgreen}88.30 & \cellcolor{brightlavender}62.28 & \cellcolor{lightgreen}86.37 & \cellcolor{mediumspringgreen}88.26 \\ 
  \textsc{German} & 90.63 & \cellcolor{brightlavender}28.93 & \cellcolor{lightgreen}88.26 & \cellcolor{lightgreen}89.53 & \cellcolor{pink}82.26 & \cellcolor{lightgreen}88.39 & \cellcolor{brightlavender}71.63 & \cellcolor{lightgreen}88.30 & \cellcolor{lightgreen}90.26 & \cellcolor{mediumspringgreen}90.83 & \cellcolor{brightlavender}59.16 & \cellcolor{lightgreen}89.12 & \cellcolor{mediumspringgreen}90.64 \\ 
 \textsc{Turkish} & 72.65 & \cellcolor{brightlavender}32.23 & \cellcolor{brightlavender}62.17 & \cellcolor{pink}66.17 & \cellcolor{brightlavender}54.50 & \cellcolor{pink}63.22 & \cellcolor{brightlavender}47.77 & \cellcolor{pink}66.37 & \cellcolor{lightgreen}70.91 & \cellcolor{mediumspringgreen}72.92 & \cellcolor{brightlavender}44.16 & \cellcolor{lightgreen}69.30 & \cellcolor{mediumspringgreen}73.08 \\ 
 \textsc{Indonesian} & 84.06 & \cellcolor{brightlavender}36.98 & \cellcolor{lightgreen}82.15 & \cellcolor{lightgreen}82.89 & \cellcolor{lightgreen}80.13 & \cellcolor{lightgreen}81.40 & \cellcolor{pink}75.94 & \cellcolor{lightgreen}81.99 & \cellcolor{lightgreen}83.78 & \cellcolor{mediumspringgreen}84.42 & \cellcolor{brightlavender}72.42 & \cellcolor{lightgreen}82.59 & \cellcolor{mediumspringgreen}84.09 \\ 
  \textsc{Russian} & 82.97 & \cellcolor{brightlavender}32.63 & \cellcolor{mediumspringgreen}83.14 & \cellcolor{mediumspringgreen}83.63 & \cellcolor{lightgreen}81.95 & \cellcolor{lightgreen}82.26 & \cellcolor{pink}77.93 & \cellcolor{lightgreen}81.69 & \cellcolor{mediumspringgreen}82.98 & \cellcolor{lightgreen}81.76 & \cellcolor{brightlavender}70.33 & \cellcolor{lightgreen}82.56 & \cellcolor{mediumspringgreen}83.19 \\ 
  \textsc{Arabic} & 56.66 & \cellcolor{brightlavender}19.61 & \cellcolor{mediumspringgreen}58.10 & \cellcolor{mediumspringgreen}58.06 & \cellcolor{mediumspringgreen}57.89 & \cellcolor{lightgreen}55.62 & \cellcolor{mediumspringgreen}57.93 & \cellcolor{lightgreen}54.69 & \cellcolor{lightgreen}56.04 & \cellcolor{lightgreen}55.97 & \cellcolor{lightgreen}52.28 & \cellcolor{lightgreen}53.60 & \cellcolor{mediumspringgreen}58.84 \\ 
\midrule
& \multicolumn{13}{c}{\textit{NER}} \\
 \textsc{English Dev} & 83.29 & \cellcolor{brightlavender}56.99 & \cellcolor{lightgreen}82.04 & \cellcolor{lightgreen}82.26 & \cellcolor{lightgreen}79.52 & \cellcolor{lightgreen}80.36 & \cellcolor{pink}76.22 & \cellcolor{lightgreen}79.53 & \cellcolor{lightgreen}82.18 & \cellcolor{lightgreen}82.53 & \cellcolor{brightlavender}69.31 & \cellcolor{lightgreen}80.05 & \cellcolor{lightgreen}82.47 \\ 

  \textsc{English Test} & 83.06 & \cellcolor{brightlavender}56.56 & \cellcolor{lightgreen}81.46 & \cellcolor{lightgreen}82.00 & \cellcolor{lightgreen}79.63 & \cellcolor{lightgreen}79.25 & \cellcolor{pink}76.68 & \cellcolor{lightgreen}78.93 & \cellcolor{lightgreen}81.64 & \cellcolor{lightgreen}82.39 & \cellcolor{brightlavender}69.08 & \cellcolor{lightgreen}79.91 & \cellcolor{lightgreen}82.27 \\ 
 \hdashline
 \textsc{French} & 76.76 & \cellcolor{brightlavender}35.35 & \cellcolor{lightgreen}75.46 & \cellcolor{mediumspringgreen}77.57 & \cellcolor{pink}69.94 & \cellcolor{lightgreen}72.83 & \cellcolor{brightlavender}65.14 & \cellcolor{pink}70.34 & \cellcolor{lightgreen}75.42 & \cellcolor{lightgreen}75.90 & \cellcolor{brightlavender}55.79 & \cellcolor{lightgreen}73.12 & \cellcolor{lightgreen}75.77 \\ 
 \textsc{German} & 76.68 & \cellcolor{brightlavender}18.95 & \cellcolor{lightgreen}73.73 & \cellcolor{lightgreen}75.39 & \cellcolor{brightlavender}66.18 & \cellcolor{pink}70.12 & \cellcolor{brightlavender}56.50 & \cellcolor{pink}69.53 & \cellcolor{lightgreen}75.38 & \cellcolor{mediumspringgreen}77.11 & \cellcolor{brightlavender}42.37 & \cellcolor{pink}71.14 & \cellcolor{lightgreen}75.50 \\ 
  \textsc{Turkish} & 67.64 & \cellcolor{brightlavender}20.76 & \cellcolor{pink}62.54 & \cellcolor{lightgreen}64.84 & \cellcolor{brightlavender}52.20 & \cellcolor{brightlavender}57.11 & \cellcolor{brightlavender}53.03 & \cellcolor{pink}60.59 & \cellcolor{lightgreen}65.66 & \cellcolor{lightgreen}64.87 & \cellcolor{brightlavender}39.38 & \cellcolor{pink}61.43 & \cellcolor{lightgreen}66.62 \\ 
 \textsc{Indonesian} & 53.47 & \cellcolor{brightlavender}21.20 & \cellcolor{lightgreen}49.19 & \cellcolor{lightgreen}49.27 & \cellcolor{pink}46.50 & \cellcolor{pink}46.87 & \cellcolor{pink}43.75 & \cellcolor{pink}47.83 & \cellcolor{mediumspringgreen}54.39 & \cellcolor{lightgreen}48.71 & \cellcolor{brightlavender}36.11 & \cellcolor{pink}46.06 & \cellcolor{pink}48.23 \\ 
  \textsc{Russian} & 58.23 & \cellcolor{brightlavender}7.43 & \cellcolor{lightgreen}55.63 & \cellcolor{lightgreen}58.08 & \cellcolor{pink}50.67 & \cellcolor{pink}52.89 & \cellcolor{brightlavender}42.83 & \cellcolor{brightlavender}46.13 & \cellcolor{lightgreen}53.38 & \cellcolor{lightgreen}58.09 & \cellcolor{brightlavender}34.66 & \cellcolor{pink}52.03 & \cellcolor{mediumspringgreen}59.12 \\ 
  \textsc{Arabic} & 41.81 & \cellcolor{brightlavender}5.49 & \cellcolor{pink}35.79 & \cellcolor{pink}34.80 & \cellcolor{pink}32.37 & \cellcolor{pink}32.31 & \cellcolor{brightlavender}26.21 & \cellcolor{lightgreen}38.88 & \cellcolor{lightgreen}38.55 & \cellcolor{lightgreen}40.83 & \cellcolor{brightlavender}21.85 & \cellcolor{lightgreen}38.67 & \cellcolor{lightgreen}41.23 \\

\bottomrule

\end{tabular}
}
\caption{Zero-shot cross-lingual performance  when applying \reinit to specific set of consecutive layers compared to the \standardFineTune model. Source language is English. Baseline model \textsc{ALL} (for all layers randomly initialized) corresponds to a model trained from scratch on the task.
For reproducibility purposes, we report performance on the Validation set \textsc{English Dev}. For all target languages, we report the scores on the test split of each dataset.
Each score is the average of 5 runs with different random seeds.
For more insights into the variability of our results, we report the min., median and max. value  of the standard deviations (std) across runs with different random seeds for each task:  Parsing:0.02/0.34/1.48, POS:0.01/0.5/2.38, NER:0.0/0.47/2.62 (std min/median/max). 
\\
\colorbox{mediumspringgreen}{$\geq$ \textsc{\standardFineTune}}
\colorbox{lightgreen}{$<$  \textsc{\standardFineTune}}
\colorbox{pink}{$\leq$ 5 points}
\colorbox{brightlavender}{$\leq$ 10 points }
}
\label{tab:re_init_full_ablation}
\end{table*}

\begin{table*}[h!]
\footnotesize\centering
\scalebox{0.78}{ 
\begin{tabular}{lrrrrrrrr}
    \toprule
        & \multicolumn{8}{c}{\underline{\textit{\reinit \textit{of layers}}}} \\
     \textsc{Source - Target} &\textsc{\standardFineTune}  & $\Delta$ \textsc{0-1} & $\Delta$ \textsc{2-3} & $\Delta$ \textsc{4-5} & $\Delta$ \textsc{6-7} & $\Delta$ \textsc{8-9} &  $\Delta$ \textsc{10-11} \\
     &&\multicolumn{7}{c}{\textbf{\textit{Parsing}}}\\
\textsc{EN - English} & 88.98 & \cellcolor{lightgreen}-0.96 & \cellcolor{lightgreen}-0.66 & \cellcolor{lightgreen}-0.93 & \cellcolor{lightgreen}-0.55 & \cellcolor{mediumspringgreen}0.04 & \cellcolor{lightgreen}-0.09 \\ 
\hdashline
\textsc{EN - Arabic} & 35.88 & \cellcolor{pink}-4.05 & \cellcolor{pink}-2.38 & \cellcolor{pink}-3.16 & \cellcolor{lightgreen}-0.78 & \cellcolor{mediumspringgreen}1.74 & \cellcolor{mediumspringgreen}1.68 \\ 
\textsc{EN - French} & 74.04 & \cellcolor{brightlavender}-21.30 & \cellcolor{brightlavender}-6.84 & \cellcolor{pink}-2.93 & \cellcolor{lightgreen}-0.69 & \cellcolor{mediumspringgreen}0.03 & \cellcolor{mediumspringgreen}0.76 \\ 
\textsc{EN - German} & 70.34 & \cellcolor{brightlavender}-15.06 & \cellcolor{brightlavender}-9.26 & \cellcolor{pink}-4.75 & \cellcolor{lightgreen}-1.54 & \cellcolor{lightgreen}-0.29 & \cellcolor{mediumspringgreen}1.82 \\ 
\textsc{EN - Turkish} & 34.03 & \cellcolor{brightlavender}-16.37 & \cellcolor{brightlavender}-10.10 & \cellcolor{brightlavender}-5.11 & \cellcolor{pink}-3.71 & \cellcolor{mediumspringgreen}0.43 & \cellcolor{mediumspringgreen}1.43 &\\
\textsc{EN - Indo} & 44.11 & \cellcolor{brightlavender}-10.57 & \cellcolor{brightlavender}-5.87 & \cellcolor{pink}-2.66 & \cellcolor{lightgreen}-0.96 & \cellcolor{lightgreen}-0.74 & \cellcolor{mediumspringgreen}0.73 \\ 
\textsc{EN - Russian} & 62.52 & \cellcolor{brightlavender}-7.31 & \cellcolor{brightlavender}-5.37 & \cellcolor{pink}-2.84 & \cellcolor{lightgreen}-1.09 & \cellcolor{mediumspringgreen}0.44 & \cellcolor{mediumspringgreen}0.71 \\ 
\textsc{EN - Porthughese} & 68.59 & \cellcolor{brightlavender}-25.83 & \cellcolor{brightlavender}-6.22 & \cellcolor{pink}-2.97 & \cellcolor{lightgreen}-0.77 & \cellcolor{mediumspringgreen}0.15 & \cellcolor{mediumspringgreen}0.82 \\ 
\textsc{EN - SPanish} & 69.96 & \cellcolor{brightlavender}-18.05 & \cellcolor{brightlavender}-5.74 & \cellcolor{pink}-2.78 & \cellcolor{lightgreen}-0.96 & \cellcolor{mediumspringgreen}0.13 & \cellcolor{mediumspringgreen}0.72 \\ 
\textsc{EN - Finish} & 48.42 & \cellcolor{brightlavender}-24.25 & \cellcolor{brightlavender}-9.48 & \cellcolor{pink}-4.39 & \cellcolor{pink}-2.51 & \cellcolor{lightgreen}-0.28 & \cellcolor{mediumspringgreen}0.22 \\ 
\textsc{EN - Italian} & 74.54 & \cellcolor{brightlavender}-30.54 & \cellcolor{brightlavender}-9.63 & \cellcolor{pink}-4.18 & \cellcolor{lightgreen}-1.32 & \cellcolor{lightgreen}-0.12 & \cellcolor{mediumspringgreen}0.90 \\ 
\textsc{EN - Slovenian} & 73.04 & \cellcolor{brightlavender}-29.89 & \cellcolor{brightlavender}-6.52 & \cellcolor{pink}-3.00 & \cellcolor{lightgreen}-1.68 & \cellcolor{lightgreen}-0.05 & \cellcolor{mediumspringgreen}0.18 \\ 
\textsc{EN - Czech} & 60.44 & \cellcolor{brightlavender}-31.84 & \cellcolor{brightlavender}-10.69 & \cellcolor{pink}-4.61 & \cellcolor{lightgreen}-1.82 & \cellcolor{mediumspringgreen}0.18 & \cellcolor{mediumspringgreen}1.17 \\ 
\textsc{EN - Polish} & 55.23 & \cellcolor{brightlavender}-23.57 & \cellcolor{brightlavender}-9.11 & \cellcolor{pink}-3.34 & \cellcolor{lightgreen}-1.83 & \cellcolor{mediumspringgreen}0.28 & \cellcolor{mediumspringgreen}0.89 \\ 
\textsc{EN - Hindi} & 28.86 & \cellcolor{brightlavender}-9.13 & \cellcolor{brightlavender}-7.58 & \cellcolor{brightlavender}-5.84 & \cellcolor{pink}-2.50 & \cellcolor{mediumspringgreen}1.35 & \cellcolor{mediumspringgreen}1.49 \\ 
\textsc{EN - Chinese} & 27.48 & \cellcolor{brightlavender}-7.31 & \cellcolor{pink}-4.47 & \cellcolor{lightgreen}-1.65 & \cellcolor{lightgreen}-0.62 & \cellcolor{mediumspringgreen}0.65 & \cellcolor{mediumspringgreen}1.32 \\ 
\textsc{EN - Japanese} & 11.99 & \cellcolor{pink}-4.36 & \cellcolor{pink}-2.76 & \cellcolor{lightgreen}-1.91 & \cellcolor{lightgreen}-1.19 & \cellcolor{mediumspringgreen}0.47 & \cellcolor{mediumspringgreen}1.12 \\ 
\hdashline
\textsc{EN - X (mean)}  & 53.23 & \cellcolor{brightlavender} -15.77 & \cellcolor{brightlavender} -6.51 & \cellcolor{pink} -3.39 & \cellcolor{lightgreen} -1.47 & \cellcolor{mediumspringgreen} 0.29 & \cellcolor{mediumspringgreen} 1.00 \\ 
\midrule
\textsc{Ru - Russian} & 85.15 & \cellcolor{lightgreen}-0.82 & \cellcolor{lightgreen}-1.38 & \cellcolor{lightgreen}-1.51 & \cellcolor{lightgreen}-0.86 & \cellcolor{lightgreen}-0.29 & \cellcolor{mediumspringgreen}0.18 & \\
\hdashline
\textsc{Ru - English} & 61.40 & \cellcolor{brightlavender}-8.37 & \cellcolor{pink}-3.55 & \cellcolor{pink}-3.90 & \cellcolor{lightgreen}-0.72 & \cellcolor{mediumspringgreen}1.77 & \cellcolor{mediumspringgreen}1.14 \\ 
\textsc{Ru - Arabic} & 59.41 & \cellcolor{brightlavender}-5.65 & \cellcolor{brightlavender}-5.26 & \cellcolor{brightlavender}-5.15 & \cellcolor{lightgreen}-1.47 & \cellcolor{mediumspringgreen}0.24 & \cellcolor{mediumspringgreen}0.16 \\ 
\textsc{Ru - French} & 65.84 & \cellcolor{brightlavender}-8.87 & \cellcolor{pink}-2.93 & \cellcolor{lightgreen}-1.81 & \cellcolor{lightgreen}-1.05 & \cellcolor{mediumspringgreen}3.81 & \cellcolor{mediumspringgreen}1.24 \\ 
\textsc{Ru - German} & 65.90 & \cellcolor{brightlavender}-7.02 & \cellcolor{pink}-4.19 & \cellcolor{lightgreen}-1.97 & \cellcolor{lightgreen}-1.45 & \cellcolor{mediumspringgreen}2.58 & \cellcolor{mediumspringgreen}2.05 \\ 
\textsc{Ru - Turkish} & 32.20 & \cellcolor{brightlavender}-13.13 & \cellcolor{brightlavender}-7.18 & \cellcolor{brightlavender}-6.82 & \cellcolor{pink}-3.77 & \cellcolor{lightgreen}-0.85 & \cellcolor{mediumspringgreen}1.21 &\\  
\textsc{Ru - Indo} & 47.59 & \cellcolor{pink}-4.74 & \cellcolor{pink}-2.99 & \cellcolor{pink}-2.30 & \cellcolor{lightgreen}-1.81 & \cellcolor{mediumspringgreen}0.04 & \cellcolor{mediumspringgreen}1.02 \\ 
\textsc{Ru - Porthughese} & 66.41 & \cellcolor{brightlavender}-11.17 & \cellcolor{lightgreen}-1.61 & \cellcolor{lightgreen}-1.09 & \cellcolor{lightgreen}-1.25 & \cellcolor{mediumspringgreen}4.16 & \cellcolor{mediumspringgreen}1.94 \\ 
\textsc{Ru - SPanish} & 66.74 & \cellcolor{pink}-4.52 & \cellcolor{lightgreen}-1.38 & \cellcolor{lightgreen}-0.69 & \cellcolor{lightgreen}-0.97 & \cellcolor{mediumspringgreen}2.95 & \cellcolor{mediumspringgreen}1.37 \\ 
\textsc{Ru - Finish} & 52.92 & \cellcolor{brightlavender}-15.43 & \cellcolor{brightlavender}-6.59 & \cellcolor{pink}-4.09 & \cellcolor{lightgreen}-1.35 & \cellcolor{mediumspringgreen}0.12 & \cellcolor{mediumspringgreen}0.77 \\ 
\textsc{Ru - Italian} & 65.28 & \cellcolor{brightlavender}-12.97 & \cellcolor{pink}-3.56 & \cellcolor{pink}-2.34 & \cellcolor{lightgreen}-1.46 & \cellcolor{mediumspringgreen}3.16 & \cellcolor{mediumspringgreen}1.55 \\ 
\textsc{Ru - Slovenian} & 62.91 & \cellcolor{brightlavender}-16.67 & \cellcolor{pink}-2.71 & \cellcolor{pink}-3.18 & \cellcolor{lightgreen}-1.03 & \cellcolor{mediumspringgreen}0.31 & \cellcolor{mediumspringgreen}1.08 \\ 
\textsc{Ru - Czech} & 72.77 & \cellcolor{brightlavender}-11.95 & \cellcolor{pink}-4.17 & \cellcolor{pink}-3.13 & \cellcolor{lightgreen}-1.57 & \cellcolor{lightgreen}-0.33 & \cellcolor{mediumspringgreen}0.30 \\ 
\textsc{Ru - Polish} & 66.07 & \cellcolor{brightlavender}-5.70 & \cellcolor{pink}-3.22 & \cellcolor{pink}-2.57 & \cellcolor{lightgreen}-1.54 & \cellcolor{lightgreen}-0.12 & \cellcolor{mediumspringgreen}0.54 \\ 
\textsc{Ru - Hindi} & 28.67 & \cellcolor{brightlavender}-6.02 & \cellcolor{brightlavender}-5.77 & \cellcolor{brightlavender}-5.27 & \cellcolor{pink}-3.75 & \cellcolor{lightgreen}-0.06 & \cellcolor{mediumspringgreen}0.99 \\ 
\textsc{Ru - Chinese} & 28.77 & \cellcolor{pink}-4.66 & \cellcolor{pink}-4.38 & \cellcolor{pink}-3.22 & \cellcolor{lightgreen}-1.80 & \cellcolor{mediumspringgreen}0.15 & \cellcolor{mediumspringgreen}1.12 \\ 
\textsc{Ru - Japanese} & 15.10 & \cellcolor{pink}-4.89 & \cellcolor{pink}-3.56 & \cellcolor{pink}-3.95 & \cellcolor{pink}-3.11 & \cellcolor{mediumspringgreen}0.68 & \cellcolor{mediumspringgreen}0.73 \\ 
\hdashline
\textsc{Ru - X (Mean)}  &  55.41 & \cellcolor{brightlavender} -7.69 & \cellcolor{pink} -3.71 & \cellcolor{pink} -3.13 & \cellcolor{lightgreen} -1.70 & \cellcolor{mediumspringgreen} 0.92 & \cellcolor{mediumspringgreen} 0.94 \\ 
 \midrule
\textsc{Ar - Arabic} & 59.54 & \cellcolor{lightgreen}-0.78 & \cellcolor{pink}-2.14 & \cellcolor{lightgreen}-1.20 & \cellcolor{lightgreen}-0.67 & \cellcolor{lightgreen}-0.27 & \cellcolor{mediumspringgreen}0.08 \\ 
\hdashline
\textsc{Ar - English} & 25.46 & \cellcolor{pink}-2.09 & \cellcolor{pink}-2.92 & \cellcolor{lightgreen}-0.90 & \cellcolor{lightgreen}-1.40 & \cellcolor{lightgreen}-0.97 & \cellcolor{lightgreen}-0.61 \\ 
\textsc{Ar - French} & 28.92 & \cellcolor{pink}-4.85 & \cellcolor{lightgreen}-1.45 & \cellcolor{lightgreen}-0.25 & \cellcolor{pink}-2.72 & \cellcolor{lightgreen}-1.60 & \cellcolor{lightgreen}-0.88 \\ 
\textsc{Ar - German} & 27.14 & \cellcolor{brightlavender}-6.38 & \cellcolor{pink}-4.51 & \cellcolor{lightgreen}-0.98 & \cellcolor{pink}-2.24 & \cellcolor{mediumspringgreen}0.13 & \cellcolor{mediumspringgreen}0.09 \\ 
\textsc{Ar - Turkish} & 9.58 & \cellcolor{pink}-3.90 & \cellcolor{pink}-3.14 & \cellcolor{pink}-2.76 & \cellcolor{pink}-2.33 & \cellcolor{mediumspringgreen}0.31 & \cellcolor{mediumspringgreen}0.15 & \\
\textsc{Ar - Indo} & 36.16 & \cellcolor{brightlavender}-5.85 & \cellcolor{pink}-4.86 & \cellcolor{lightgreen}-1.71 & \cellcolor{lightgreen}-0.68 & \cellcolor{lightgreen}-0.17 & \cellcolor{mediumspringgreen}0.58 \\ 
\textsc{Ar - Russian} & 42.25 & \cellcolor{pink}-3.52 & \cellcolor{brightlavender}-5.28 & \cellcolor{pink}-2.46 & \cellcolor{lightgreen}-1.66 & \cellcolor{lightgreen}-0.67 & \cellcolor{lightgreen}-0.27 \\ 
\textsc{Ar - Porthughese} & 34.71 & \cellcolor{pink}-4.80 & \cellcolor{lightgreen}-1.22 & \cellcolor{mediumspringgreen}0.10 & \cellcolor{pink}-2.98 & \cellcolor{lightgreen}-0.33 & \cellcolor{lightgreen}-0.24 \\ 
\textsc{Ar - SPanish} & 31.95 & \cellcolor{pink}-4.02 & \cellcolor{lightgreen}-0.15 & \cellcolor{lightgreen}-0.44 & \cellcolor{lightgreen}-1.46 & \cellcolor{lightgreen}-0.77 & \cellcolor{mediumspringgreen}0.38 \\ 
\textsc{Ar - Finish} & 28.18 & \cellcolor{brightlavender}-9.89 & \cellcolor{brightlavender}-7.03 & \cellcolor{pink}-3.17 & \cellcolor{lightgreen}-1.81 & \cellcolor{lightgreen}-0.58 & \cellcolor{lightgreen}-0.42 \\ 
\textsc{Ar - Italian} & 28.85 & \cellcolor{pink}-3.01 & \cellcolor{mediumspringgreen}0.60 & \cellcolor{mediumspringgreen}1.45 & \cellcolor{pink}-2.26 & \cellcolor{lightgreen}-1.47 & \cellcolor{lightgreen}-0.70 \\ 
\textsc{Ar - Slovenian} & 35.78 & \cellcolor{brightlavender}-9.73 & \cellcolor{pink}-4.97 & \cellcolor{pink}-2.21 & \cellcolor{lightgreen}-1.43 & \cellcolor{lightgreen}-0.41 & \cellcolor{lightgreen}-0.56 \\ 
\textsc{Ar - Czech} & 40.04 & \cellcolor{brightlavender}-13.61 & \cellcolor{brightlavender}-6.82 & \cellcolor{pink}-3.20 & \cellcolor{pink}-2.38 & \cellcolor{lightgreen}-1.12 & \cellcolor{lightgreen}-0.21 \\ 
\textsc{Ar - Polish} & 41.16 & \cellcolor{brightlavender}-8.46 & \cellcolor{brightlavender}-5.52 & \cellcolor{pink}-2.48 & \cellcolor{lightgreen}-1.48 & \cellcolor{lightgreen}-0.47 & \cellcolor{lightgreen}-0.55 \\ 
\textsc{Ar - Hindi} & 10.24 & \cellcolor{pink}-2.46 & \cellcolor{pink}-2.86 & \cellcolor{pink}-2.57 & \cellcolor{lightgreen}-1.55 & \cellcolor{mediumspringgreen}1.00 & \cellcolor{mediumspringgreen}0.14 \\ 
\textsc{Ar - Chinese} & 11.46 & \cellcolor{pink}-2.42 & \cellcolor{pink}-2.43 & \cellcolor{lightgreen}-1.26 & \cellcolor{lightgreen}-0.82 & \cellcolor{mediumspringgreen}0.23 & \cellcolor{lightgreen}-0.05 \\ 
\textsc{Ar - Japanese} & 6.66 & \cellcolor{lightgreen}-1.28 & \cellcolor{lightgreen}-0.79 & \cellcolor{lightgreen}-1.20 & \cellcolor{lightgreen}-1.04 & \cellcolor{mediumspringgreen}0.74 & \cellcolor{mediumspringgreen}0.30 \\ 
\hdashline
 \textsc{Ar - X (Mean) }  &  27.97 & \cellcolor{pink} -4.91 & \cellcolor{pink} -3.17 & \cellcolor{lightgreen} -1.48 & \cellcolor{lightgreen} -1.68 & \cellcolor{lightgreen} -0.36 & \cellcolor{lightgreen} -0.14 \\ 


\bottomrule
\end{tabular}}
\caption{Parsing (LAS score) Relative Zero shot Cross-Lingual performance of \mbert with \reinit (section \ref{sec:controlled_exp}) on pairs of consecutive layers compared to \mbert without any random-initialization (\standardFineTune). In \textsc{Src - Trg}, \textsc{Src} indicates the source language on which we fine-tune \mbert, and \textsc{Trg} the target language on which we evaluate it. 
\textsc{Src-X} is the average across all 17 target language with X $\neq$ \textsc{Src}
\colorbox{mediumspringgreen}{$\geq$ \textsc{\standardFineTune}}
\colorbox{lightgreen}{$<$  \textsc{\standardFineTune}}
\colorbox{pink}{$\leq$ -2 points}
\colorbox{brightlavender}{$\leq$ -5 points}
}
\label{tab:delta_pud_reinit_parsing}
\end{table*}

\begin{table*}[h!]
\footnotesize\centering
\scalebox{0.78}{ 
\begin{tabular}{lrrrrrrrr}
    \toprule
        & \multicolumn{8}{c}{\underline{\textit{\reinit \textit{of layers}}}} \\
     \textsc{Source - Target} &\textsc{\standardFineTune}  & $\Delta$ \textsc{0-1} & $\Delta$ \textsc{2-3} & $\Delta$ \textsc{4-5} & $\Delta$ \textsc{6-7} & $\Delta$ \textsc{8-9} &  $\Delta$ \textsc{10-11} \\
     &&\multicolumn{7}{c}{\textbf{\textit{POS}}}\\
\textsc{En - English} & 96.51 & \cellcolor{lightgreen}-0.30 & \cellcolor{lightgreen}-0.25 & \cellcolor{lightgreen}-0.40 & \cellcolor{lightgreen}-0.00 & \cellcolor{mediumspringgreen}0.05 & \cellcolor{mediumspringgreen}0.02 \\ 
\hdashline
\textsc{En - Arabic} & 70.20 & \cellcolor{pink}-3.63 & \cellcolor{lightgreen}-1.88 & \cellcolor{pink}-2.40 & \cellcolor{lightgreen}-1.26 & \cellcolor{lightgreen}-1.89 & \cellcolor{pink}-2.74 \\ 
\textsc{En - French} & 89.16 & \cellcolor{brightlavender}-9.68 & \cellcolor{pink}-2.09 & \cellcolor{lightgreen}-1.49 & \cellcolor{lightgreen}-1.03 & \cellcolor{mediumspringgreen}0.29 & \cellcolor{mediumspringgreen}0.59 \\ 
\textsc{En - German} & 89.32 & \cellcolor{brightlavender}-7.81 & \cellcolor{pink}-2.12 & \cellcolor{lightgreen}-1.27 & \cellcolor{lightgreen}-0.99 & \cellcolor{lightgreen}-0.46 & \cellcolor{lightgreen}-0.68 \\ 
\textsc{En - Turkish} & 71.67 & \cellcolor{brightlavender}-11.62 & \cellcolor{pink}-4.43 & \cellcolor{lightgreen}-1.48 & \cellcolor{lightgreen}-0.95 & \cellcolor{mediumspringgreen}0.04 & \cellcolor{lightgreen}-0.95 & \\
\textsc{En - Indo} & 71.44 & \cellcolor{brightlavender}-6.39 & \cellcolor{pink}-2.80 & \cellcolor{lightgreen}-1.74 & \cellcolor{lightgreen}-0.59 & \cellcolor{lightgreen}-0.41 & \cellcolor{lightgreen}-1.10 \\ 
\textsc{En - Russian} & 86.26 & \cellcolor{pink}-2.66 & \cellcolor{lightgreen}-0.94 & \cellcolor{lightgreen}-0.27 & \cellcolor{mediumspringgreen}0.13 & \cellcolor{mediumspringgreen}0.37 & \cellcolor{mediumspringgreen}0.62 \\ 
\textsc{En - Porthughese} & 86.51 & \cellcolor{brightlavender}-10.84 & \cellcolor{lightgreen}-1.83 & \cellcolor{lightgreen}-1.44 & \cellcolor{lightgreen}-0.81 & \cellcolor{lightgreen}-0.01 & \cellcolor{lightgreen}-0.14 \\ 
\textsc{En - Spanish} & 87.26 & \cellcolor{brightlavender}-8.09 & \cellcolor{lightgreen}-1.30 & \cellcolor{lightgreen}-1.36 & \cellcolor{lightgreen}-1.13 & \cellcolor{mediumspringgreen}0.20 & \cellcolor{mediumspringgreen}0.17 \\ 
\textsc{En - Finish} & 84.85 & \cellcolor{brightlavender}-20.00 & \cellcolor{brightlavender}-8.09 & \cellcolor{pink}-2.77 & \cellcolor{lightgreen}-0.97 & \cellcolor{lightgreen}-0.06 & \cellcolor{lightgreen}-0.86 \\ 
\textsc{En - Italian} & 91.35 & \cellcolor{brightlavender}-13.97 & \cellcolor{pink}-3.35 & \cellcolor{pink}-2.66 & \cellcolor{lightgreen}-1.34 & \cellcolor{lightgreen}-0.01 & \cellcolor{mediumspringgreen}0.27 \\ 
\textsc{En - Slovenian} & 89.64 & \cellcolor{brightlavender}-16.46 & \cellcolor{pink}-2.41 & \cellcolor{lightgreen}-1.09 & \cellcolor{lightgreen}-0.18 & \cellcolor{mediumspringgreen}0.34 & \cellcolor{mediumspringgreen}0.19 \\ 
\textsc{En - Czech} & 83.39 & \cellcolor{brightlavender}-19.62 & \cellcolor{pink}-3.93 & \cellcolor{lightgreen}-0.73 & \cellcolor{lightgreen}-0.56 & \cellcolor{mediumspringgreen}0.21 & \cellcolor{mediumspringgreen}0.29 \\ 
\textsc{En - Polish} & 81.45 & \cellcolor{brightlavender}-13.33 & \cellcolor{pink}-3.52 & \cellcolor{lightgreen}-1.19 & \cellcolor{lightgreen}-1.22 & \cellcolor{lightgreen}-0.50 & \cellcolor{lightgreen}-0.16 \\ 
\textsc{En - Hindi} & 65.43 & \cellcolor{brightlavender}-10.04 & \cellcolor{pink}-2.70 & \cellcolor{pink}-2.89 & \cellcolor{pink}-3.25 & \cellcolor{mediumspringgreen}3.00 & \cellcolor{mediumspringgreen}0.28 \\ 
\textsc{En - Chinese} & 67.89 & \cellcolor{pink}-3.04 & \cellcolor{pink}-2.82 & \cellcolor{pink}-3.59 & \cellcolor{lightgreen}-0.29 & \cellcolor{mediumspringgreen}0.66 & \cellcolor{mediumspringgreen}0.29 \\ 
\textsc{En - Japanese} & 48.86 & \cellcolor{pink}-2.19 & \cellcolor{mediumspringgreen}1.52 & \cellcolor{lightgreen}-1.51 & \cellcolor{lightgreen}-1.13 & \cellcolor{mediumspringgreen}1.42 & \cellcolor{mediumspringgreen}1.79 \\ 
\hdashline
\textsc{En - X (Mean)}  &  79.37 & \cellcolor{brightlavender} -8.94 & \cellcolor{pink} -2.49 & \cellcolor{lightgreen} -1.66 & \cellcolor{lightgreen} -0.88 & \cellcolor{mediumspringgreen} 0.20 & \cellcolor{lightgreen} -0.14 \\ 
 \midrule 
\textsc{Ru - Russian} & 96.90 & \cellcolor{lightgreen}-0.52 & \cellcolor{lightgreen}-0.55 & \cellcolor{lightgreen}-0.40 & \cellcolor{lightgreen}-0.07 & \cellcolor{mediumspringgreen}0.02 & \cellcolor{lightgreen}-0.03 & \\
\hdashline
\textsc{Ru - English} & 82.55 & \cellcolor{brightlavender}-20.72 & \cellcolor{brightlavender}-7.06 & \cellcolor{brightlavender}-5.01 & \cellcolor{pink}-3.93 & \cellcolor{mediumspringgreen}0.74 & \cellcolor{lightgreen}-1.57 \\ 
\textsc{Ru - Arabic} & 79.30 & \cellcolor{pink}-4.04 & \cellcolor{lightgreen}-1.48 & \cellcolor{pink}-2.06 & \cellcolor{mediumspringgreen}0.64 & \cellcolor{mediumspringgreen}0.01 & \cellcolor{mediumspringgreen}0.47 \\ 
\textsc{Ru - French} & 86.02 & \cellcolor{brightlavender}-18.66 & \cellcolor{pink}-4.64 & \cellcolor{pink}-4.10 & \cellcolor{brightlavender}-9.00 & \cellcolor{lightgreen}-0.13 & \cellcolor{lightgreen}-1.84 \\ 
\textsc{Ru - German} & 84.90 & \cellcolor{brightlavender}-12.50 & \cellcolor{pink}-4.80 & \cellcolor{pink}-2.79 & \cellcolor{pink}-3.90 & \cellcolor{mediumspringgreen}0.47 & \cellcolor{lightgreen}-1.82 \\ 
\textsc{Ru - Turkish} & 69.92 & \cellcolor{brightlavender}-15.20 & \cellcolor{pink}-2.06 & \cellcolor{lightgreen}-0.55 & \cellcolor{lightgreen}-1.41 & \cellcolor{lightgreen}-0.11 & \cellcolor{mediumspringgreen}0.68 &\\
\textsc{Ru - Indo} & 71.16 & \cellcolor{brightlavender}-8.33 & \cellcolor{pink}-3.44 & \cellcolor{lightgreen}-1.03 & \cellcolor{lightgreen}-0.56 & \cellcolor{lightgreen}-0.73 & \cellcolor{mediumspringgreen}0.15 \\ 
\textsc{Ru - Porthughese} & 84.24 & \cellcolor{brightlavender}-19.56 & \cellcolor{brightlavender}-7.15 & \cellcolor{pink}-3.00 & \cellcolor{brightlavender}-7.78 & \cellcolor{lightgreen}-0.15 & \cellcolor{pink}-2.08 \\ 
\textsc{Ru - SPanish} & 84.84 & \cellcolor{brightlavender}-13.64 & \cellcolor{pink}-4.09 & \cellcolor{pink}-2.66 & \cellcolor{brightlavender}-7.67 & \cellcolor{lightgreen}-0.35 & \cellcolor{pink}-2.48 \\ 
\textsc{Ru - Finish} & 81.08 & \cellcolor{brightlavender}-18.55 & \cellcolor{brightlavender}-5.42 & \cellcolor{lightgreen}-1.37 & \cellcolor{lightgreen}-1.00 & \cellcolor{lightgreen}-0.16 & \cellcolor{mediumspringgreen}0.02 \\ 
\textsc{Ru - Italian} & 85.56 & \cellcolor{brightlavender}-21.04 & \cellcolor{brightlavender}-5.11 & \cellcolor{pink}-3.41 & \cellcolor{brightlavender}-8.21 & \cellcolor{lightgreen}-0.20 & \cellcolor{pink}-3.36 \\ 
\textsc{Ru - Slovenian} & 85.37 & \cellcolor{brightlavender}-14.65 & \cellcolor{pink}-3.53 & \cellcolor{lightgreen}-1.72 & \cellcolor{lightgreen}-2.00 & \cellcolor{lightgreen}-0.15 & \cellcolor{lightgreen}-0.15 \\ 
\textsc{Ru - Czech} & 87.37 & \cellcolor{brightlavender}-8.43 & \cellcolor{lightgreen}-1.99 & \cellcolor{lightgreen}-0.71 & \cellcolor{lightgreen}-1.16 & \cellcolor{lightgreen}-0.50 & \cellcolor{lightgreen}-0.28 \\ 
\textsc{Ru - Polish} & 86.42 & \cellcolor{pink}-4.41 & \cellcolor{lightgreen}-1.89 & \cellcolor{lightgreen}-0.64 & \cellcolor{lightgreen}-0.44 & \cellcolor{lightgreen}-0.21 & \cellcolor{mediumspringgreen}0.09 \\ 
\textsc{Ru - Hindi} & 65.49 & \cellcolor{lightgreen}-1.16 & \cellcolor{mediumspringgreen}0.41 & \cellcolor{lightgreen}-1.49 & \cellcolor{pink}-2.17 & \cellcolor{mediumspringgreen}1.13 & \cellcolor{mediumspringgreen}3.20 \\ 
\textsc{Ru - Chinese} & 65.85 & \cellcolor{brightlavender}-5.12 & \cellcolor{lightgreen}-1.43 & \cellcolor{lightgreen}-0.32 & \cellcolor{lightgreen}-0.74 & \cellcolor{lightgreen}-0.13 & \cellcolor{lightgreen}-0.47 \\ 
\textsc{Ru - Japanese} & 46.91 & \cellcolor{lightgreen}-0.72 & \cellcolor{mediumspringgreen}2.16 & \cellcolor{mediumspringgreen}0.00 & \cellcolor{lightgreen}-1.30 & \cellcolor{mediumspringgreen}1.15 & \cellcolor{mediumspringgreen}1.12 \\ 
\hdashline
 \textsc{Ru - X (Mean)}  &  79.25 & \cellcolor{brightlavender} -10.08 & \cellcolor{pink} -2.83 & \cellcolor{lightgreen} -1.65 & \cellcolor{pink} -2.74 & \cellcolor{mediumspringgreen} 0.01 & \cellcolor{lightgreen} -0.45 \\ 
 \midrule 
\textsc{Ar - Arabic} & 79.28 & \cellcolor{lightgreen}-0.35 & \cellcolor{lightgreen}-0.49 & \cellcolor{lightgreen}-0.36 & \cellcolor{lightgreen}-0.19 & \cellcolor{lightgreen}-0.05 & \cellcolor{lightgreen}-0.00 \\ 
\hdashline
\textsc{Ar - English} & 63.26 & \cellcolor{pink}-3.32 & \cellcolor{lightgreen}-1.09 & \cellcolor{lightgreen}-1.72 & \cellcolor{lightgreen}-1.68 & \cellcolor{lightgreen}-1.03 & \cellcolor{lightgreen}-1.78 \\ 
\textsc{Ar - French} & 63.33 & \cellcolor{pink}-4.41 & \cellcolor{lightgreen}-1.53 & \cellcolor{lightgreen}-1.14 & \cellcolor{lightgreen}-1.30 & \cellcolor{lightgreen}-0.44 & \cellcolor{lightgreen}-0.92 \\ 
\textsc{Ar - German} & 63.23 & \cellcolor{pink}-4.95 & \cellcolor{pink}-2.97 & \cellcolor{lightgreen}-1.04 & \cellcolor{lightgreen}-1.58 & \cellcolor{lightgreen}-0.53 & \cellcolor{pink}-2.09 \\ 
\textsc{Ar - Turkish} & 60.99 & \cellcolor{brightlavender}-13.76 & \cellcolor{brightlavender}-8.74 & \cellcolor{pink}-2.86 & \cellcolor{pink}-4.49 & \cellcolor{lightgreen}-1.08 & \cellcolor{lightgreen}-1.88 & \\
\textsc{Ar - Indo} & 64.24 & \cellcolor{brightlavender}-5.11 & \cellcolor{pink}-3.43 & \cellcolor{lightgreen}-1.87 & \cellcolor{lightgreen}-0.58 & \cellcolor{lightgreen}-0.28 & \cellcolor{lightgreen}-0.63 \\ 
\textsc{Ar - Russian} & 74.52 & \cellcolor{pink}-4.01 & \cellcolor{pink}-2.37 & \cellcolor{pink}-2.40 & \cellcolor{lightgreen}-1.84 & \cellcolor{lightgreen}-1.69 & \cellcolor{pink}-2.03 \\ 
\textsc{Ar - Porthughese} & 67.28 & \cellcolor{brightlavender}-6.51 & \cellcolor{pink}-2.84 & \cellcolor{lightgreen}-1.30 & \cellcolor{lightgreen}-1.23 & \cellcolor{mediumspringgreen}0.04 & \cellcolor{lightgreen}-0.96 \\ 
\textsc{Ar - SPanish} & 64.84 & \cellcolor{pink}-3.08 & \cellcolor{lightgreen}-0.51 & \cellcolor{lightgreen}-0.74 & \cellcolor{lightgreen}-0.48 & \cellcolor{mediumspringgreen}0.02 & \cellcolor{lightgreen}-0.14 \\ 
\textsc{Ar - Finish} & 64.28 & \cellcolor{brightlavender}-19.72 & \cellcolor{brightlavender}-8.32 & \cellcolor{pink}-3.72 & \cellcolor{pink}-2.56 & \cellcolor{lightgreen}-1.64 & \cellcolor{pink}-3.03 \\ 
\textsc{Ar - Italian} & 63.55 & \cellcolor{pink}-4.25 & \cellcolor{lightgreen}-1.60 & \cellcolor{lightgreen}-0.94 & \cellcolor{lightgreen}-1.15 & \cellcolor{mediumspringgreen}0.14 & \cellcolor{lightgreen}-0.64 \\ 
\textsc{Ar - Slovenian} & 68.06 & \cellcolor{brightlavender}-12.21 & \cellcolor{pink}-4.31 & \cellcolor{pink}-2.17 & \cellcolor{lightgreen}-1.85 & \cellcolor{mediumspringgreen}0.68 & \cellcolor{lightgreen}-1.81 \\ 
\textsc{Ar - Czech} & 72.65 & \cellcolor{brightlavender}-13.57 & \cellcolor{pink}-3.14 & \cellcolor{lightgreen}-1.88 & \cellcolor{lightgreen}-1.77 & \cellcolor{lightgreen}-1.35 & \cellcolor{lightgreen}-1.57 \\ 
\textsc{Ar - Polish} & 75.00 & \cellcolor{brightlavender}-8.87 & \cellcolor{pink}-2.94 & \cellcolor{lightgreen}-1.46 & \cellcolor{lightgreen}-0.62 & \cellcolor{lightgreen}-1.00 & \cellcolor{lightgreen}-1.37 \\ 
\textsc{Ar - Hindi} & 62.29 & \cellcolor{brightlavender}-7.31 & \cellcolor{brightlavender}-6.07 & \cellcolor{pink}-2.42 & \cellcolor{lightgreen}-1.26 & \cellcolor{mediumspringgreen}0.19 & \cellcolor{lightgreen}-1.72 \\ 
\textsc{Ar - Chinese} & 56.51 & \cellcolor{brightlavender}-5.02 & \cellcolor{pink}-4.94 & \cellcolor{pink}-2.10 & \cellcolor{lightgreen}-1.35 & \cellcolor{lightgreen}-1.02 & \cellcolor{lightgreen}-1.77 \\ 
\textsc{Ar - Japanese} & 47.06 & \cellcolor{pink}-3.34 & \cellcolor{pink}-3.34 & \cellcolor{lightgreen}-0.65 & \cellcolor{lightgreen}-0.89 & \cellcolor{lightgreen}-1.54 & \cellcolor{lightgreen}-0.35 \\ 
\hdashline
\textsc{Ar - X (Mean)}  &  64.81 & \cellcolor{brightlavender} -6.73 & \cellcolor{pink} -3.50 & \cellcolor{lightgreen} -1.63 & \cellcolor{lightgreen} -1.56 & \cellcolor{lightgreen} -0.73 & \cellcolor{lightgreen} $-1.29$ \\ 


\bottomrule

\end{tabular}
}
\caption{POS tagging Relative Zero shot Cross-Lingual performance of \mbert with \reinit (section \ref{sec:controlled_exp}) on pairs of consecutive layers compared to \mbert without any random-initialization (\standardFineTune). In \textsc{Src - Trg}, \textsc{Src} indicates the source language on which we fine-tune \mbert, and \textsc{Trg} the target language on which we evaluate it. 
\textsc{Src-X} is the average across all 17 target language with $X\neq$ \textsc{Src}.
\colorbox{mediumspringgreen}{$ \geq $ \textsc{\standardFineTune}}
\colorbox{lightgreen}{$ < $  \textsc{\standardFineTune}}
\colorbox{pink}{$ \leq $ -2 points}
\colorbox{brightlavender}{$ \leq $ -5 points }
}
\label{tab:delta_pud_reinit_pos}
\end{table*}

\begin{table*}[h!]
\footnotesize\centering
\scalebox{0.78}{ 
\begin{tabular}{lrrrrrrrr}
    \toprule
        & \multicolumn{8}{c}{\underline{\textit{\reinit \textit{of layers}}}} \\
     Source - Target &\textsc{\standardFineTune}  & $\Delta$ \textsc{0-1} & $\Delta$ \textsc{2-3} & $\Delta$ \textsc{4-5} & $\Delta$ \textsc{6-7} & $\Delta$ \textsc{8-9} &  $\Delta$ \textsc{10-11} \\
     &&\multicolumn{7}{c}{\textbf{\textit{NER}}}\\
 
\textsc{EN - English} & 83.27 & \cellcolor{pink}-2.64 & \cellcolor{pink}-2.12 & \cellcolor{lightgreen}-1.41 & \cellcolor{lightgreen}-0.61 & \cellcolor{lightgreen}-0.21 & \cellcolor{lightgreen}-0.14 \\ 
\hdashline
\textsc{EN - French} & 76.20 & \cellcolor{pink}-4.41 & \cellcolor{pink}-2.72 & \cellcolor{pink}-2.09 & \cellcolor{lightgreen}-0.30 & \cellcolor{mediumspringgreen}0.51 & \cellcolor{mediumspringgreen}0.08 \\ 
\textsc{EN - German} & 75.58 & \cellcolor{brightlavender}-8.25 & \cellcolor{pink}-4.65 & \cellcolor{pink}-2.50 & \cellcolor{lightgreen}-0.40 & \cellcolor{mediumspringgreen}0.06 & \cellcolor{mediumspringgreen}0.26 \\ 
\textsc{EN - Turkish} & 66.23 & \cellcolor{brightlavender}-8.71 & \cellcolor{brightlavender}-6.57 & \cellcolor{pink}-2.16 & \cellcolor{lightgreen}-1.01 & \cellcolor{mediumspringgreen}0.51 & \cellcolor{mediumspringgreen}0.51 & \\
\textsc{EN - Indo} & 50.24 & \cellcolor{pink}-2.94 & \cellcolor{lightgreen}-1.43 & \cellcolor{pink}-2.54 & \cellcolor{mediumspringgreen}2.49 & \cellcolor{lightgreen}-0.70 & \cellcolor{mediumspringgreen}0.82 \\ 
\textsc{EN - Porthughese} & 76.09 & \cellcolor{pink}-4.66 & \cellcolor{lightgreen}-0.88 & \cellcolor{lightgreen}-1.16 & \cellcolor{lightgreen}-0.57 & \cellcolor{mediumspringgreen}0.62 & \cellcolor{lightgreen}-0.70 \\ 
\textsc{EN - SPanish} & 67.00 & \cellcolor{lightgreen}-0.99 & \cellcolor{mediumspringgreen}4.37 & \cellcolor{mediumspringgreen}2.03 & \cellcolor{lightgreen}-1.69 & \cellcolor{mediumspringgreen}1.57 & \cellcolor{lightgreen}-1.38 \\ 
\textsc{EN - Finish} & 75.61 & \cellcolor{brightlavender}-11.89 & \cellcolor{pink}-4.47 & \cellcolor{pink}-2.29 & \cellcolor{mediumspringgreen}0.63 & \cellcolor{mediumspringgreen}0.54 & \cellcolor{lightgreen}-0.37 \\ 
\textsc{EN - Italian} & 78.48 & \cellcolor{brightlavender}-6.65 & \cellcolor{pink}-3.64 & \cellcolor{pink}-3.08 & \cellcolor{lightgreen}-1.32 & \cellcolor{lightgreen}-0.30 & \cellcolor{lightgreen}-0.28 \\ 
\textsc{EN - Slovenian} & 72.80 & \cellcolor{brightlavender}-10.37 & \cellcolor{pink}-2.96 & \cellcolor{pink}-3.11 & \cellcolor{lightgreen}-0.36 & \cellcolor{mediumspringgreen}0.10 & \cellcolor{lightgreen}-0.72 \\ 
\textsc{EN - Czech} & 76.90 & \cellcolor{brightlavender}-8.02 & \cellcolor{brightlavender}-6.81 & \cellcolor{pink}-3.17 & \cellcolor{mediumspringgreen}0.09 & \cellcolor{mediumspringgreen}1.00 & \cellcolor{mediumspringgreen}0.39 \\ 
\textsc{EN - Russian} & 60.20 & \cellcolor{brightlavender}-5.87 & \cellcolor{brightlavender}-6.65 & \cellcolor{brightlavender}-5.71 & \cellcolor{pink}-2.82 & \cellcolor{lightgreen}-0.82 & \cellcolor{lightgreen}-0.37 \\ 
\textsc{EN - Arabic} & 39.15 & \cellcolor{brightlavender}-8.98 & \cellcolor{brightlavender}-5.31 & \cellcolor{lightgreen}-1.97 & \cellcolor{mediumspringgreen}1.56 & \cellcolor{mediumspringgreen}0.31 & \cellcolor{lightgreen}-0.98 \\ 
\textsc{EN - Polish} & 77.20 & \cellcolor{brightlavender}-8.32 & \cellcolor{brightlavender}-5.53 & \cellcolor{pink}-3.05 & \cellcolor{lightgreen}-0.06 & \cellcolor{mediumspringgreen}0.67 & \cellcolor{mediumspringgreen}0.09 \\ 
\textsc{EN - Hindi} & 60.61 & \cellcolor{brightlavender}-12.08 & \cellcolor{brightlavender}-13.88 & \cellcolor{brightlavender}-9.23 & \cellcolor{lightgreen}-0.91 & \cellcolor{lightgreen}-1.25 & \cellcolor{mediumspringgreen}2.08 \\ 
\textsc{EN - Chinese} & 37.74 & \cellcolor{brightlavender}-13.68 & \cellcolor{brightlavender}-6.49 & \cellcolor{pink}-4.59 & \cellcolor{pink}-2.41 & \cellcolor{brightlavender}-5.23 & \cellcolor{lightgreen}-1.00 \\ 
\textsc{EN - Japanese} & 25.19 & \cellcolor{brightlavender}-11.40 & \cellcolor{brightlavender}-7.54 & \cellcolor{pink}-4.67 & \cellcolor{pink}-2.53 & \cellcolor{pink}-3.45 & \cellcolor{lightgreen}-0.23 \\ 
\hdashline
\textsc{EN - X (mean)}  &  64.17 & \cellcolor{brightlavender} -8.28 & \cellcolor{brightlavender} -5.09 & \cellcolor{pink} -3.07 & \cellcolor{lightgreen} -0.79 & \cellcolor{lightgreen} -0.47 & \cellcolor{lightgreen} -0.13 \\ 

\midrule
\textsc{Ru - Russian} & 88.20 & \cellcolor{pink}-2.08 & \cellcolor{pink}-2.13 & \cellcolor{lightgreen}-1.52 & \cellcolor{lightgreen}-0.64 & \cellcolor{lightgreen}-0.33 & \cellcolor{lightgreen}-0.13 \\ 
\textsc{Ru - English} & 56.62 & \cellcolor{brightlavender}-13.83 & \cellcolor{brightlavender}-8.52 & \cellcolor{pink}-4.70 & \cellcolor{lightgreen}-1.50 & \cellcolor{lightgreen}-0.76 & \cellcolor{mediumspringgreen}1.38 \\ 
\textsc{Ru - French} & 67.35 & \cellcolor{brightlavender}-18.45 & \cellcolor{brightlavender}-9.70 & \cellcolor{pink}-4.32 & \cellcolor{lightgreen}-1.76 & \cellcolor{lightgreen}-1.77 & \cellcolor{mediumspringgreen}2.29 \\ 
\textsc{Ru - German} & 69.23 & \cellcolor{brightlavender}-13.94 & \cellcolor{brightlavender}-9.01 & \cellcolor{brightlavender}-5.80 & \cellcolor{pink}-2.98 & \cellcolor{lightgreen}-1.65 & \cellcolor{mediumspringgreen}0.40 \\ 
\textsc{Ru - Turkish} & 63.64 & \cellcolor{brightlavender}-18.52 & \cellcolor{brightlavender}-10.06 & \cellcolor{brightlavender}-6.01 & \cellcolor{pink}-4.16 & \cellcolor{lightgreen}-0.67 & \cellcolor{lightgreen}-0.27 & \\ 
\textsc{Ru - Indo} & 41.92 & \cellcolor{brightlavender}-10.29 & \cellcolor{brightlavender}-7.20 & \cellcolor{brightlavender}-5.19 & \cellcolor{lightgreen}-1.20 & \cellcolor{lightgreen}-1.91 & \cellcolor{mediumspringgreen}0.50 \\ 
\textsc{Ru - Porthughese} & 67.33 & \cellcolor{brightlavender}-21.23 & \cellcolor{brightlavender}-8.27 & \cellcolor{brightlavender}-8.84 & \cellcolor{pink}-2.83 & \cellcolor{lightgreen}-1.83 & \cellcolor{mediumspringgreen}1.51 \\ 
\textsc{Ru - SPanish} & 69.15 & \cellcolor{brightlavender}-16.74 & \cellcolor{brightlavender}-10.00 & \cellcolor{brightlavender}-8.16 & \cellcolor{brightlavender}-5.80 & \cellcolor{lightgreen}-1.66 & \cellcolor{mediumspringgreen}0.26 \\ 
\textsc{Ru - Finish} & 73.03 & \cellcolor{brightlavender}-17.17 & \cellcolor{brightlavender}-8.70 & \cellcolor{brightlavender}-5.88 & \cellcolor{pink}-2.12 & \cellcolor{mediumspringgreen}0.86 & \cellcolor{mediumspringgreen}1.48 \\ 
\textsc{Ru - Italian} & 70.05 & \cellcolor{brightlavender}-19.47 & \cellcolor{brightlavender}-9.54 & \cellcolor{brightlavender}-6.90 & \cellcolor{pink}-3.06 & \cellcolor{mediumspringgreen}0.73 & \cellcolor{mediumspringgreen}1.04 \\ 
\textsc{Ru - Slovenian} & 71.18 & \cellcolor{brightlavender}-12.02 & \cellcolor{brightlavender}-9.48 & \cellcolor{pink}-3.61 & \cellcolor{lightgreen}-0.70 & \cellcolor{mediumspringgreen}1.16 & \cellcolor{mediumspringgreen}2.14 \\ 
\textsc{Ru - Czech} & 74.87 & \cellcolor{brightlavender}-17.93 & \cellcolor{brightlavender}-10.59 & \cellcolor{brightlavender}-6.34 & \cellcolor{pink}-4.02 & \cellcolor{mediumspringgreen}0.17 & \cellcolor{lightgreen}-0.23 \\ 
\textsc{Ru - Arabic} & 38.63 & \cellcolor{brightlavender}-8.67 & \cellcolor{brightlavender}-6.81 & \cellcolor{lightgreen}-0.13 & \cellcolor{lightgreen}-0.65 & \cellcolor{lightgreen}-1.34 & \cellcolor{lightgreen}-0.29 \\ 
\textsc{Ru - Polish} & 75.16 & \cellcolor{brightlavender}-15.38 & \cellcolor{brightlavender}-7.97 & \cellcolor{brightlavender}-6.33 & \cellcolor{pink}-3.07 & \cellcolor{lightgreen}-0.63 & \cellcolor{mediumspringgreen}1.34 \\ 
\textsc{Ru - Hindi} & 58.01 & \cellcolor{brightlavender}-19.60 & \cellcolor{brightlavender}-12.36 & \cellcolor{brightlavender}-6.18 & \cellcolor{mediumspringgreen}0.93 & \cellcolor{lightgreen}-1.64 & \cellcolor{mediumspringgreen}1.17 \\ 
\textsc{Ru - Chinese} & 43.86 & \cellcolor{brightlavender}-23.73 & \cellcolor{brightlavender}-11.68 & \cellcolor{brightlavender}-6.80 & \cellcolor{pink}-4.27 & \cellcolor{pink}-4.13 & \cellcolor{brightlavender}-6.01 \\ 
\textsc{Ru - Japanese} & 30.79 & \cellcolor{brightlavender}-16.80 & \cellcolor{brightlavender}-11.29 & \cellcolor{brightlavender}-5.26 & \cellcolor{pink}-2.77 & \cellcolor{pink}-3.99 & \cellcolor{brightlavender}-6.91 \\ 
\hdashline
 \textsc{Ru - X (Mean)} & 62.13 & \cellcolor{brightlavender} -15.85 & \cellcolor{brightlavender} -9.36 & \cellcolor{brightlavender} -5.50 & \cellcolor{pink} -2.44 & \cellcolor{lightgreen} -1.16 & \cellcolor{lightgreen} -0.06 \\ 
 \midrule 
\textsc{Ar - Arabic} & 87.97 & \cellcolor{pink}-2.37 & \cellcolor{pink}-2.11 & \cellcolor{lightgreen}-0.96 & \cellcolor{lightgreen}-0.39 & \cellcolor{lightgreen}-0.15 & \cellcolor{mediumspringgreen}0.21 \\ 
\textsc{Ar - French} & 75.21 & \cellcolor{brightlavender}-18.71 & \cellcolor{brightlavender}-8.31 & \cellcolor{pink}-3.76 & \cellcolor{lightgreen}-0.19 & \cellcolor{mediumspringgreen}0.82 & \cellcolor{mediumspringgreen}1.07 \\ 
\textsc{Ar - German} & 74.24 & \cellcolor{brightlavender}-15.25 & \cellcolor{brightlavender}-7.19 & \cellcolor{pink}-3.72 & \cellcolor{lightgreen}-1.38 & \cellcolor{lightgreen}-0.04 & \cellcolor{mediumspringgreen}0.27 \\ 
\textsc{Ar - Turkish} & 68.45 & \cellcolor{brightlavender}-14.89 & \cellcolor{brightlavender}-8.65 & \cellcolor{pink}-2.78 & \cellcolor{lightgreen}-0.30 & \cellcolor{mediumspringgreen}0.98 & \cellcolor{mediumspringgreen}1.90 & \\ 
\textsc{Ar - Indo} & 54.65 & \cellcolor{brightlavender}-13.86 & \cellcolor{brightlavender}-10.95 & \cellcolor{brightlavender}-8.53 & \cellcolor{pink}-4.66 & \cellcolor{pink}-2.82 & \cellcolor{mediumspringgreen}0.09 \\ 
\textsc{Ar - Porthughese} & 74.67 & \cellcolor{brightlavender}-20.42 & \cellcolor{brightlavender}-10.54 & \cellcolor{pink}-3.17 & \cellcolor{lightgreen}-1.59 & \cellcolor{mediumspringgreen}0.10 & \cellcolor{mediumspringgreen}1.28 \\ 
\textsc{Ar - SPanish} & 74.88 & \cellcolor{brightlavender}-18.16 & \cellcolor{brightlavender}-12.18 & \cellcolor{pink}-3.06 & \cellcolor{lightgreen}-1.95 & \cellcolor{mediumspringgreen}0.52 & \cellcolor{mediumspringgreen}0.63 \\ 
\textsc{Ar - Finish} & 78.01 & \cellcolor{brightlavender}-18.79 & \cellcolor{brightlavender}-8.84 & \cellcolor{pink}-4.30 & \cellcolor{pink}-2.03 & \cellcolor{lightgreen}-0.30 & \cellcolor{mediumspringgreen}0.19 \\ 
\textsc{Ar - Italian} & 75.76 & \cellcolor{brightlavender}-16.37 & \cellcolor{brightlavender}-7.73 & \cellcolor{pink}-3.98 & \cellcolor{lightgreen}-1.49 & \cellcolor{lightgreen}-0.06 & \cellcolor{mediumspringgreen}0.74 \\ 
\textsc{Ar - Slovenian} & 63.08 & \cellcolor{brightlavender}-11.13 & \cellcolor{brightlavender}-5.49 & \cellcolor{mediumspringgreen}4.79 & \cellcolor{mediumspringgreen}0.88 & \cellcolor{mediumspringgreen}2.17 & \cellcolor{mediumspringgreen}0.79 \\ 
\textsc{Ar - Czech} & 74.70 & \cellcolor{brightlavender}-21.93 & \cellcolor{brightlavender}-10.95 & \cellcolor{brightlavender}-5.84 & \cellcolor{pink}-2.42 & \cellcolor{lightgreen}-1.36 & \cellcolor{mediumspringgreen}0.09 \\ 
\textsc{Ar - Russian} & 45.51 & \cellcolor{brightlavender}-7.59 & \cellcolor{brightlavender}-5.81 & \cellcolor{pink}-2.63 & \cellcolor{mediumspringgreen}0.15 & \cellcolor{lightgreen}-0.22 & \cellcolor{mediumspringgreen}0.47 \\ 
\textsc{Ar - English} & 57.94 & \cellcolor{brightlavender}-12.79 & \cellcolor{brightlavender}-6.03 & \cellcolor{pink}-4.57 & \cellcolor{lightgreen}-0.32 & \cellcolor{mediumspringgreen}0.29 & \cellcolor{mediumspringgreen}1.65 \\ 
\textsc{Ar - Polish} & 77.29 & \cellcolor{brightlavender}-20.61 & \cellcolor{brightlavender}-9.47 & \cellcolor{brightlavender}-5.93 & \cellcolor{pink}-2.64 & \cellcolor{lightgreen}-1.09 & \cellcolor{lightgreen}-0.19 \\ 
\textsc{Ar - Hindi} & 65.31 & \cellcolor{brightlavender}-14.95 & \cellcolor{brightlavender}-9.12 & \cellcolor{pink}-3.84 & \cellcolor{lightgreen}-1.48 & \cellcolor{mediumspringgreen}0.72 & \cellcolor{mediumspringgreen}0.98 \\ 
\textsc{Ar - Chinese} & 45.88 & \cellcolor{brightlavender}-25.72 & \cellcolor{brightlavender}-10.67 & \cellcolor{pink}-3.99 & \cellcolor{lightgreen}-1.41 & \cellcolor{pink}-2.72 & \cellcolor{mediumspringgreen}0.57 \\ 
\textsc{Ar - Japanese} & 24.75 & \cellcolor{brightlavender}-14.66 & \cellcolor{brightlavender}-5.19 & \cellcolor{pink}-3.82 & \cellcolor{lightgreen}-0.99 & \cellcolor{lightgreen}-1.17 & \cellcolor{mediumspringgreen}1.50 \\ 
\hdashline
 \textsc{Ar - X (Mean)} &  65.59 & \cellcolor{brightlavender} -16.10 & \cellcolor{brightlavender} -8.42 & \cellcolor{pink} -3.73 & \cellcolor{lightgreen} -1.40 & \cellcolor{lightgreen} -0.25 & \cellcolor{mediumspringgreen} 0.67 \\

\bottomrule

\end{tabular}
}
\caption{NER (F1 score) Relative Zero shot Cross-Lingual performance of \mbert with \reinit (section \ref{sec:controlled_exp}) on pairs of consecutive layers compared to \mbert without any random-initialization (\standardFineTune). In \textsc{Src - Trg}, \textsc{Src} indicates the source language on which we fine-tune \mbert, and \textsc{Trg} the target language on which we evaluate it. 
\textsc{Src-X} is the average across all 17 target language with X $\neq$ \textsc{Src}
\colorbox{mediumspringgreen}{$\geq$ \textsc{\standardFineTune}}
\colorbox{lightgreen}{$<$  \textsc{\standardFineTune}}
\colorbox{pink}{$\leq$ -2 points}
\colorbox{brightlavender}{$\leq$ -5 points }
}
\label{tab:delta_pud_reinit_ner}
\end{table*}

\newpage
\textcolor{white}{.}

\newpage
\textcolor{white}{.}
\newpage
\textcolor{white}{.}
\newpage
\textcolor{white}{.}
\newpage
\textcolor{white}{.}

\begin{figure*}[ht]
\centering
\subfloat[]{
\includegraphics[width=1\columnwidth]{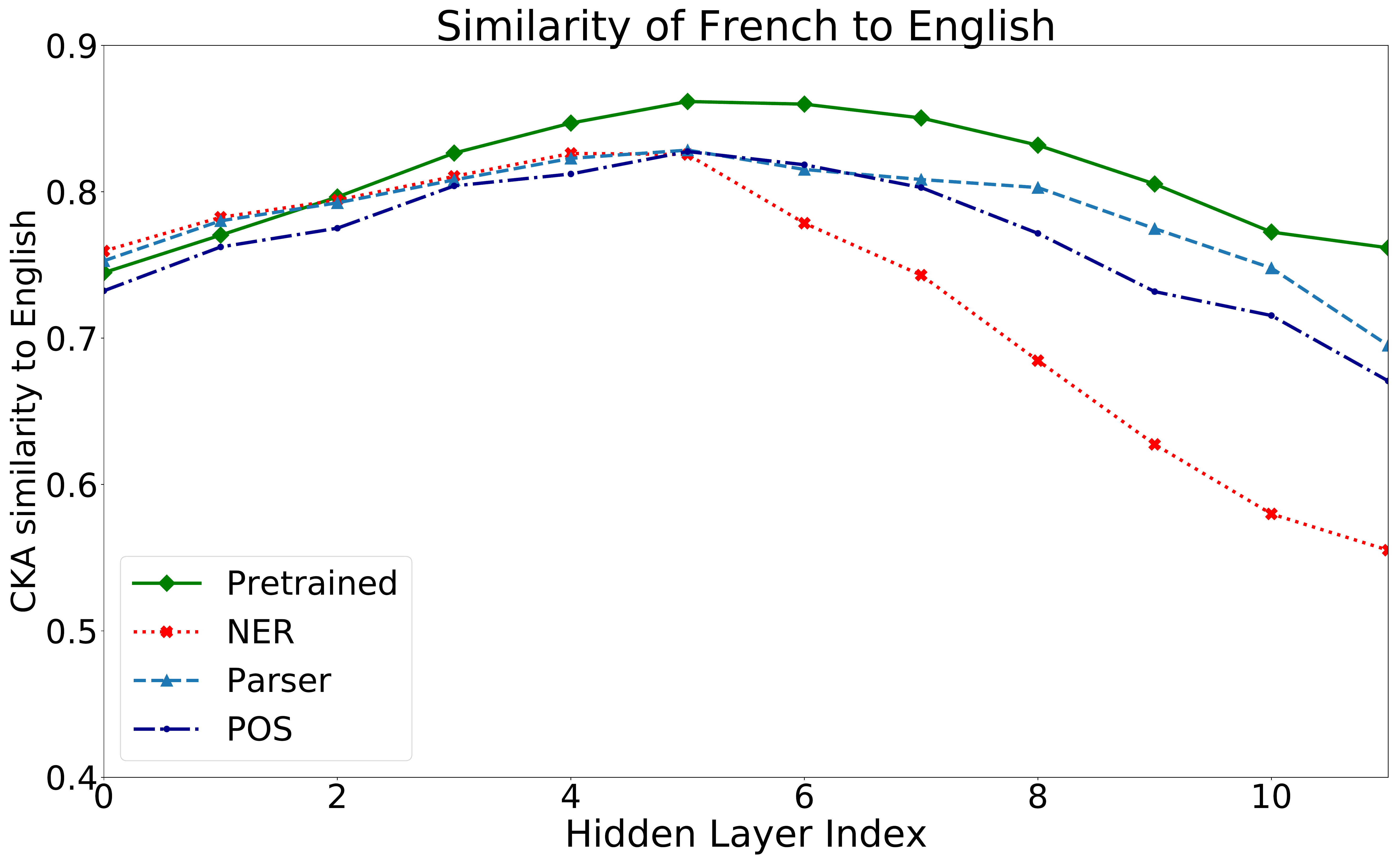}
\label{fig:vehicle_speed}
}
\subfloat[]{
\includegraphics[width=1\columnwidth]{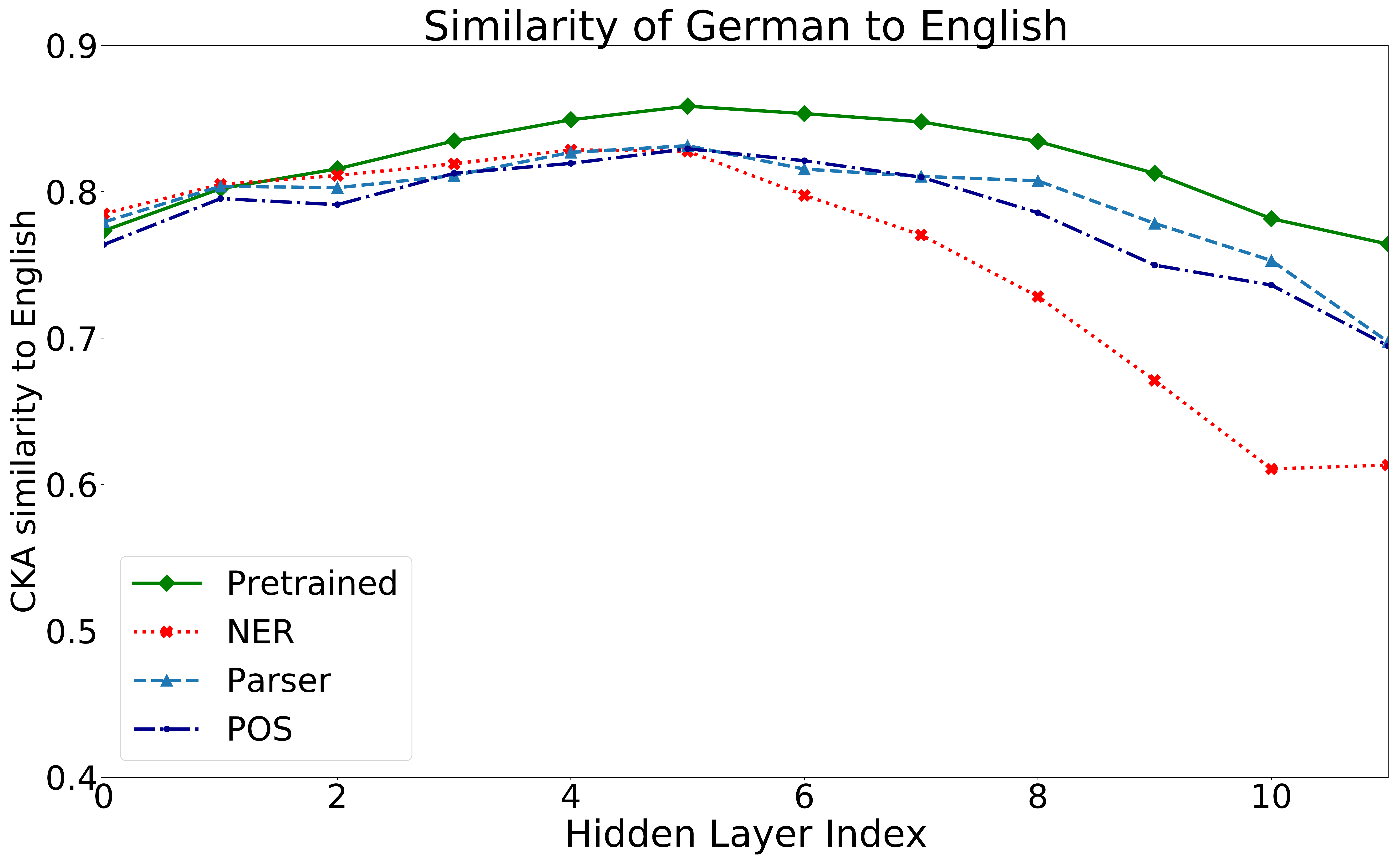}
\label{fig:accessories_dollar}
} \\

\subfloat[]{
\includegraphics[width=1\columnwidth]{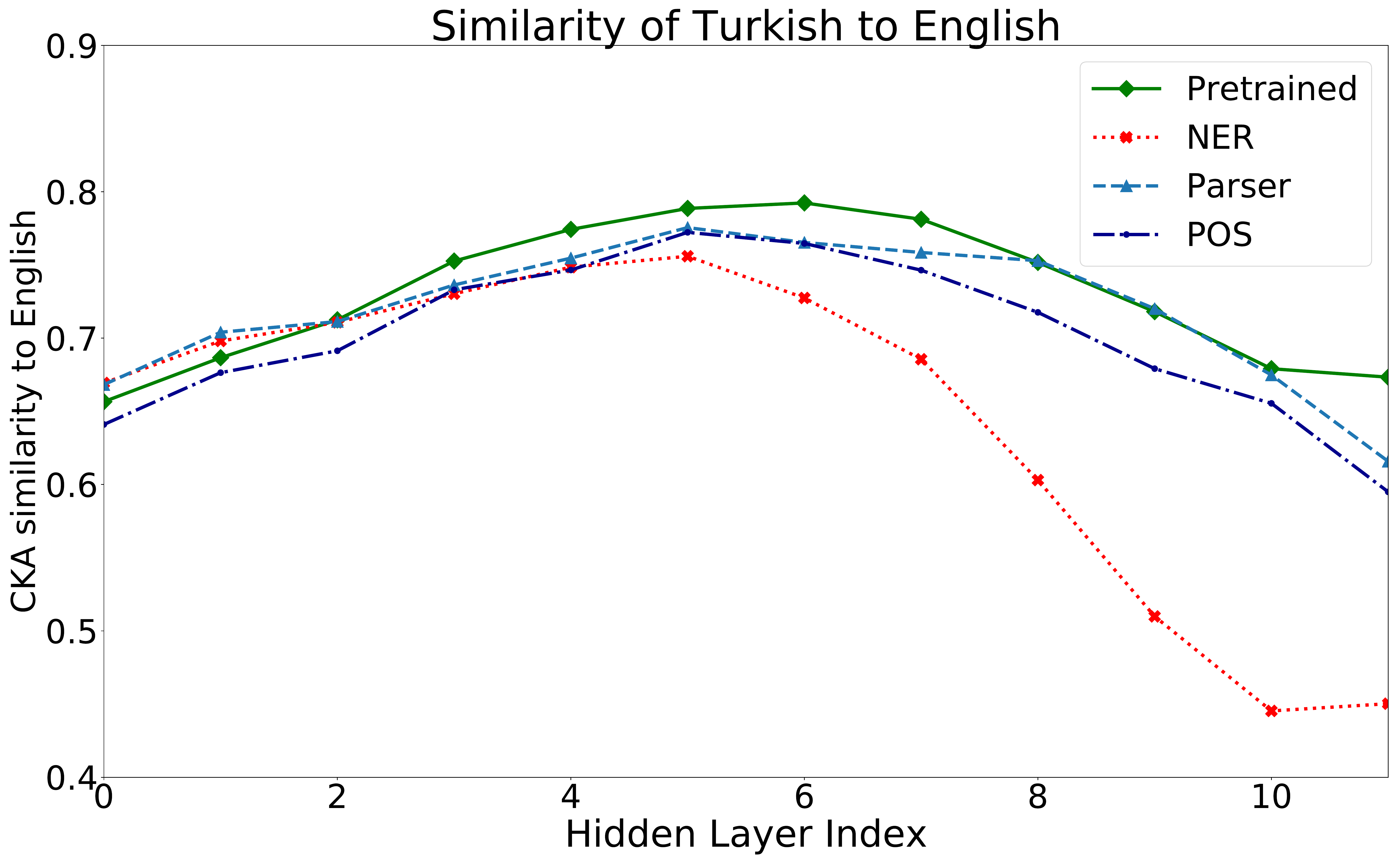}
\label{fig:vehicle_speed}
}
\subfloat[]{
\includegraphics[width=1\columnwidth]{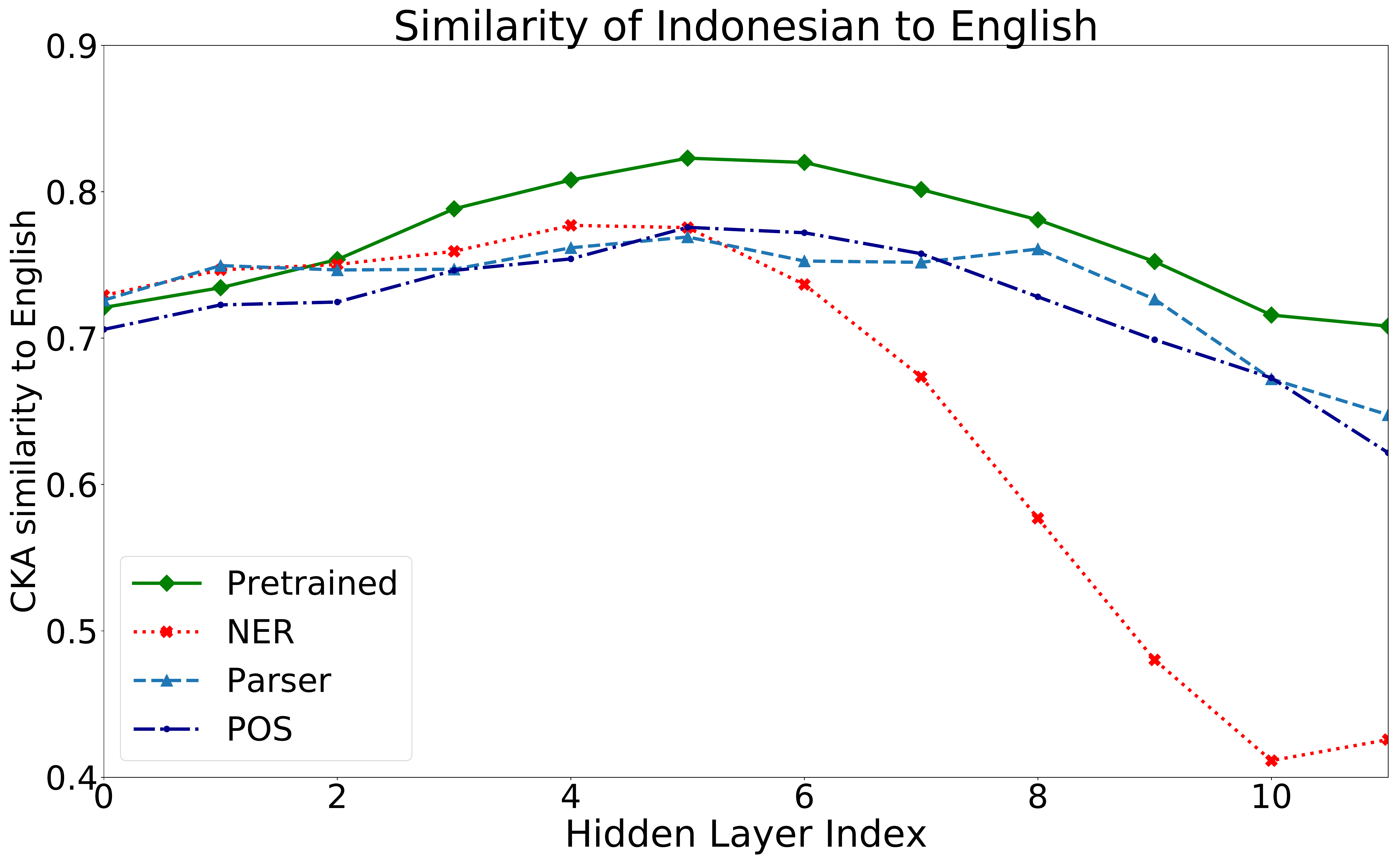}
\label{fig:accessories_dollar}
} \\

\subfloat[]{
\includegraphics[width=1\columnwidth]{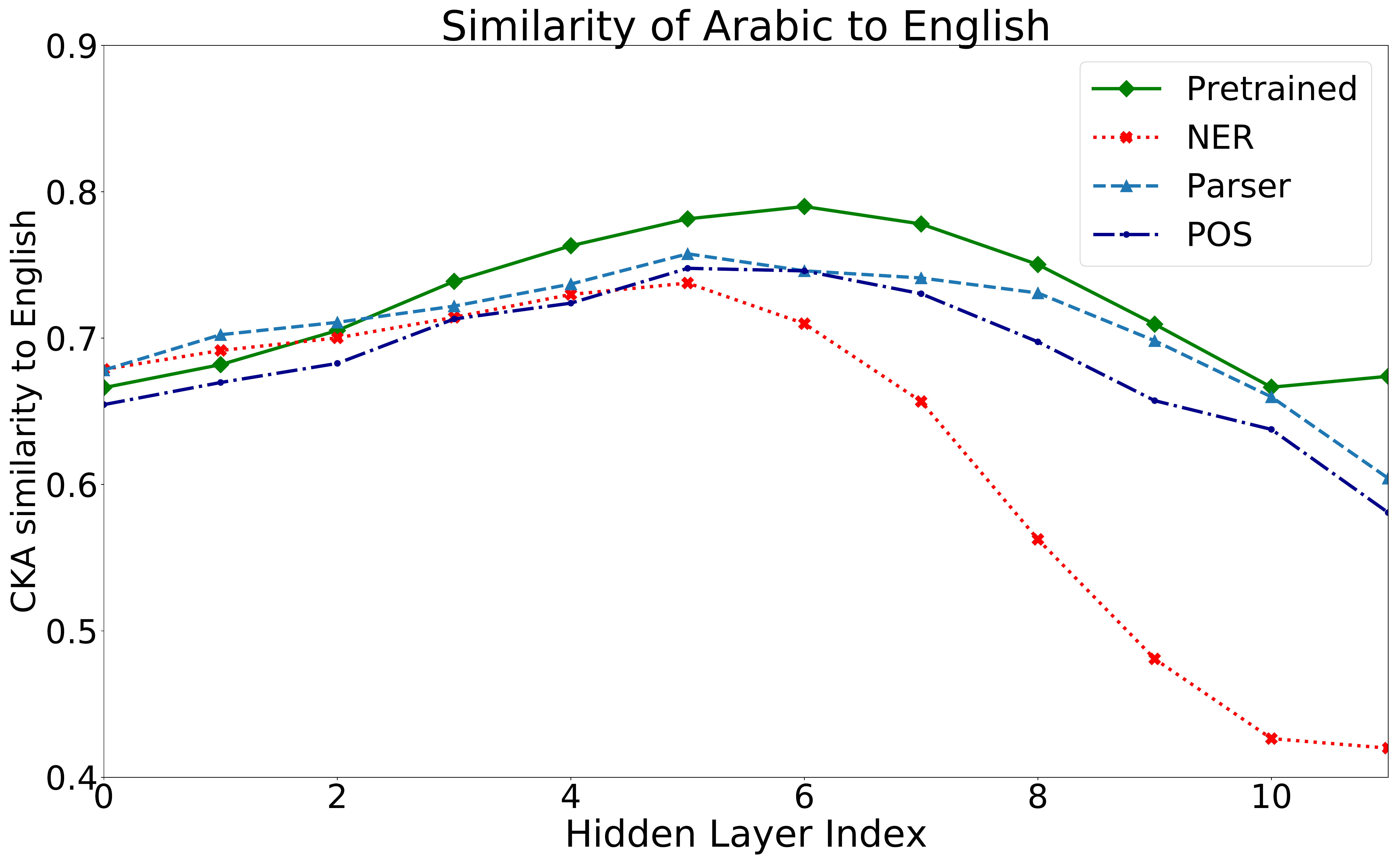}
\label{fig:vehicle_speed}
}

\caption{Cross-Lingual similarity (CKA) similarity (\S \ref{sec:alignement}) of hidden representations of a source language (English) sentences with a target language sentences on fine-tuned and pretrained \mbert. The higher the CKA value the greater the similarity.}
\label{fig:pretrained_tuned_models}
\end{figure*}


\newpage


\begin{figure*}[ht]
\centering
\subfloat[]{
\includegraphics[width=8.0cm]{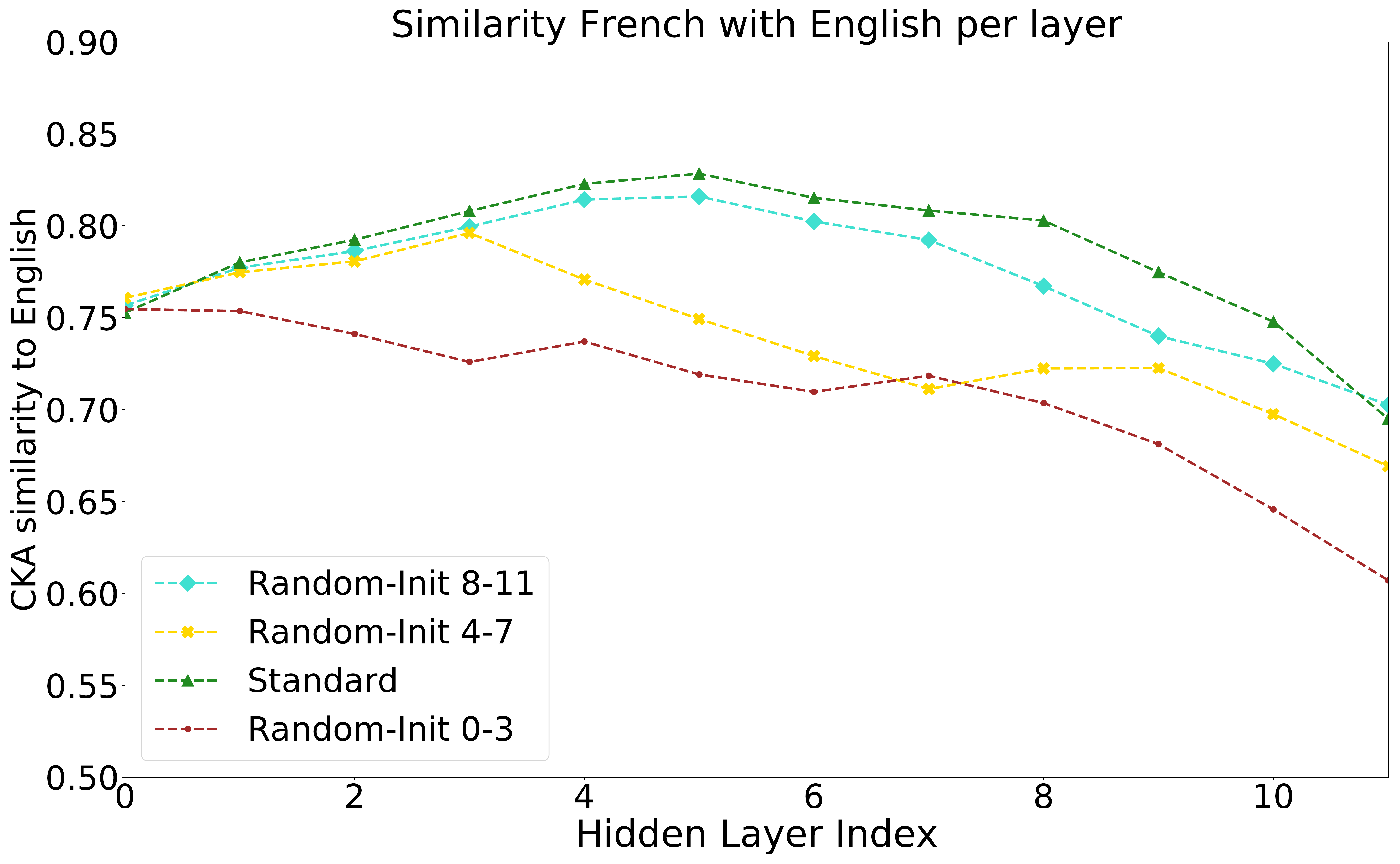}

\label{fig:vehicle_speed}
}
\subfloat[]{
\includegraphics[width=8.0cm]{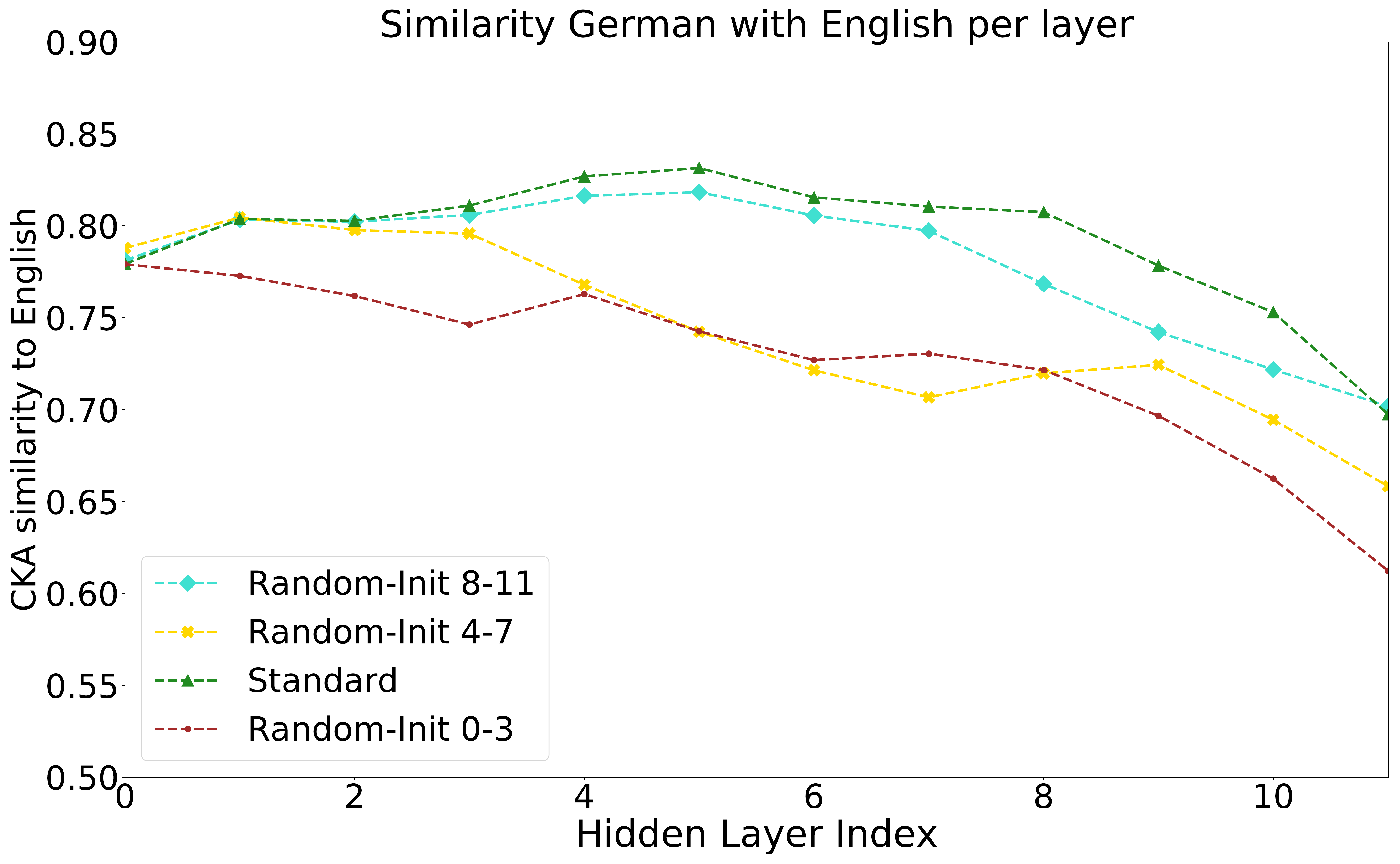}

\label{fig:vehicle_speed}
}\\
\subfloat[]{
\includegraphics[width=8.0cm]{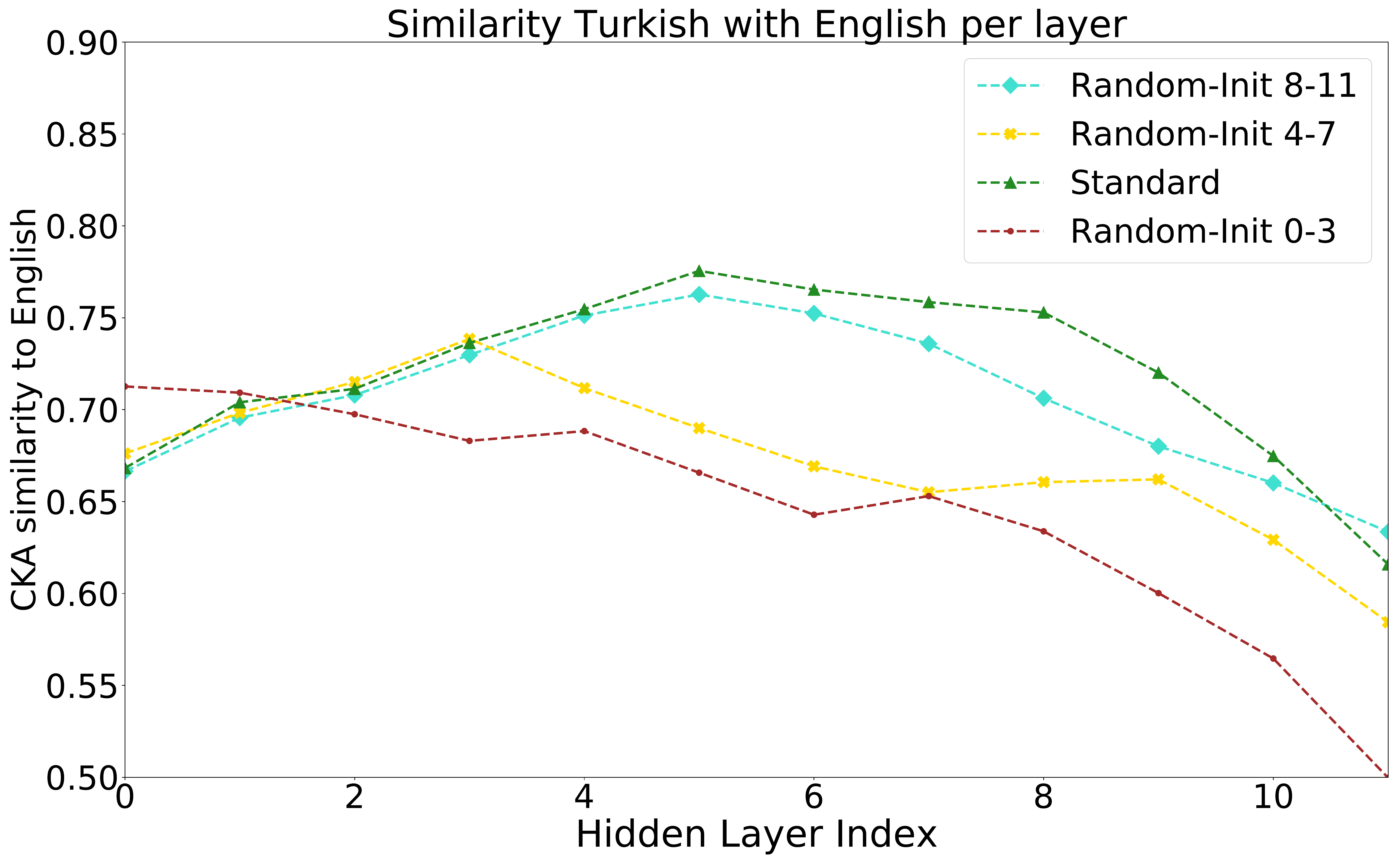}
\label{fig:vehicle_speed}
}
\subfloat[]{
\includegraphics[width=8.0cm]{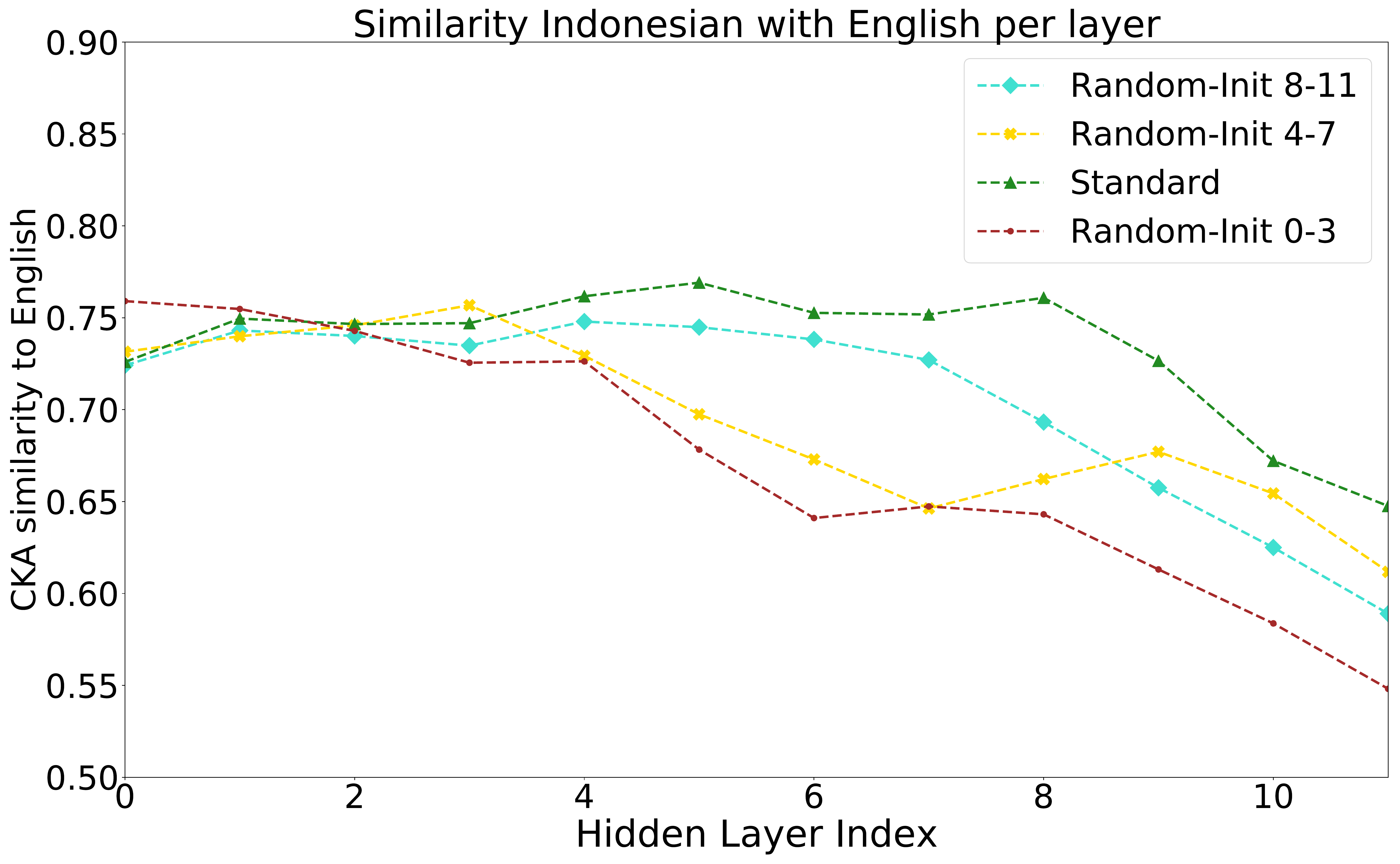}
\label{fig:vehicle_speed}
}
\\
\subfloat[]{
\includegraphics[width=8.0cm]{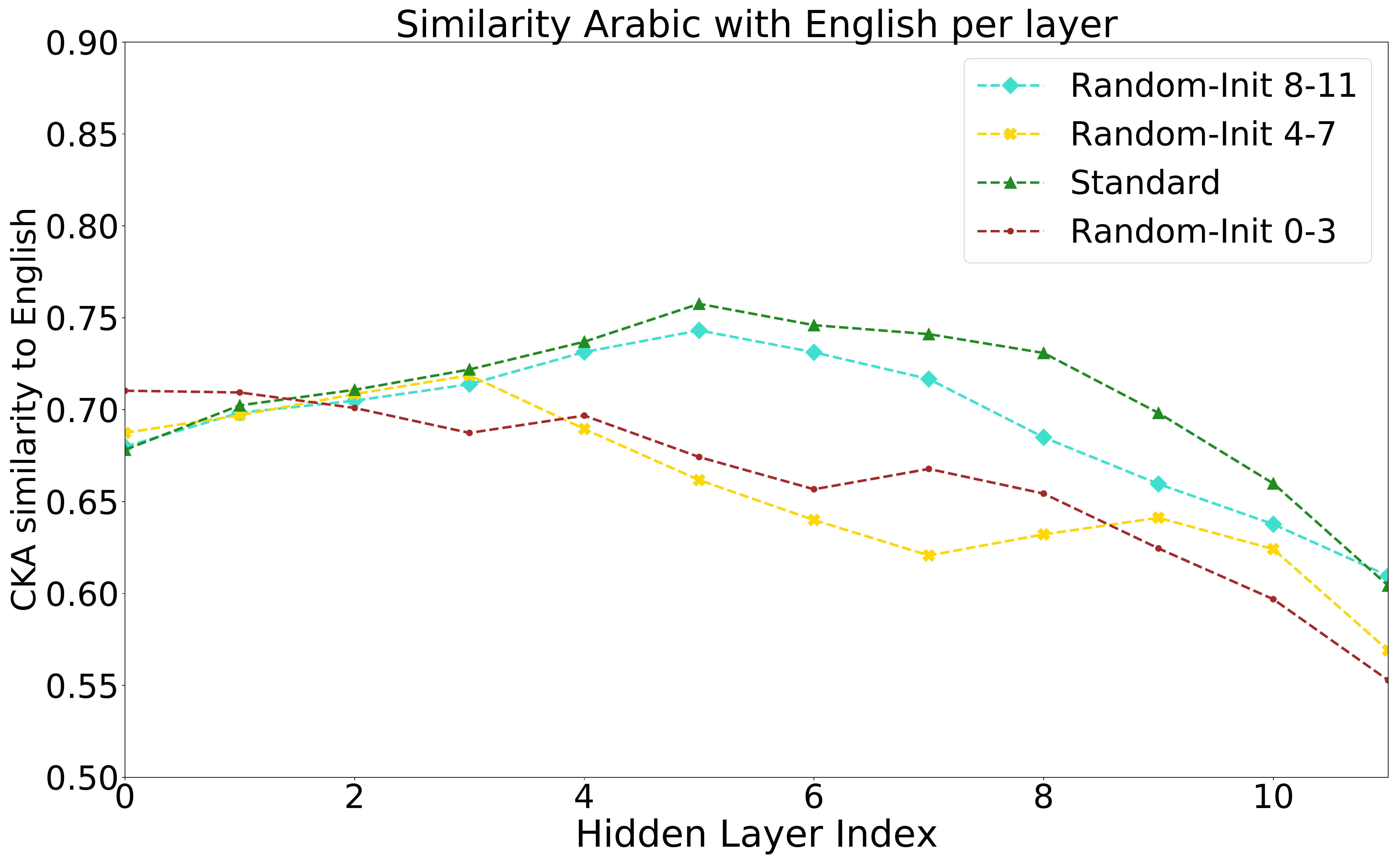}
}
\caption{Cross-Lingual similarity (CKA) (\S \ref{sec:alignement}) of hidden representations of a source language (English) sentences with target languages sentences on fine-tuned \textbf{Parsing} models with and without \reinit. The higher the CKA value the greater the similarity.}
\label{fig:plot_parsing_random_init}
\end{figure*}

\newpage


\begin{figure*}[ht]
\centering
\subfloat[]{
\includegraphics[width=1\columnwidth]{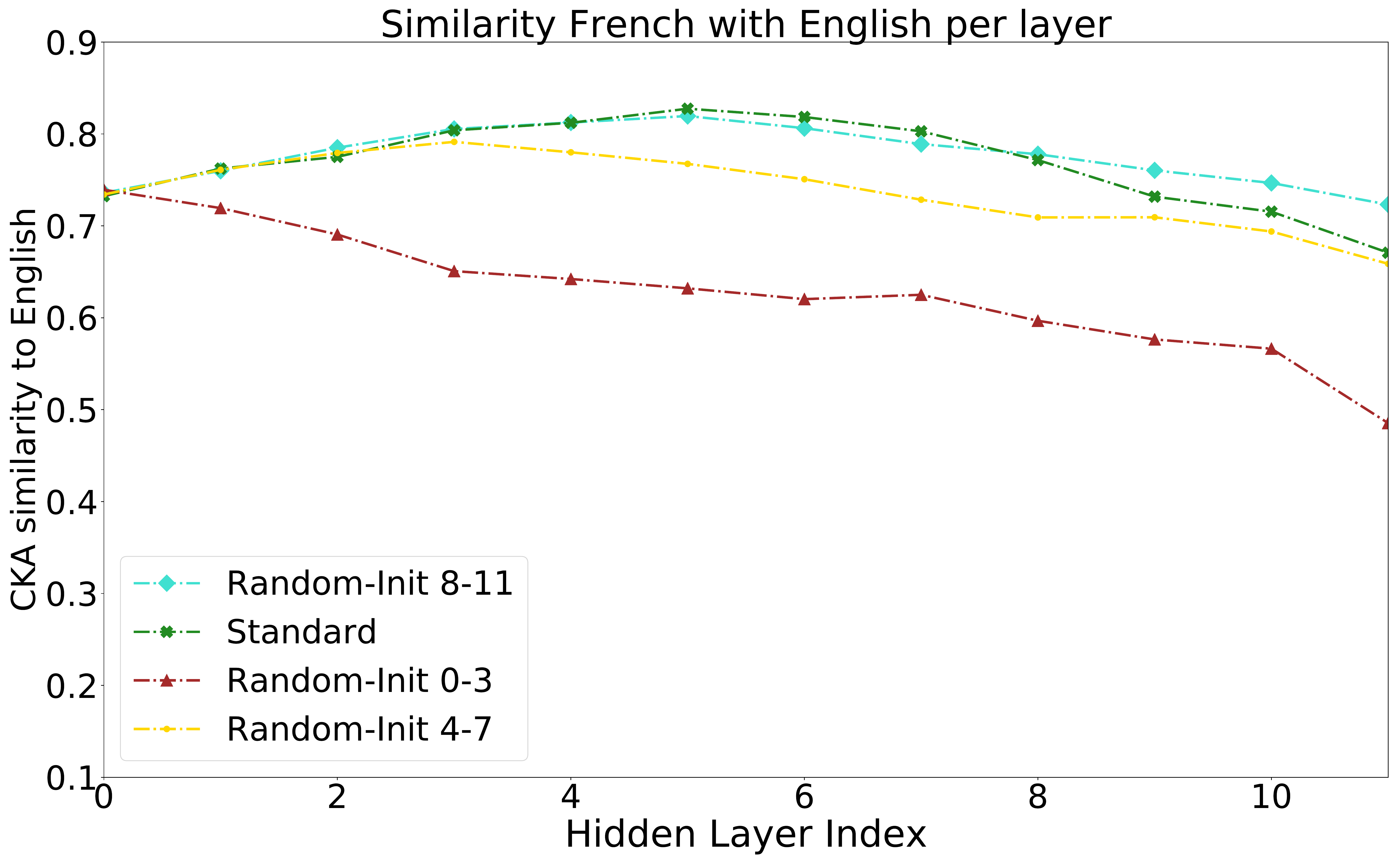}
\label{fig:vehicle_speed}
}
\subfloat[]{
\includegraphics[width=1\columnwidth]{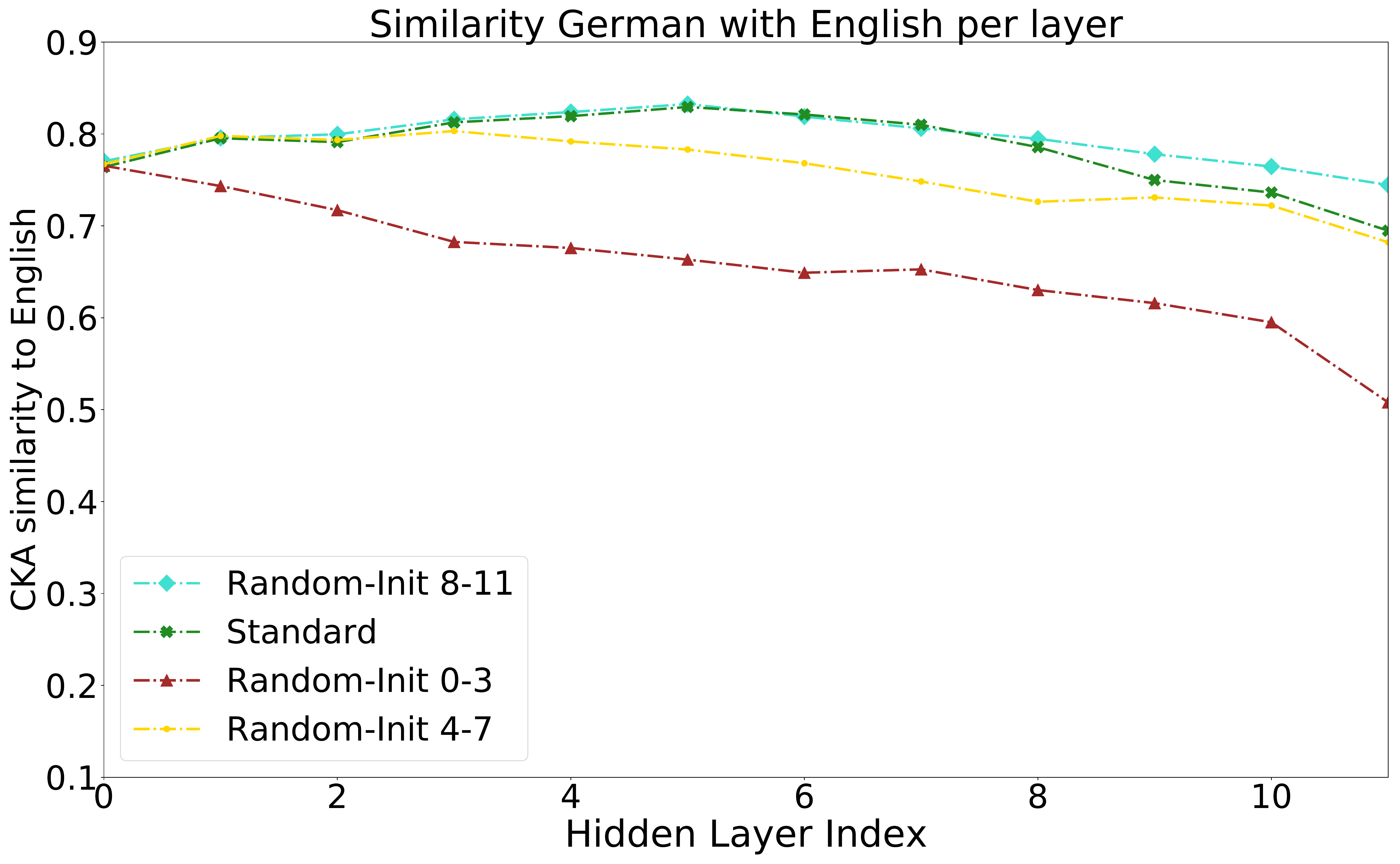}
\label{fig:vehicle_speed}
}\\
\subfloat[]{
\includegraphics[width=1\columnwidth]{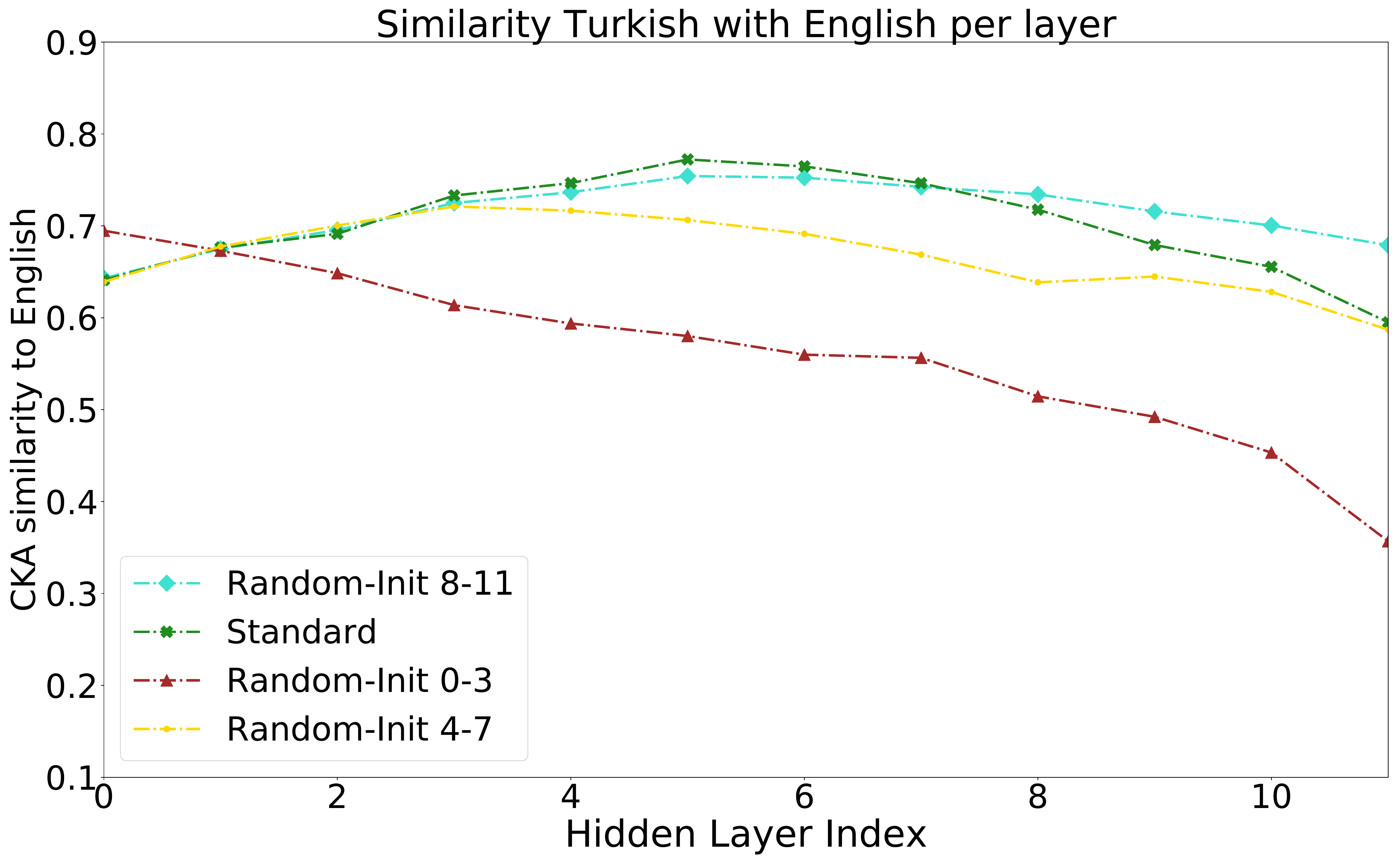}
\label{fig:vehicle_speed}
}
\subfloat[]{
\includegraphics[width=1\columnwidth]{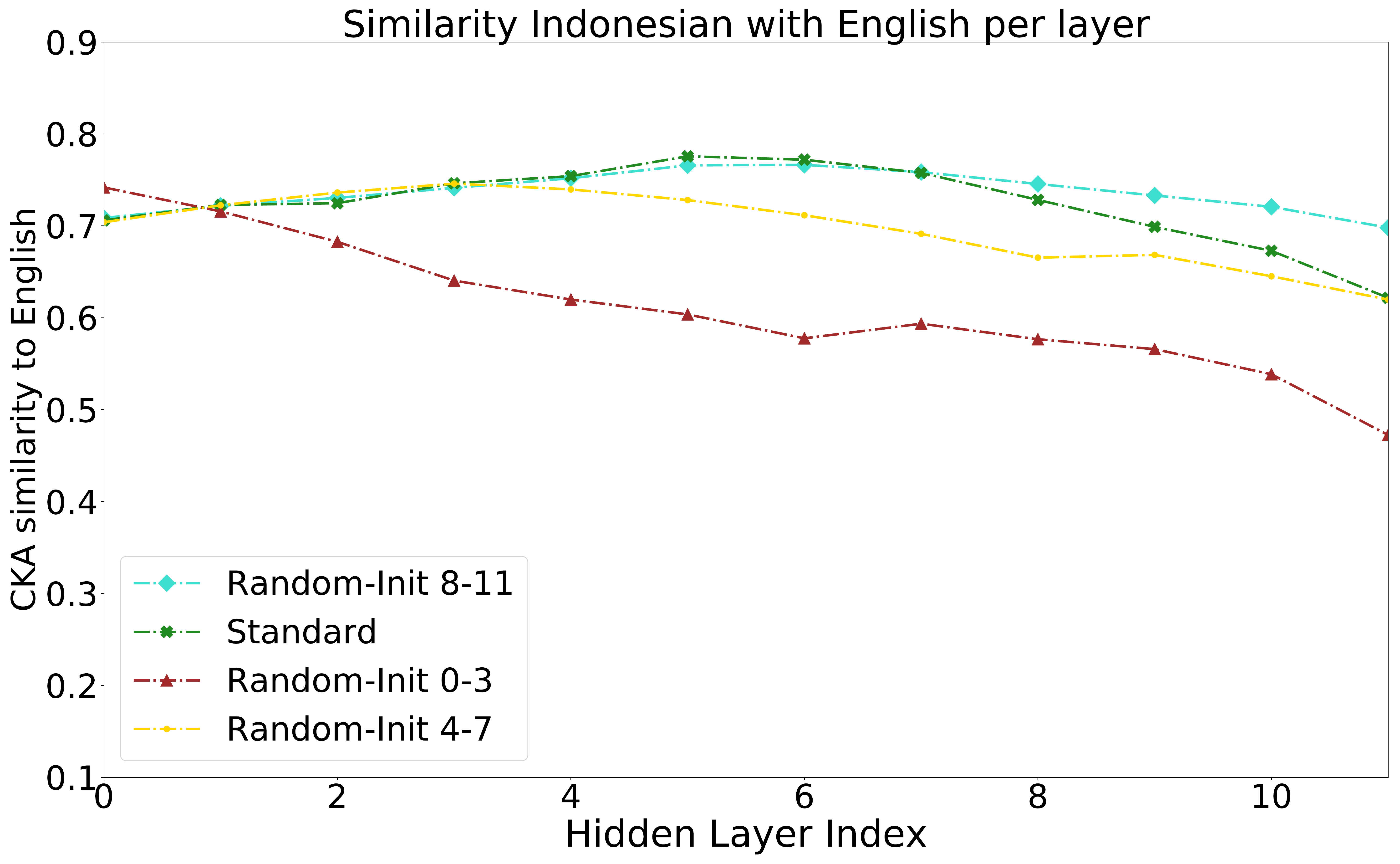}
\label{fig:vehicle_speed}
}
\\
\subfloat[]{
\includegraphics[width=1\columnwidth]{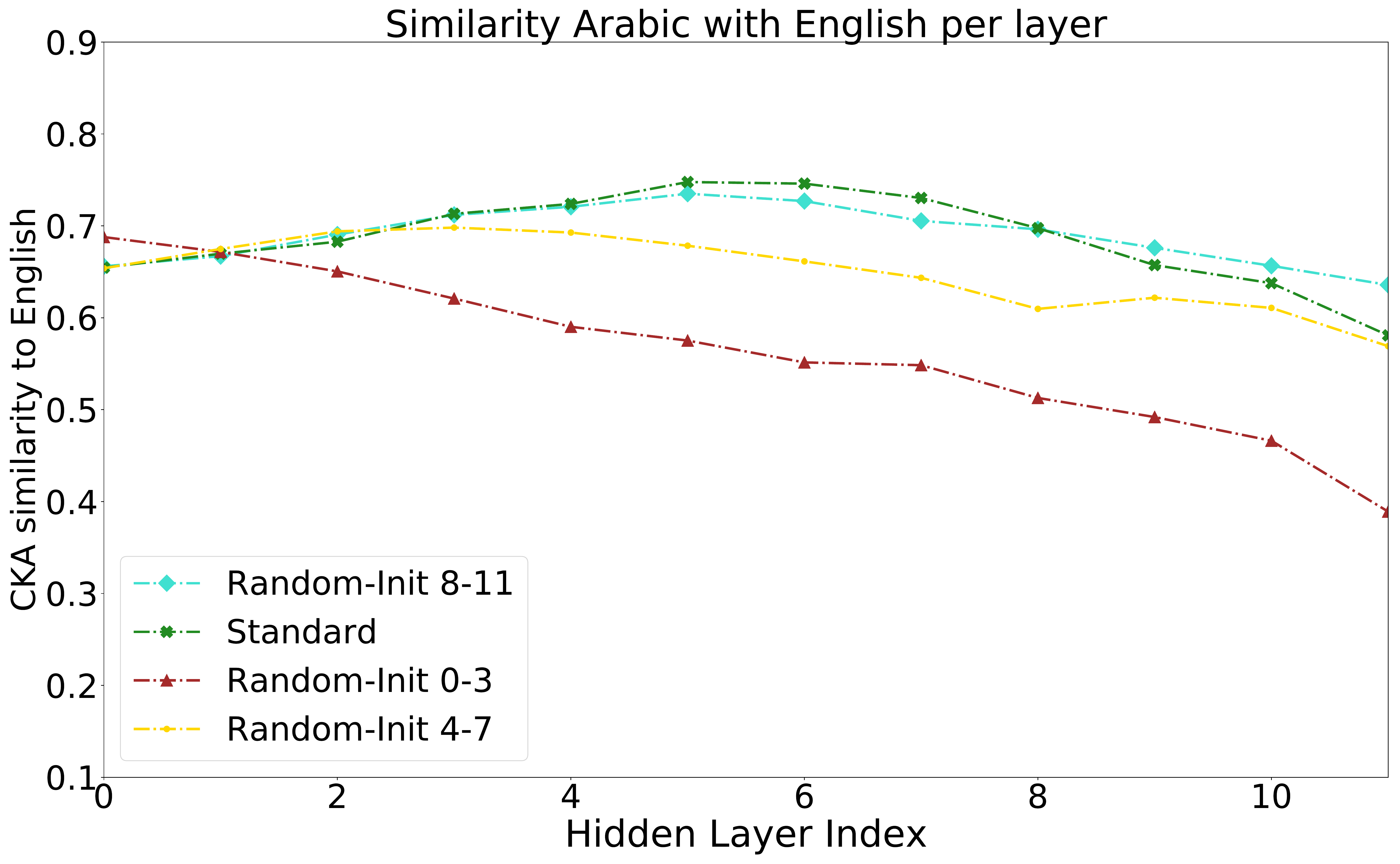}
}
\caption{Cross-Lingual similarity (CKA) (\S \ref{sec:alignement}) of hidden representations of a source language (English) sentences with target languages sentences on fine-tuned \textbf{POS} models with and w/o \reinit. The higher the CKA value the greater the similarity.}
\label{fig:plot_pos_random_init}
\end{figure*}

\newpage


\begin{figure*}[ht]
\centering
\subfloat[]{
\includegraphics[width=1\columnwidth]{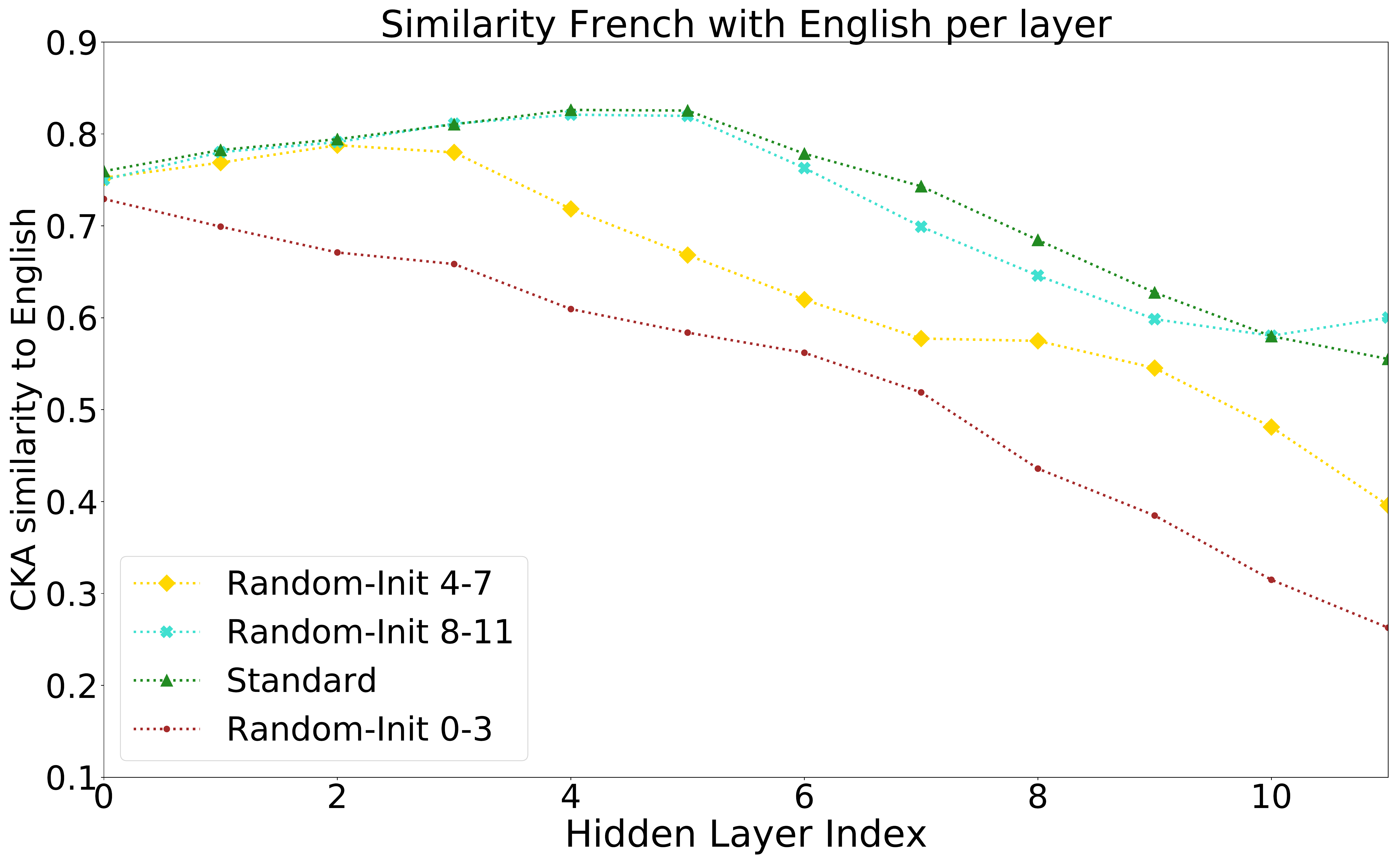}
\label{fig:vehicle_speed}
}
\subfloat[]{
\includegraphics[width=1\columnwidth]{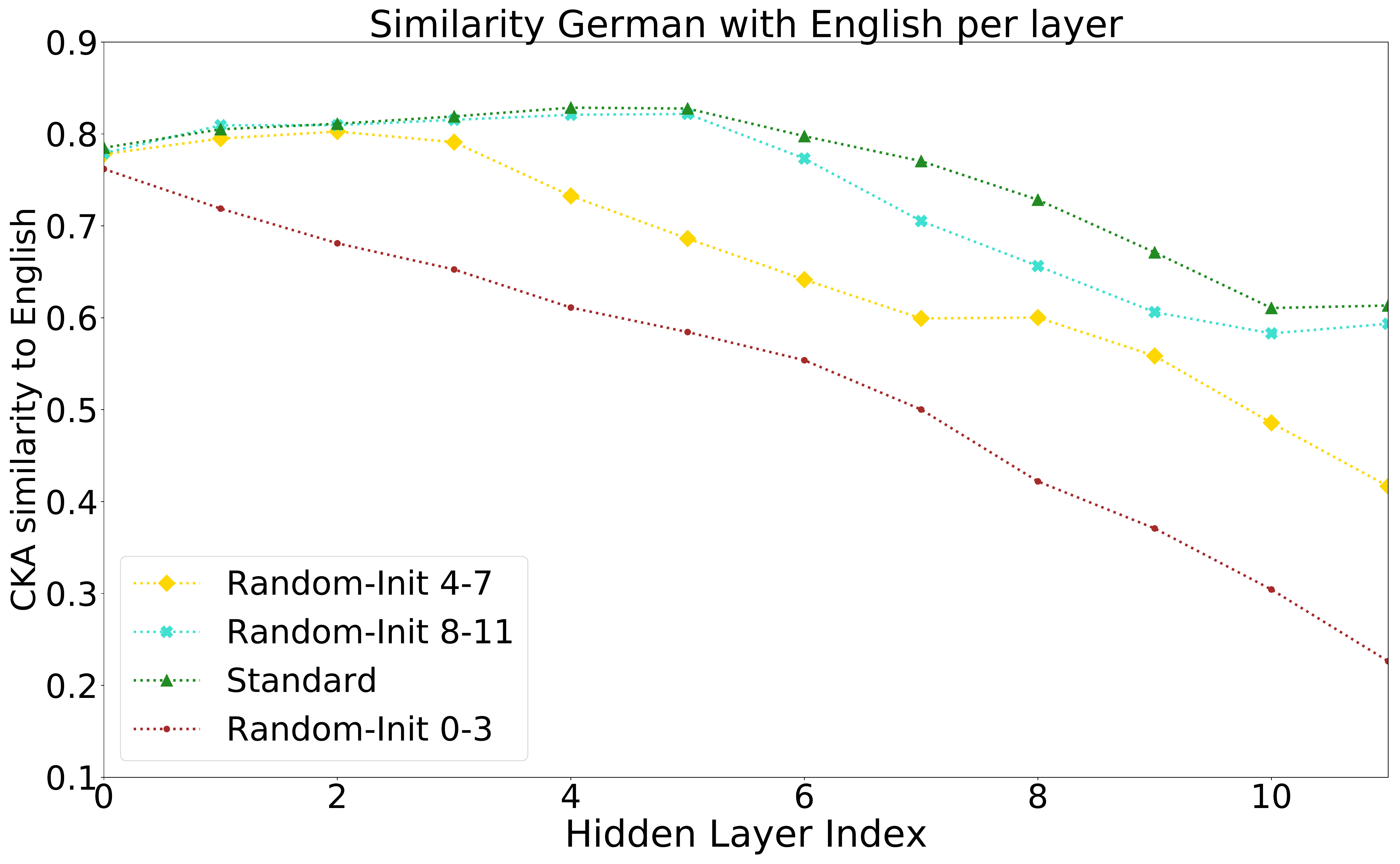}
\label{fig:vehicle_speed}
}\\
\subfloat[]{
\includegraphics[width=1\columnwidth]{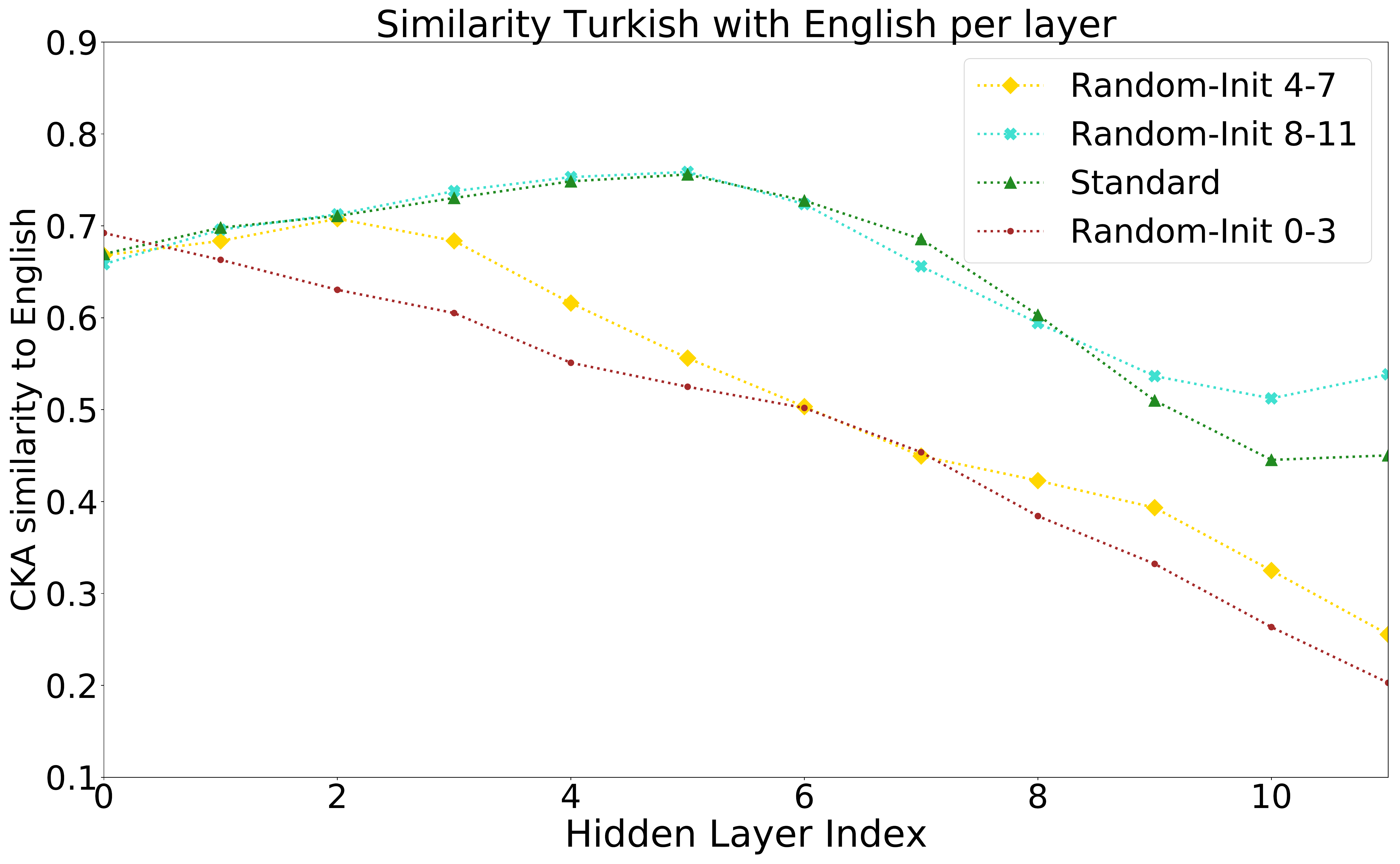}
\label{fig:vehicle_speed}
}
\subfloat[]{
\includegraphics[width=1\columnwidth]{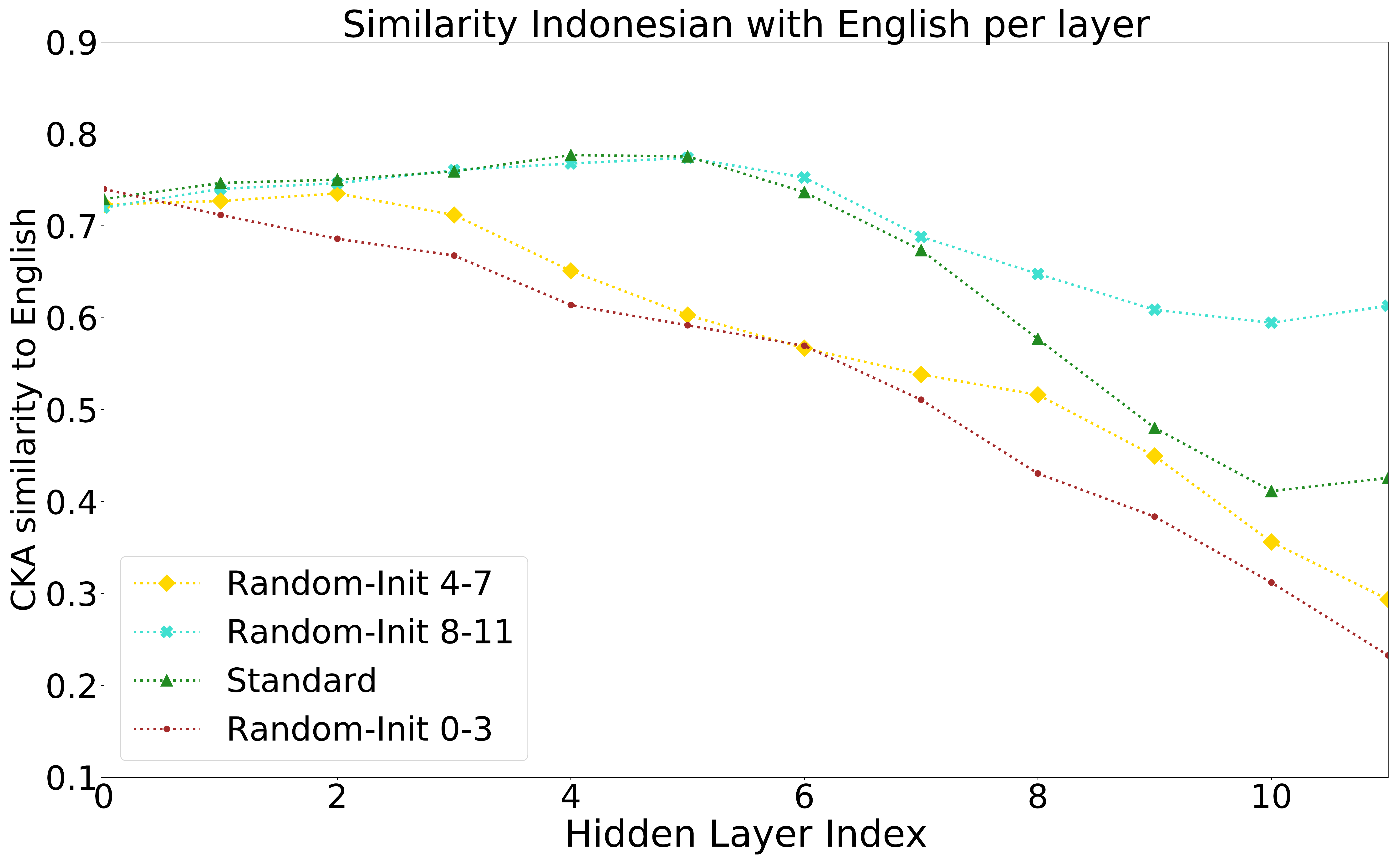}
\label{fig:vehicle_speed}
}
\\
\subfloat[]{
\includegraphics[width=1\columnwidth]{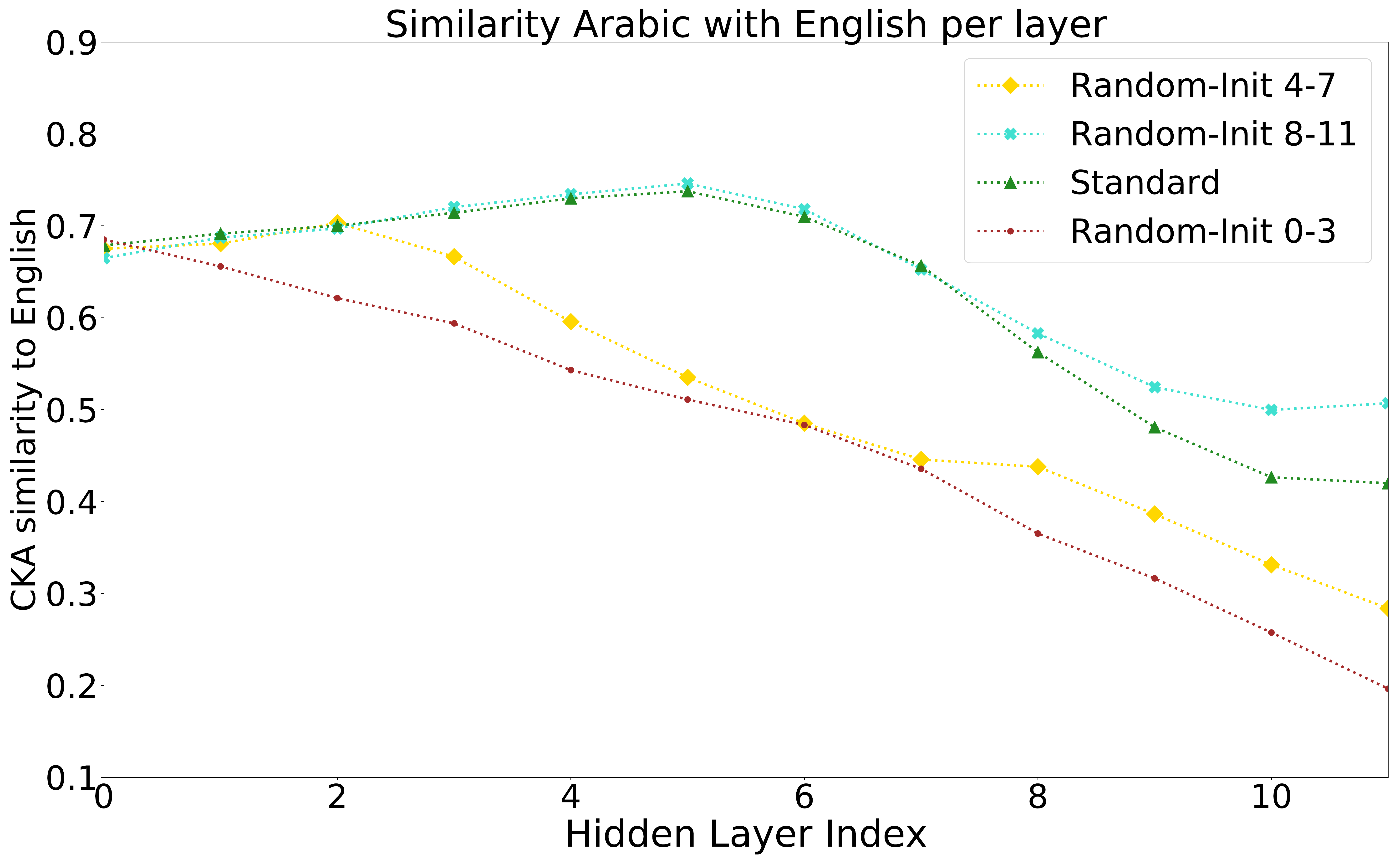}
}
\caption{Cross-Lingual similarity (CKA) (\S \ref{sec:alignement}) of hidden representations of a source language (English) sentences with target languages sentences on fine-tuned \textbf{NER} models with and w/o \reinit. The higher the CKA value the greater the similarity.}
\label{fig:plot_ner_random_init}
\end{figure*}




\newpage

\end{document}